\documentclass[pdflatex,sn-mathphys-num]{sn-jnl}

\usepackage{graphicx}%
\usepackage{multirow}%
\usepackage{amsmath,amssymb,amsfonts}%
\usepackage{amsthm}%
\usepackage{mathrsfs}%
\usepackage[title]{appendix}%
\usepackage{xcolor}%
\usepackage{textcomp}%
\usepackage{manyfoot}%
\usepackage{booktabs}%
\usepackage{algorithm}%
\usepackage{algorithmicx}%
\usepackage{algpseudocode}%
\usepackage{listings}%
\usepackage{array}
\usepackage{tabularx}
\usepackage{caption}
\usepackage{ulem}
\usepackage{circuitikz}
\usepackage{afterpage}
\usepackage{ulem}
\usepackage{pdflscape}
\usepackage{enumitem}

\theoremstyle{thmstyleone}%
\newtheorem{theorem}{Theorem}
\newtheorem{proposition}{Proposition}%
\newtheorem{lemma}{Lemma}

\theoremstyle{thmstyletwo}%

\theoremstyle{thmstylethree}%
\newtheorem{definition}{Definition}%

\usepackage[utf8]{inputenc} 

\begin{document}

\title[Article Title]{SAT Encoding for Set of At-Most-K Cardinality Constraints with Ladder Shape}

\author[1]{\fnm{Hieu} \sur{Truong Xuan}}\email{truongxuanhieu11@vnu.edu.vn}
\equalcont{These authors contributed equally to this work.}

\author[1]{\fnm{Tuyen} \sur{Kieu Van}}\email{tuyenkv@vnu.edu.vn}

\author*[1]{\fnm{Khanh} \sur{To Van}}\email{khanhtv@vnu.edu.vn}
\equalcont{These authors contributed equally to this work.}

\affil[1]{\orgdiv{University of Engineering and Technology}, \par \orgname{Vietnam National University}, \city{Hanoi}, \country{Vietnam}}

\abstract{This paper presents a novel SAT encoding designed to represent a sequence of At-Most-K (AMK) cardinality constraints arranged in a ladder-like structure, referred to as the Ladder AMK constraint. 
These AMK constraints feature a ladder-like arrangement, with shared common variables in the center and two variables differing at the beginning and the end. 
The proposed encoding first decomposes the sequence of AMK constraints into separated blocks, then encodes each block as a whole rather than encoding each AMK constraint individually. 
This approach improves the efficiency of constraint solving in SAT solvers by significantly reducing the size of the encoding, precisely the number of variables, and the generated clauses required.
To achieve this, the encoding employs \textit{Sequential Counter} to encode each block and constructs clauses that link adjacent blocks, ensuring equivalence with the input constraints. 
Compared to the previously proposed encoding that uses \textit{Binary Decision Diagrams} for Ladder AMO constraint (\( k = 1 \)), our encoding requires fewer variables and clauses. 
Our study is the first study that develops SAT encoding for Ladder AMK constraints with \( k > 1 \).
Experimental results on single Ladder AMO and Ladder AMK constraints indicate that our encoding significantly enhances both the size of the encoding and the efficiency of solving time compared to existing encodings. 
Furthermore, by evaluating on the two real-world combinatorial problems, namely Anti-bandwidth Problem and Nurse Rostering Problem, we found that the proposed encoding surpasses alternative encodings, as well as more advanced Constraint Programming (CP) and Mixed Integer Programming (MIP) solvers.
In summary, the SAT encoding technique presented in this paper provides a more efficient method for encoding sequence constraint, specifically Ladder AMK constraint, which could enhance the effectiveness of SAT solving for various problems involving this type of constraint.}


\maketitle

\section{Introduction}\label{sec:introduction}

Sequence constraint is a type of constraint that frequently appears in combinatorial optimization problems, such as production line scheduling, employee scheduling, Nurse Rostering Problem, Crew Rostering Problem, sports timetabling, and OR problems. 
For instance, in the Car Sequencing Problem \cite{artigues2014sat, siala2015search}, sequence constraints restrict the maximum number of cars with a specific option to no more than two in any sequence of five consecutive cars on a production line. 
In the Nurse Rostering Problem \cite{ceschia2015second, kletzander2020solving}, sequence constraints may limit a nurse to working at most four night shifts and eight evening shifts, while also requiring at least four off-days within any sequence of fourteen consecutive days \cite{bergman2014mdd}.
These constraints help ensure proper rest and recovery by preventing excessively intense work schedules.
Similarly, in the Crew Rostering Problem, an employee may be allowed to work a maximum of five days in any seven consecutive days \cite{siala2014optimal} to comply with labor regulations.
In the taxi driver shift scheduling problem at an airport for a car rental company \cite{nieuwenhuis2021employee}, sequence constraints are used to ensure a balanced distribution of morning and afternoon shifts among drivers.
In the RobinX sports timetabling problem \cite{lester2022pseudo}, sequence constraints limit the maximum number of home or away games a team can play within any fixed-length sequence of consecutive matches. 
Sequence constraints limit the number of occurrences of certain specific values in a sequence of $k$ variables and are often referred to as AmongSeq \cite{bessiere2007reformulating} and AtMostSeq \cite{artigues2014sat}.
Certain problems, such as the Anti-bandwidth Problem (ABP) \cite{sinnl2021note}, can be reformulated to incorporate sequence constraints. This problem has applications in scheduling, radio frequency assignment, obnoxious facility location, and map coloring.

The impressive progress of SAT solvers has initiated what is now referred to as the SAT revolution \cite{vardi2015sat}. These solvers are capable of addressing enormous formulas that contain millions of variables and clauses. In the near future, they are expected to handle formulas with significantly more than just millions of variables and clauses. Today, SAT solving has become a serious alternative to Integer Linear Programming (ILP) solvers for addressing NP-complete problems. Due to substantial improvements in the performance of SAT solvers, SAT solving is increasingly being used in various real-world applications. However, when applying SAT encoding to combinatorial problems, a significant number of clauses can be generated. To mitigate this issue, several encoding techniques have been proposed to reduce the overwhelming number of clauses \cite{haberlandt2023effective, vasconcellos2020abacus}. 

Sequence constraints have been extensively studied in the field of Constraint Programming (CP) to develop efficient solutions, as highlighted in the work of Bergman et al. \cite{bergman2014mdd} and Siala et al. \cite{siala2015search}. 
However, effective SAT encodings for sequence constraints, such as AmongSeq \cite{bessiere2007reformulating} and AtMostSeq \cite{artigues2014sat}, remain largely unexamined. 
A notable exception is \textit{Duplex} encoding \cite{fazekas2020duplex}, which is specifically designed for sequences of At-Most-One (AMO) constraints. 
In our research, we propose a block-based encoding method for a sequence of At-Most-K (AMK) constraints, which encodes them as a whole rather than handling each AMK constraint independently. 
In particular, we leverage shared variables among neighboring AMK constraints to decompose the input sequence constraint into blocks. 
Then, we encode these blocks using an encoding inspired by the \textit{Sequential Counter} \cite{sinz2005towards} encoding. 
Auxiliary variables generated from the block encoding are used to connect adjacent blocks, thereby reformulating the original constraint. 
This approach not only demonstrates greater efficiency for sequences of AMO constraints compared to the \textit{Duplex} encoding, as reported by Hieu et al. \cite{Truong2025}, but also proves to be effective for sequences of AMK constraints with \( k > 1 \).

\begin{itemize}
    \item \textit{We propose a novel SAT encoding for a set of AMK cardinality constraints arranged in a ladder shape. Our approach leverages shared variables among neighboring AMK constraints, allowing us to organize them into blocks for encoding rather than encoding each constraint independently. To the best of our knowledge, this is the first SAT encoding method for the sequence of AMK cardinality constraints with a ladder shape where \( k > 1 \)}. 
    \item \textit{Additionally, we present solutions to the Anti-bandwidth Problem and Nurse Rostering Problem, using our new SAT encoding for the sequence of At-Most-One and At-Most-K cardinality constraints, respectively. Experimental results show that our approach outperforms existing SAT encodings, as well as Constraint Programming (CP) and Mixed Integer Programming (MIP) methods that utilize powerful solvers such as CPLEX, CP-SAT, and Gurobi.} 
\end{itemize}

This paper extends our previously published research \cite{Truong2025} in the following ways:
First, we generalize the encoding from AMO constraints to AMK constraints with \( k \geq 1 \) for sequences of cardinality constraints in a ladder shape, whereas the previous work focused solely on AMO constraints. 
Second, we also include an experiment on the Nurse Rostering Problem to evaluate the performance of the proposed encoding on combinatorial problems that include sequences of AMK constraints.
Third, for the Anti-bandwidth Problem, we develop a parallel search implementation, leading to improved performance.
We also expanded our experiment to include comparisons and evaluations with \textit{Adder Network} encoding \cite{een2006translating} and powerful CP and MIP solvers, namely \textit{Gurobi} and \textit{CP-SAT}. This comparison highlights the effectiveness of our proposed SAT encoding in solving the Anti-bandwidth Problem and Nurse Rostering Problem when compared to CP and MIP solvers.
 
The paper is organized as follows: Section 1 introduces the research problem and summarizes the contributions of this study. Section 2 presents the basic definitions of the Ladder AMK constraint and discusses how to represent one AMK constraint using two sub-AMK constraints.
Section 3 details the block encoding for the Ladder AMK constraint and includes a comparative evaluation against existing SAT encodings. Section 4 introduces the encoding for the Ladder AMO constraint, which is a specific case of the Ladder AMK constraint with \( k = 1 \).
Sections 5 and 6 sequentially explain the approach to solving the Anti-bandwidth Problem and Nurse Rostering Problem using SAT solving. Section 7 provides experimental results that compare our proposed encoding with existing SAT encodings, as well as state-of-the-art CP and MIP solvers. Finally, Section 8 concludes the paper.

\section{Preliminaries}
\label{sec:preliminaries}

\subsection{Ladder AMK constraint}
\label{sec:ladder-amk-definition}

A Ladder AMK constraint is defined as a sequence of AMK cardinality constraints that involve \( w \) boolean variables. In this structure, each pair of adjacent AMK constraints shares \( w - 1 \) variables. They differ only in the first variable of the first constraint and the last variable of the second constraint. This overlapping feature results in a ladder-like shape. The formal definition of the Ladder AMK constraint is provided in Definition \ref{def:ladder_amk}.

\vspace{0.5cm}
\begin{definition}
    \label{def:ladder_amk}
    Given a sequence of $n$ boolean variables $\Omega = \{x_{1}, x_{2},..., x_{n}\}$ and a number $w$ such that $1 < w \leq n$, a Ladder AMK constraint of width $w$ is formulated as follows:
    \[Ladder(\Omega, w)=\bigwedge_{i=0}^{n-w}(\sum_{j=i+1}^{i+w}x_{j} \leq k) \hspace{1cm} 1 \leq k < w\]
\end{definition}

\vspace{0.5cm}

\begin{figure}[ht!]
    \centering
    \resizebox{\textwidth}{!}{
        \begin{tabular}{llllllllllllllllllll}
        \cline{1-8}
         &  &  &  &  &  &  & \multicolumn{1}{l|}{} &  &  &  &  &  &  &  &  &  &  &  &  \\ \cline{9-10}
        $x_{1}$ & + & $x_{2}$ & + & $x_{3}$ & + & $x_{4}$ &  &  & \multicolumn{1}{l|}{} &  &  &  &  &  &  &  &  &  & $\leq k\wedge$ \\ \cline{11-12}
         &  & $x_{2}$ & + & $x_{3}$ & + & $x_{4}$ & + & $x_{5}$ &  &  & \multicolumn{1}{l|}{} &  &  &  &  &  &  &  & $\leq k\wedge$ \\ \cline{13-14}
         &  &  &  & $x_{3}$ & + & $x_{4}$ & + & $x_{5}$ & + & $x_{6}$ &  &  & \multicolumn{1}{l|}{} &  &  &  &  &  & $\leq k\wedge$ \\ \cline{15-16}
         &  &  &  &  &  & $x_{4}$ & + & $x_{5}$ & + & $x_{6}$ & + & $x_{7}$ &  &  & \multicolumn{1}{l|}{} &  &  &  & $\leq k\wedge$ \\ \cline{17-18}
         &  &  &  &  &  &  &  & $x_{5}$ & + & $x_{6}$ & + & $x_{7}$ & + & $x_{8}$ &  &  & \multicolumn{1}{l|}{} &  & $\leq k\wedge$ \\ \cline{19-19}
         &  &  &  &  &  &  &  &  &  & $x_{6}$ & + & $x_{7}$ & + & $x_{8}$ & + & $x_{9}$ &  &  & $\leq k\wedge$ \\
         &  &  &  &  &  &  &  &  &  &  &  & $x_{7}$ & + & $x_{8}$ & + & $x_{9}$ & + & $x_{10}$ & $\leq k$
        \end{tabular}
    }
    \caption{Ladder AMK constraint of 10 variables and width 4.}
    \label{fig:ladder_amk_constraint_10_vars_4_width}
\end{figure}

Figure \ref{fig:ladder_amk_constraint_10_vars_4_width} illustrates a Ladder AMK constraint over a sequence of 10 boolean variables $\Omega = \{x_{1}, x_{2},..., x_{10}\}$ of width $w=4$. This Ladder AMK constraint consists of seven AMK constraints, each involving four consecutive variables in the sequence.

As discussed in Section~\ref{sec:introduction}, a wide range of combinatorial problems can be modeled using sequence constraints, particularly the Ladder AMK constraint. For instance, in the Car Sequencing Problem~\cite{artigues2014sat, siala2015search}, let $c_{i,o}$ be a Boolean variable that indicates whether the $i^{th}$ car is equipped with option $o$, i.e., $c_{i,o}$ is $true$ if the $i^{th}$ car is equipped with option $o$ and $false$ otherwise. Given a production sequence of 10 cars, the constraint requiring that, in any 5 consecutive cars, at most 2 cars are equipped with the option $o$ can be modeled as a Ladder AMK constraint with $n=10$, $w = 5$, and $k=2$, as illustrated in Figure~\ref{fig:ladder-amk-constraint-car-sequencing-10-5-2}.

\begin{figure}[ht!]
    \centering
    \resizebox{!}{!}{
    \begin{tabular}{p{0.25cm}p{0.1cm}p{0.25cm}p{0.1cm}p{0.25cm}p{0.1cm}p{0.25cm}p{0.1cm}p{0.25cm}p{0.1cm}p{0.25cm}p{0.1cm}p{0.25cm}p{0.1cm}p{0.25cm}p{0.1cm}p{0.25cm}p{0.1cm}p{0.25cm}ll}
    $c_{1,o}$ & + & $c_{2,o}$ & + & $c_{3,o}$ & + & $c_{4,o}$ & + & $c_{5,o}$ &  &  &  &  &  &  &  &  &  &  &  & $\leq 2\wedge$ \\
     &  & $c_{2,o}$ & + & $c_{3,o}$ & + & $c_{4,o}$ & + & $c_{5,o}$ & + & $c_{6,o}$ &  &  &  &  &  &  &  &  &  & $\leq 2\wedge$ \\
     &  &  &  & $c_{3,o}$ & + & $c_{4,o}$ & + & $c_{5,o}$ & + & $c_{6,o}$ & + & $c_{7,o}$ &  &  &  &  &  &  &  & $\leq 2\wedge$ \\
     &  &  &  &  &  & $c_{4,o}$ & + & $c_{5,o}$ & + & $c_{6,o}$ & + & $c_{7,o}$ & + & $c_{8,o}$ &  &  &  &  &  & $\leq 2\wedge$ \\
     &  &  &  &  &  &  &  & $c_{5,o}$ & + & $c_{6,o}$ & + & $c_{7,o}$ & + & $c_{8,o}$ & + & $c_{9,o}$ &  &  &  & $\leq 2\wedge$ \\
     &  &  &  &  &  &  &  &  &  & $c_{6,o}$ & + & $c_{7,o}$ & + & $c_{8,o}$ & + & $c_{9,o}$ & + & $c_{10,o}$ &  & $\leq 2$
    \end{tabular}
    }
    \caption{Ladder AMK constraint ensuring that for any 5 consecutive cars in a sequence of 10 cars, at most 2 cars can be equipped with option $o$.}
    \label{fig:ladder-amk-constraint-car-sequencing-10-5-2}
\end{figure}

In the Nurse Rostering Problem~\cite{ceschia2015second, kletzander2020solving, bergman2014mdd}, let $x_{i,j,s}$ be a Boolean variable indicating whether nurse $i$ is assigned to shift $s \in \{D, E, N, O\}$ on day $j^{th}$, where $D$, $E$, and $N$ denote day, evening, and night shifts, respectively, while $O$ represents an off-day. 
Given a planning horizon of 28 consecutive days, the constraint requiring that nurse $i$ is assigned to at most 8 night shifts in any 14 consecutive days can be modeled as a Ladder AMK constraint with $n=28$, $w=14$, and $k=8$, as illustrated in Figure~\ref{fig:ladder-amk-constraint-nrp-most-8N-every-28}.

\begin{figure}[ht!]
    \centering
    \resizebox{\textwidth}{!}{
    \begin{tabular}{p{0.75cm}p{0.1cm}p{0.75cm}p{0.1cm}p{0.75cm}p{0.1cm}p{0.75cm}p{0.1cm}p{0.75cm}p{0.1cm}p{0.75cm}p{0.1cm}p{0.75cm}p{0.1cm}p{0.75cm}cl}
    $x_{i,1,N}$ & + & $x_{i,2,N}$ & + & $\ldots$ & + & $x_{i,14,N}$ &  &  &  &  &  &  &  &  &  & $\leq 8\wedge$ \\
     &  & $x_{i,2,N}$ & + & $\ldots$ & + & $x_{i,14,N}$ & + & $x_{i,15,N}$ &  &  &  &  &  &  &  & $\leq 8\wedge$ \\
     &  & \multicolumn{11}{c}{$\ldots$} &  &  & \multicolumn{1}{l}{} & $\leq 8\wedge$ \\
     &  &  &  &  &  & $x_{i,14,N}$ & + & $x_{i,15,N}$ & + & $\ldots$ & + & $x_{i,27,N}$ &  &  & \multicolumn{1}{l}{} & $\leq 8\wedge$ \\
     &  &  &  &  &  &  &  & $x_{i,15,N}$ & + & $\ldots$ & + & $x_{i,27,N}$ & + & $x_{i,28,N}$ &  & $\leq 8$
    \end{tabular}
    }
    \caption{Ladder AMK constraint ensuring that nurse $i$ works at most 8 night shifts in any 14 consecutive days over a 28-day planning horizon.}
    \label{fig:ladder-amk-constraint-nrp-most-8N-every-28}
\end{figure}

In addition to the examples above, many constraints arising in other combinatorial and scheduling problems can also be modeled as Ladder AMK constraints, such as the Crew Rostering Problem \cite{siala2014optimal}, the taxi driver shift scheduling problem \cite{nieuwenhuis2021employee}, the RobinX sports timetabling problem \cite{lester2022pseudo}, and the Anti-bandwidth Problem \cite{fazekas2020duplex, Truong2025}, as discussed in Section~\ref{sec:introduction}. In general, any requirement that limits the number of selected activities, resources, or events within a window over an ordered sequence can be expressed in this form. Therefore, the Ladder AMK constraint is not restricted to specific benchmark problems, but constitutes a general and practically relevant modeling primitive. Its ability to capture sequence cardinality restrictions highlights its wide applicability and practical significance in real-world optimization problems.

\subsection{Decomposition of AMK constraint}
\label{sec:amk-decomposition}
An AMK constraint can be decomposed into two separate AMK constraints applied to two subsets of the original sequence of variables. This decomposition is formally explained in Proposition \ref{prop:decompose_AMK_constraint}. It also incorporates connecting clauses to ensure that the decomposed constraints comply with the original constraint without violating it. \\

\begin{proposition}
    For any integer number $i$ s.t. $1 \leq i < w$, the constraint 
    \[x_{1} + x_{2} + \ldots + x_{w} \leq k\]
    holds if and only if the following constraint holds:
    \[(x_{1} + \ldots + x_{i}\leq k)\wedge (x_{i+1} + \ldots + x_{w}\leq k)\]
    \[\wedge \bigwedge_{j = 1}^{k} (x_{1} + \ldots + x_{i}\leq k - j \vee x_{i+1} + \ldots + x_{w}\leq j - 1)\]
    \label{prop:decompose_AMK_constraint}
\end{proposition}

Proposition \ref{prop:decompose_AMK_constraint} can be proven to be correct by considering all possible cases in which one of the two sub-expressions has a total of \( k, k-1, k-2, \ldots, 1 \). In each of these cases, the other sub-expression must have a corresponding total that is less than or equal to \( 0, 1, 2, \ldots, k-1 \), thereby the sum of these two expressions never exceeds $k$.
The following example illustrates the decomposition of an AMK constraint with $k = 3$ over 8 variables $\{x_1, \ldots, x_8\}$.

\begin{minipage}{0.25\textwidth}
    \centering
    \[\sum_{i=1}^8 x_i \leq 3 \hspace{0.5cm} \equiv\]
\end{minipage}
\begin{minipage}{0.6\textwidth}
    \centering
    \begin{align*}
    &(x_1 + x_2 + x_3 + x_4 \leq 3)\wedge (x_5 + x_6 + x_7 + x_8 \leq 3) \\
    \wedge (&(x_1 + x_2 + x_3 + x_4 \leq 2) \vee (x_5 + x_6 + x_7 + x_8 \leq 0)) \\
    \wedge (&(x_1 + x_2 + x_3 + x_4 \leq 1) \vee (x_5 + x_6 + x_7 + x_8 \leq 1)) \\
    \wedge (&(x_1 + x_2 + x_3 + x_4 \leq 0) \vee (x_5 + x_6 + x_7 + x_8 \leq 2)),
    \\
    \end{align*}
\end{minipage}

The decomposition of the AMK constraint serves a crucial role in the proposed encoding. In subsequent sections, the proposed encoding leverages this decomposition to replace a large AMK constraint with smaller sub-constraints that can be reused across consecutive constraints. As a result, the proposed encoding achieves a reduced encoding size of the resulting SAT formulation.

The Ladder AMO constraint is a specific case of the Ladder AMK constraint, which occurs when $k = 1$. To derive the decomposition of the AMO constraint, one can simply substitute $k$ with $1$ in the decomposition of the AMK constraint.
Lemma \ref{lemma:decompose_AMO_constraint} provides a formal decomposition of an AMO constraint that involves $w$ variables.

\begin{lemma}
    For any integer number $i$ s.t. $1 \leq i < w$, the constraint 
    \[x_{1} + x_{2} + \ldots + x_{w} \leq 1\]
    holds if and only if the following constraint holds:
   \[(x_{1} + \ldots + x_{i}\leq 1)\wedge (x_{i+1} + \ldots + x_{w}\leq 1)\]
   \[\wedge (x_{1} + \ldots + x_{i}\leq 0 \vee x_{i+1} + \ldots + x_{w}\leq 0)\]
    \label{lemma:decompose_AMO_constraint}
\end{lemma}

\section{Sequential Counter encoding for Ladder constraint}
\label{sec:ladder_amk_constraint}

This section introduces the proposed encoding called \textbf{\textit{SCL}} (\textit{Sequential Counter encoding for Ladder constraints}) for encoding Ladder AMK constraints. The \textit{SCL} encoding leverages shared variables among neighboring AMK constraints to partition a Ladder AMK constraint into blocks. These blocks are then represented using an encoding inspired by the \textit{Sequential Counter} \cite{sinz2005towards}, which we refer to as the \textit{SC} encoding. After that, auxiliary variables generated by the \textit{SC} encoding are incorporated into clauses that connect adjacent blocks through Proposition \ref{prop:decompose_AMK_constraint}, thus reformulating the original constraint. By exploiting block decomposition to encode the sequence of AMK constraints as a whole, instead of encoding each AMK constraint independently, the proposed encoding achieves a more compact and efficient representation.

\subsection{Partitioning the Ladder AMK constraint into blocks}

We define a block consisting of a sequence of $w$ variables $\{v_1, \ldots, v_w\}$ as a set of $w$ expressions \(\{\mathcal{E}_1, \ldots, \mathcal{E}_w\}\) where $\forall i \in [1, w]$, $\mathcal{E}_i = \sum_{j=1}^{i} v_j$. 
Figure~\ref{fig:block-definition} illustrates the structure of such a block.
A block can be efficiently represented by encoding the constraint $\mathcal{E}_w \leq k$ using a single \textit{Sequential Counter} (\textit{Seq}) encoding~\cite{sinz2005towards}, as \textit{Seq} encoding of $\mathcal{E}_w \leq k$ sequentially encodes $\mathcal{E}_1 \leq k$, \ldots, $ \mathcal{E}_{w-1} \leq k$.

\begin{figure}[ht!]
    \centering
    \begin{tabular}{cc|ccccccc}
    \cline{3-3}
    $\mathcal{E}_1$ & = & \multicolumn{1}{c|}{$v_1$} &  &  &  &  &  &  \\ \cline{4-5}
    $\mathcal{E}_2$ & = & $v_1$ & + & \multicolumn{1}{c|}{$v_2$} &  &  &  &  \\ \cline{6-7}
    $\ldots$ & = & \multicolumn{5}{c|}{$\ldots$} &  &  \\ \cline{8-9} 
    $\mathcal{E}_{w}$ & = & $v_1$ & + & $v_2$ & + & $\ldots$ & + & \multicolumn{1}{c|}{$v_w$} \\ \cline{3-9} 
    \end{tabular}
    \caption{A block consisting of a sequence of $w$ variables $\{v_1, \ldots, v_w\}$.}
    \label{fig:block-definition}
\end{figure}

Leverage this insight, we partition the input Ladder AMK constraint into blocks. Given a Ladder AMK constraint with width $w$ over a sequence of $n$ variables $\Omega = \{x_{1}, x_{2}, \ldots, x_{n}\}$, we first divide $\Omega$ into $M = \left\lceil \frac{n}{w} \right\rceil$ subsets $\{\omega_{1}, \ldots, \omega_{M}\}$. Each subset $\omega_{i}$ $(1 \leq i \leq M - 1)$ contains exactly $w$ unique consecutive variables s.t. $w_i = \{x_{i,1}, \ldots, x_{i,w}\}$, while the last subset $\omega_{M}$ may contain fewer variables if $n$ is not divisible by $w$. 
We then consider each subset $\omega_{i}$ as sequences of variables and construct blocks over these sequences so that the resulting blocks collectively cover all expressions required by the Ladder AMK constraint.

\begin{figure*}[h!]
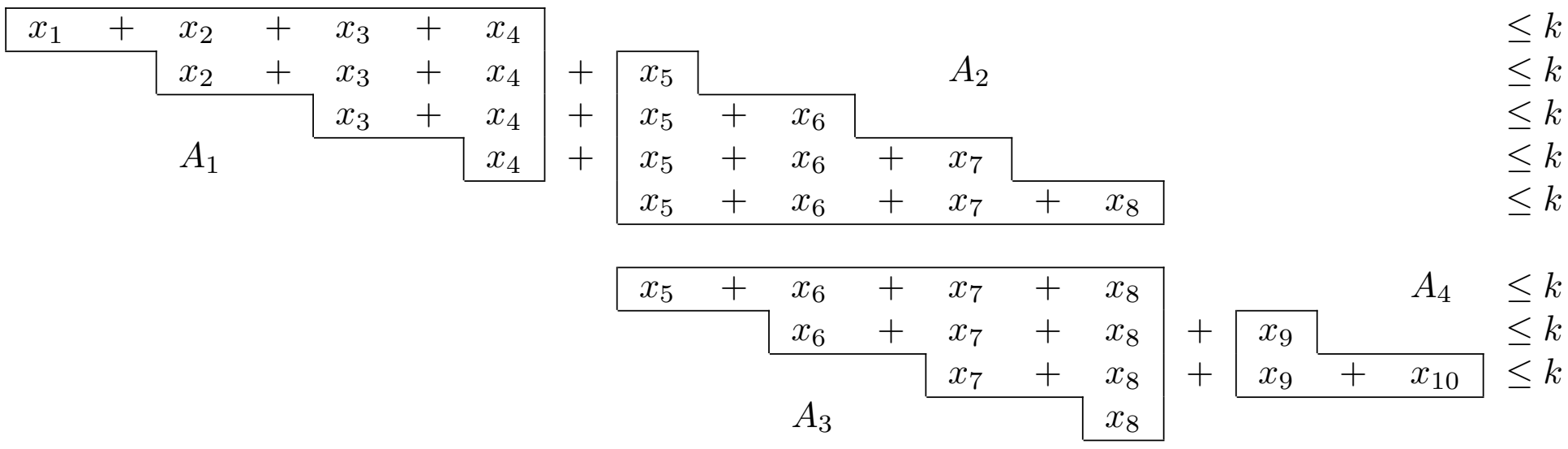

\centering
\resizebox{\textwidth}{!}{
    \begin{tabular}{llllllllllllllllllll}
    \cline{1-7}
    \multicolumn{1}{|l}{$x_1$} & + & $x_2$ & + & $x_3$ & + & \multicolumn{1}{l|}{$x_4$} &  &  &  &  &  &  &  &  &  &  &  &  & $\leq k$ \\ \cline{1-2} \cline{9-9}
     & \multicolumn{1}{l|}{} & $x_2$ & + & $x_3$ & + & \multicolumn{1}{l|}{$x_4$} & \multicolumn{1}{l|}{+} & \multicolumn{1}{l|}{$x_5$} &  &  &  & $A_2$ &  &  &  &  &  &  & $\leq k$ \\ \cline{3-4} \cline{10-11}
     &  &  & \multicolumn{1}{l|}{} & $x_3$ & + & \multicolumn{1}{l|}{$x_4$} & \multicolumn{1}{l|}{+} & $x_5$ & + & \multicolumn{1}{l|}{$x_6$} &  &  &  &  &  &  &  &  & $\leq k$ \\ \cline{5-6} \cline{12-13}
     &  & $A_1$ &  &  & \multicolumn{1}{l|}{} & \multicolumn{1}{l|}{$x_4$} & \multicolumn{1}{l|}{+} & $x_5$ & + & $x_6$ & + & \multicolumn{1}{l|}{$x_7$} &  &  &  &  &  &  & $\leq k$ \\ \cline{7-7} \cline{14-15}
     &  &  &  &  &  &  & \multicolumn{1}{l|}{} & $x_5$ & + & $x_6$ & + & $x_7$ & + & \multicolumn{1}{l|}{$x_8$} &  &  &  &  & $\leq k$ \\ \cline{9-15}
     &  &  &  &  &  &  &  &  &  &  &  &  &  &  &  &  &  &  &  \\ \cline{9-15}
     &  &  &  &  &  &  & \multicolumn{1}{l|}{} & $x_5$ & + & $x_6$ & + & $x_7$ & + & \multicolumn{1}{l|}{$x_8$} &  &  &  & $A_4$ & $\leq k$ \\ \cline{9-10} \cline{17-17}
     &  &  &  &  &  &  &  &  & \multicolumn{1}{l|}{} & $x_6$ & + & $x_7$ & + & \multicolumn{1}{l|}{$x_8$} & \multicolumn{1}{l|}{+} & \multicolumn{1}{l|}{$x_9$} &  &  & $\leq k$ \\ \cline{11-12} \cline{18-19}
     &  &  &  &  &  &  &  &  &  &  & \multicolumn{1}{l|}{} & $x_7$ & + & \multicolumn{1}{l|}{$x_8$} & \multicolumn{1}{l|}{+} & $x_9$ & + & \multicolumn{1}{l|}{$x_{10}$} & $\leq k$ \\ \cline{13-14} \cline{17-19}
     &  &  &  &  &  &  &  &  &  & $A_3$ &  &  & \multicolumn{1}{l|}{} & \multicolumn{1}{l|}{$x_8$} &  &  &  &  &  \\ \cline{15-15}
    \end{tabular}
}
\caption{Decomposition of an Ladder AMK constraint with $n = 10$ and $w = 4$.}
\label{fig:decompose_ladder_amk}
\end{figure*}

For instance, Figure \ref{fig:decompose_ladder_amk} illustrates the partitioning of the Ladder AMK constraint presented in Figure \ref{fig:ladder_amk_constraint_10_vars_4_width}. 
The sequence of variables in this constraint is divided into three subsets $\{\{x_1, \ldots, x_4\}, \{x_5, \ldots, x_8\}, \{x_9, x_{10}\}\}$. 
From these subsets, we construct blocks by considering appropriate variable sequences. 
The first subset (resp. the last subset) is considered as the sequence $\{x_4, \ldots, x_1\}$ (resp. $\{x_9, x_{10}\}$), from which we construct block $A_1$ (resp. block $A_4$). 
For the second subset, to cover all expressions required by the Ladder AMK constraint, we consider two sequences, namely $\{x_5, \ldots, x_8\}$ and $\{x_8, \ldots, x_5\}$, and construct two blocks $A_2$ and $A_3$, respectively. 
Since each block can be efficiently represented using a single \textit{Seq} encoding, the Ladder AMK constraint illustrated in Figure~\ref{fig:decompose_ladder_amk} can be encoded using four \textit{Seq} encodings, instead of the seven encodings that would be required if all seven AMK constraints were encoded independently.

\subsection{Encoding for Ladder AMK constraint}
\label{sec:encode-ladder-amk-constraint}

This section is organized into two parts.
First, we describe how each individual block obtained from the decomposition of the Ladder AMK constraint is encoded using \textit{SC} encoding.
Second, we connect adjacent blocks to reconstruct the original Ladder AMK constraint.

\vspace{0.25cm}
\begin{enumerate}[label=\textbf{\alph*.}]
\item \textbf{\textit{Encoding each block in the Ladder AMK constraint}}

\vspace{0.1cm}

\noindent We denote by $x_{i,j}$ the $j^\text{th}$ variable in the variable sequence of the block~$i$. 
For instance, consider the block $A_2$ illustrated in Figure~\ref{fig:decompose_ladder_amk}. 
The variable sequence of this block is $\{x_5, x_6, x_7, x_8\}$, which implies that $x_{2,1} \equiv x_5$, $x_{2,2} \equiv x_6$, $x_{2,3} \equiv x_7$, and $x_{2,4} \equiv x_8$.
Similarly, block $A_3$ in Figure~\ref{fig:decompose_ladder_amk} has variable sequence \{$x_8, x_7, x_6, x_5\}$.
Consequently, $x_{3,1} \equiv x_8$, $x_{3,2} \equiv x_7$, $x_{3,3} \equiv x_6$, and $x_{3,4} \equiv x_5$.

Since each $x_{i,j}$ is a binary variable, the sum $x_{i,1} + \ldots + x_{i,j}$ can take any integer value between $0$ and $j$. To represent this sum, we introduce $j$ register bits, denoted by $R_{i,j,s}$ for $s \in \{1, \ldots, j\}$. Each register bit $R_{i,j,s}$ indicates that under the condition ``the sum of the first $j$ variables in the variable sequence of block $i$ is greater than or equal to $s$''. Particularly, $R_{i,j,s}$ is \textit{true} if and only if $x_{i,1} + \ldots + x_{i,j} \geq s$, and \textit{false} otherwise, i.e., when $x_{i,1} + \ldots + x_{i,j} \leq s - 1$.

It is necessary to mention that all the expressions within each block are At-Most-K. Consequently, when encoding sub-expression $x_{i,1} + \ldots + x_{i,j}$ with $k < j$, it is sufficient to introduce only $k$ register bits $\{R_{i,j,1}, \ldots, R_{i,j,k}\}$. Any use of a register bit $R_{i,j,s}$ with $s > k$ is redundant, as such bits are always $false$. For example, to encode $x_{1,1} + x_{1,2} + x_{1,3} \leq 2$, only two register bits, namely \(R_{1,3,1}\) and \(R_{1,3,2}\), are required.

\begin{figure}[h!]
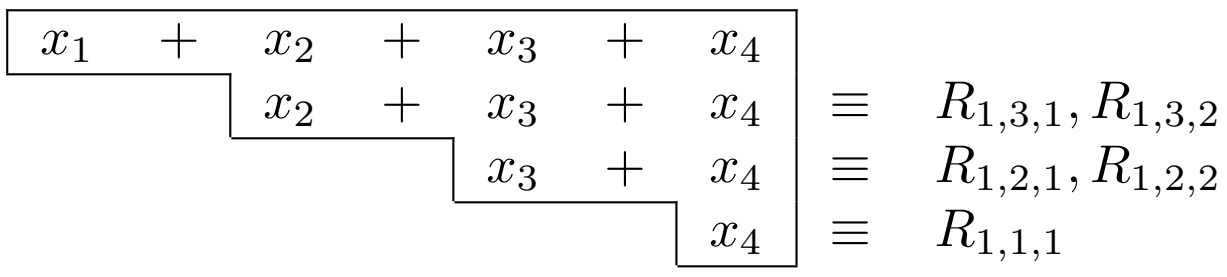

    \centering
    \begin{tabular}{lllllll|ll}
        \cline{1-7}
        \multicolumn{1}{|l}{$x_{1}$} & + & $x_{2}$ & + & $x_{3}$ & + & $x_{4}$ &  &  \\ \cline{1-2}
         & \multicolumn{1}{l|}{} & $x_{2}$ & + & $x_{3}$ & + & $x_{4}$ & $\equiv$ & $R_{1,3,1}, R_{1,3,2}$ \\ \cline{3-4}
         &  &  & \multicolumn{1}{l|}{} & $x_{3}$ & + & $x_{4}$ & $\equiv$ & $R_{1,2,1}, R_{1,2,2}$ \\ \cline{5-6}
         &  &  &  &  & \multicolumn{1}{l|}{} & $x_4$ & $\equiv$ & $R_{1,1,1}$ \\ \cline{7-7}
    \end{tabular}
    \caption{Register bit construction of block $A_1$ under the condition $\leq 2$.}
    \label{fig:register-bit-construction-of-block-x1-x2-x3-x4-under-the condition-leq-2}
\end{figure}

Figure \ref{fig:register-bit-construction-of-block-x1-x2-x3-x4-under-the condition-leq-2} illustrates the assignment of register bits for the block $A_1$ of the Ladder constraint illustrated in Figure \ref{fig:decompose_ladder_amk}, under the condition~$\leq 2$. $x_4$ has a maximum value of $1$, and thus requires only one register bit, denoted $R_{1,1,1}$. Next, $x_3 + x_4$ has a maximum value of $2$, and therefore is assigned two register bits, $R_{1,2,1}$ and $R_{1,2,2}$. For $x_2 + x_3 + x_4$, although the maximum value is $3$, due to the constraint $\leq 2$, this sub-expression is also assigned only two register bits, namely $R_{1,3,1}$ and $R_{1,3,2}$. Finally, $x_1 + x_2 + x_3 + x_4$ does not require additional register bits, since condition $\leq 2$ can be expressed based on the value of $x_1$ and the register bits already assigned to ${x_2 + x_3 + x_4}$.

We also note that, if $R_{i,j,s}$ is $true$, then $R_{i,j,s-1}$ must also be $true$, since $x_{i,1} + \ldots + x_{i,j} \geq s$ implies $x_{i,1} + \ldots + x_{i,j} \geq s-1$. Besides, the equality $x_{i,1} + \ldots + x_{i,j} = s$ is valid if and only if $R_{i,j,s}$ is $true$ and $R_{i,j,s+1}$ is $false$. Specifically, $R_{i,j,s}$ is $true$ indicating that $x_{i,1} + \ldots + x_{i,j} \geq s$, while $R_{i,j,s+1}$ is $false$ implying that $x_{i,1} + \ldots + x_{i,j} \leq (s+1) - 1 = s$. Combining these two conditions results in $s \leq x_{i,1} + \ldots + x_{i,j} \leq s$, and thus $x_{i,1} + \ldots + x_{i,j} = s$.

The \textit{SC} encoding for decomposed blocks involves the application of seven equations \eqref{eq:ladder-amk-init-bit}, \eqref{eq:ladder-amk-copy-bit}, \eqref{eq:ladder-amk-derive-true-bit}, \eqref{eq:ladder-amk-init-false-bit}, \eqref{eq:ladder-amk-copy-false-bit}, \eqref{eq:ladder-amk-derive-false-bit}, and \eqref{eq:ladder-amk-restrict-bit}. The formal proof of the correctness of this encoding is given in Appendix~\ref{app:sc-correctness}.

\begin{equation}
\label{eq:ladder-amk-init-bit}
    \bigwedge_{j = 1}^{w - 1} x_{i,j} \to R_{i, j, 1}
\end{equation}

\begin{equation}
\label{eq:ladder-amk-copy-bit}
    \bigwedge_{j = 2}^{w - 1} \bigwedge_{s = 1}^{min(j-1, k)} R_{i, j - 1, s} \to R_{i,j, s}
\end{equation}

\begin{equation}
\label{eq:ladder-amk-derive-true-bit}
    \bigwedge_{j = 2}^{w - 1} \bigwedge_{s = 2}^{min(j, k)} x_{i,j} \wedge R_{i, j-1, s - 1} \to R_{i,j,s}
\end{equation}

\begin{equation}
\label{eq:ladder-amk-init-false-bit}
    \bigwedge_{j = 1}^{k} \neg x_{i,j} \to \neg R_{i,j,j}
\end{equation}

\begin{equation}
\label{eq:ladder-amk-copy-false-bit}
    \bigwedge_{j = 2}^{w - 1} \bigwedge_{s = 2}^{min(j, k)} \neg R_{i, j-1, s-1} \to \neg R_{i,j,s}
\end{equation}

\begin{equation}
\label{eq:ladder-amk-derive-false-bit}
    \bigwedge_{j = 2}^{w - 1} \bigwedge_{s = 1}^{min(j-1, k)} \neg x_{i,j} \wedge \neg R_{i, j-1, s} \to \neg R_{i,j,s}
\end{equation}

\begin{equation}
\label{eq:ladder-amk-restrict-bit}
    \bigwedge_{j=k+1}^w x_{i,j} \to \neg R_{i, j - 1, k}  
\end{equation}

Equation \eqref{eq:ladder-amk-init-bit} specifies that the sum of the first $j$ variables in block $i$ must be at least one if $x_{i,j}$ is $true$.
Equation \eqref{eq:ladder-amk-copy-bit} states that if the sum of the first $j-1$ variables in block $i$ is greater than or equal to $s$, then the sum of the first $j$ variables must also be at least $s$.
Equation \eqref{eq:ladder-amk-derive-true-bit} ensures that if $x_{i,j}$ is $true$ and the sum of the first $j-1$ variables is greater than or equal to $s - 1$, then the sum of the first $j$ variables in block $i$ must be at least $s$.
Together, equations \eqref{eq:ladder-amk-init-bit}, \eqref{eq:ladder-amk-copy-bit}, and \eqref{eq:ladder-amk-derive-true-bit} ensure the correctness of the statement ``$R_{i,j,s}$ holds $true$ if and only if $x_{i,1} + \ldots + x_{i,j} \geq s$''.

Next, equation \eqref{eq:ladder-amk-init-false-bit} limits the sum of the first $j$ variables in block $i$ to at most $j - 1$ if $x_{i,j}$ is $false$.
The equation \eqref{eq:ladder-amk-copy-false-bit} states that if the sum of the first $j-1$ variables is less than $s - 1$, then the sum of the first $j$ variables must be less than $s$.
Equation \eqref{eq:ladder-amk-derive-false-bit} states that if $x_{i,j}$ is $false$ and the sum of the first $j-1$ variables is less than $s$, then the sum of the first $j$ variables must also be less than $s$. 
Taken together, the equations \eqref{eq:ladder-amk-init-false-bit}, \eqref{eq:ladder-amk-copy-false-bit}, and \eqref{eq:ladder-amk-derive-false-bit} validate the correctness of the statement ``$R_{i,j,s}$ is $false$ if and only if $x_{i,1} + \ldots + x_{i,j} \leq s - 1$''.

Finally, the equation \eqref{eq:ladder-amk-restrict-bit} stipulates that if \( x_{i,j} \) is \textit{true}, then the sum of the first \( j-1 \) variables in block \( i \) must be strictly less than \( k \). 
By applying this constraint to all indices \( j \) within the range of \( [k+1, w] \), we ensure that no more than \( k \) out of the \( w \) variables can be set to \textit{true}. 
Additionally, the sizes of the clauses (i.e., the number of literals in each clause) in our encoding are at most 3.

As previously mentioned, the \textit{SC} encoding is inspired by the \textit{Seq} encoding \cite{sinz2005towards}. In particular, equations~\eqref{eq:ladder-amk-init-bit}, \eqref{eq:ladder-amk-copy-bit}, \eqref{eq:ladder-amk-derive-true-bit}, and \eqref{eq:ladder-amk-restrict-bit} are adopted directly from the \textit{Seq} encoding \cite{sinz2005towards}. The remaining equations~\eqref{eq:ladder-amk-init-false-bit}, \eqref{eq:ladder-amk-copy-false-bit}, and \eqref{eq:ladder-amk-derive-false-bit} are intentionally included to explicitly characterize the conditions under which the corresponding register bits are forced to \textit{false}. This is crucial for ensuring the correctness of the proposed encoding, as these register bits will be reused to represent the block connections.

It is necessary to note that when two blocks involve the same subset of variables $\omega_i = \{x_{i,1}, \ldots, x_{i,w}\}$, if one block has already enforced the constraint $x_{i,1} + \ldots + x_{i,w} \leq k$ via equation~\eqref{eq:ladder-amk-restrict-bit}, the application of equation~\eqref{eq:ladder-amk-restrict-bit} in the other block is unnecessary.
Exploiting this insight, when encoding two different blocks that share the same subset of variables, we include equation~\eqref{eq:ladder-amk-restrict-bit} on only one of them.
We refer to the block on which equation~\eqref{eq:ladder-amk-restrict-bit} is applied to enforce the AMK constraint as the AMK block, while the other block, which primarily provides expressions for block connection, is referred to as the connection block.

\begin{figure*}[h!]
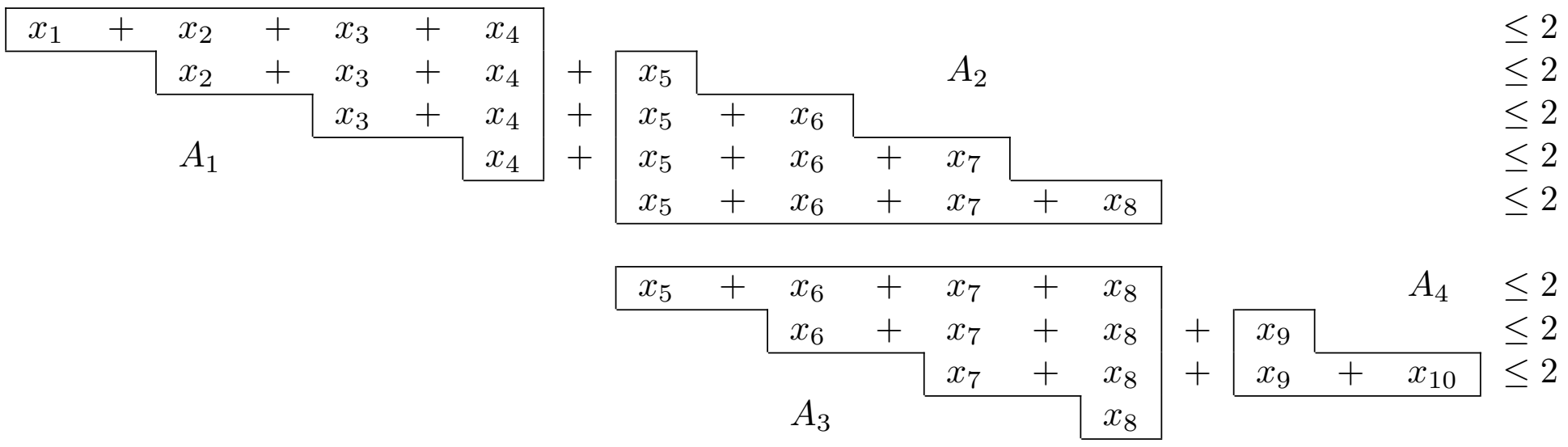

\centering
\resizebox{\textwidth}{!}{
    \begin{tabular}{llllllllllllllllllll}
    \cline{1-7}
    \multicolumn{1}{|l}{$x_1$} & + & $x_2$ & + & $x_3$ & + & \multicolumn{1}{l|}{$x_4$} &  &  &  &  &  &  &  &  &  &  &  &  & $\leq 2$ \\ \cline{1-2} \cline{9-9}
     & \multicolumn{1}{l|}{} & $x_2$ & + & $x_3$ & + & \multicolumn{1}{l|}{$x_4$} & \multicolumn{1}{l|}{+} & \multicolumn{1}{l|}{$x_5$} &  &  &  & $A_2$ &  &  &  &  &  &  & $\leq 2$ \\ \cline{3-4} \cline{10-11}
     &  &  & \multicolumn{1}{l|}{} & $x_3$ & + & \multicolumn{1}{l|}{$x_4$} & \multicolumn{1}{l|}{+} & $x_5$ & + & \multicolumn{1}{l|}{$x_6$} &  &  &  &  &  &  &  &  & $\leq 2$ \\ \cline{5-6} \cline{12-13}
     &  & $A_1$ &  &  & \multicolumn{1}{l|}{} & \multicolumn{1}{l|}{$x_4$} & \multicolumn{1}{l|}{+} & $x_5$ & + & $x_6$ & + & \multicolumn{1}{l|}{$x_7$} &  &  &  &  &  &  & $\leq 2$ \\ \cline{7-7} \cline{14-15}
     &  &  &  &  &  &  & \multicolumn{1}{l|}{} & $x_5$ & + & $x_6$ & + & $x_7$ & + & \multicolumn{1}{l|}{$x_8$} &  &  &  &  & $\leq 2$ \\ \cline{9-15}
     &  &  &  &  &  &  &  &  &  &  &  &  &  &  &  &  &  &  &  \\ \cline{9-15}
     &  &  &  &  &  &  & \multicolumn{1}{l|}{} & $x_5$ & + & $x_6$ & + & $x_7$ & + & \multicolumn{1}{l|}{$x_8$} &  &  &  & $A_4$ & $\leq 2$ \\ \cline{9-10} \cline{17-17}
     &  &  &  &  &  &  &  &  & \multicolumn{1}{l|}{} & $x_6$ & + & $x_7$ & + & \multicolumn{1}{l|}{$x_8$} & \multicolumn{1}{l|}{+} & \multicolumn{1}{l|}{$x_9$} &  &  & $\leq 2$ \\ \cline{11-12} \cline{18-19}
     &  &  &  &  &  &  &  &  &  &  & \multicolumn{1}{l|}{} & $x_7$ & + & \multicolumn{1}{l|}{$x_8$} & \multicolumn{1}{l|}{+} & $x_9$ & + & \multicolumn{1}{l|}{$x_{10}$} & $\leq 2$ \\ \cline{13-14} \cline{17-19}
     &  &  &  &  &  &  &  &  &  & $A_3$ &  &  & \multicolumn{1}{l|}{} & \multicolumn{1}{l|}{$x_8$} &  &  &  &  &  \\ \cline{15-15}
    \end{tabular}
}
\caption{Decomposition of Ladder At-Most-Two constraint.}
\label{fig:decompose_ladder_at_most_two}
\end{figure*}

\begin{figure*}[h!]
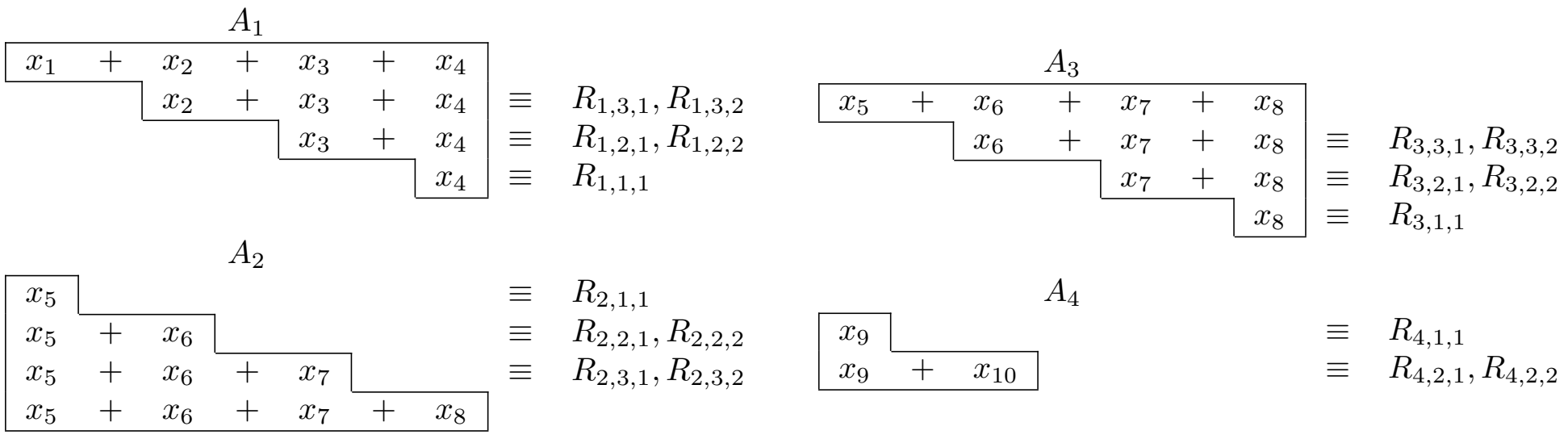

    \centering
    \begin{minipage}{0.49\textwidth}
        \centering
        \resizebox{\textwidth}{!}{
            \begin{tabular}{lllllllll}
            \multicolumn{7}{c}{$A_{1}$} &  &  \\ \cline{1-7}
            \multicolumn{1}{|l}{$x_{1}$} & + & $x_{2}$ & + & $x_{3}$ & + & \multicolumn{1}{l|}{$x_{4}$} & & \\ \cline{1-2}
             & \multicolumn{1}{l|}{} & $x_{2}$ & + & $x_{3}$ & + & \multicolumn{1}{l|}{$x_{4}$} & $\equiv$ & $R_{1,3,1}, R_{1,3,2}$ \\ \cline{3-4}
             &  &  & \multicolumn{1}{l|}{} & $x_{3}$ & + & \multicolumn{1}{l|}{$x_{4}$} & $\equiv$ & $R_{1,2,1}, R_{1,2,2}$ \\ \cline{5-6}
             &  &  &  &  & \multicolumn{1}{l|}{} & \multicolumn{1}{l|}{$x_4$} & $\equiv$ & $R_{1,1,1}$ \\ \cline{7-7}
             &  &  &  &  &  &  &  &  \\
            \multicolumn{7}{c}{$A_{2}$} &  &  \\ \cline{1-1}
            \multicolumn{1}{|l|}{$x_5$} &  &  &  &  &  &  & $\equiv$ & $R_{2,1,1}$ \\ \cline{2-3}
            \multicolumn{1}{|l}{$x_{5}$} & + & \multicolumn{1}{l|}{$x_{6}$} &  &  &  &  & $\equiv$ & $R_{2,2,1}, R_{2,2,2}$ \\ \cline{4-5}
            \multicolumn{1}{|l}{$x_{5}$} & + & $x_{6}$ & + & \multicolumn{1}{l|}{$x_{7}$} &  &  & $\equiv$ & $R_{2,3,1}, R_{2,3,2}$ \\ \cline{6-7}
            \multicolumn{1}{|l}{$x_{5}$} & + & $x_{6}$ & + & $x_{7}$ & + & \multicolumn{1}{l|}{$x_{8}$} & & \\ \cline{1-7}
            &  &  &  &  &  &  &  &  \\
            \end{tabular}
        }
    \end{minipage}%
    \hfill
    \begin{minipage}{0.49\textwidth}
        \centering
        \resizebox{\textwidth}{!}{
            \begin{tabular}{lllllllll}
            \multicolumn{7}{c}{$A_{3}$} &  &  \\ \cline{1-7}
            \multicolumn{1}{|l}{$x_{5}$} & + & $x_{6}$ & + & $x_{7}$ & + & \multicolumn{1}{l|}{$x_{8}$} & & \\ \cline{1-2}
             & \multicolumn{1}{l|}{} & $x_{6}$ & + & $x_{7}$ & + & \multicolumn{1}{l|}{$x_{8}$} & $\equiv$ & $R_{3,3,1}, R_{3,3,2}$ \\ \cline{3-4}
             &  &  & \multicolumn{1}{l|}{} & $x_{7}$ & + & \multicolumn{1}{l|}{$x_{8}$} & $\equiv$ & $R_{3,2,1}, R_{3,2,2}$ \\ \cline{5-6}
             &  &  &  &  & \multicolumn{1}{l|}{} & \multicolumn{1}{l|}{$x_8$} & $\equiv$ & $R_{3,1,1}$ \\ \cline{7-7}
             &  &  &  &  &  &  &  &  \\
            \multicolumn{7}{c}{$A_{4}$} &  &  \\ \cline{1-1}
            \multicolumn{1}{|l|}{$x_9$} &  &  &  &  &  &  & $\equiv$ &  $R_{4,1,1}$ \\ \cline{2-3}
            \multicolumn{1}{|l}{$x_9$} & + & \multicolumn{1}{l|}{$x_{10}$} &  &  &  &  & $\equiv$ & $R_{4,2,1}, R_{4,2,2}$ \\ \cline{1-3}
            &  &  &  &  &  &  &  &  \\
            \end{tabular}
        }
    \end{minipage}
    \caption{Register bits constructing of Ladder At-Most-Two constraint.}
    \label{fig:register_bit_constructing_ladder_at_most_two}
\end{figure*}

Figure \ref{fig:decompose_ladder_at_most_two} illustrates an example of a Ladder AMK constraint with \( k=2 \). This Ladder AMK constraint is divided into four blocks, labeled \( A_1, A_2, A_3, \) and \( A_4 \). Figure \ref{fig:register_bit_constructing_ladder_at_most_two} presents these four blocks along with the necessary register bits used in the \textit{SC} encoding.
The block \( A_1 \) is encoded using \textit{SC} with the formulas from equations \eqref{eq:ladder-amk-init-bit} to \eqref{eq:ladder-amk-restrict-bit}. The generated clauses for constructing the block \( A_1 \) are presented below. In this context, the variables are defined as \( x_{1,1} \equiv x_4 \), \( x_{1,2} \equiv x_3 \), \( x_{1,3} \equiv x_2 \), \( x_{1,4} \equiv x_1 \), \( x_{1,1} \equiv R_{1,1,1} \), and \( x_4 \equiv R_{1,1,1} \).

\vspace{0.25cm}

\resizebox{0.8\textwidth}{!}{
\begin{tabular}{ccccccccc}
\multirow{2}{*}{(1)} & \multirow{2}{*}{$\overset{A_1}{\iff}$} & $x_{1,2}$ & $\to$ & $R_{1,2,1}$ & \multirow{2}{*}{$\iff$} & $x_3$ & $\to$ & $R_{1,2,1}$ \\
&  & $x_{1,3}$ & $\to$ & $R_{1,3,1}$ &  & $x_2$ & $\to$ & $R_{1,3,1}$ \\
&  &  &  &  &  &  &  &  \\
\multirow{3}{*}{(2)} & \multirow{3}{*}{$\overset{A_1}{\iff}$} & $R_{1,1,1}$ & $\to$ & $R_{1,2,1}$ & \multirow{3}{*}{$\iff$} & $x_4$ & $\to$ & $R_{1,2,1}$ \\
&  & $R_{1,2,1}$ & $\to$ & $R_{1,3,1}$ &  & $R_{1,2,1}$ & $\to$ & $R_{1,3,1}$ \\
&  & $R_{1,2,2}$ & $\to$ & $R_{1,3,2}$ &  & $R_{1,2,2}$ & $\to$ & $R_{1,3,2}$ \\
&  &  &  &  &  &  &  &  \\
\multirow{2}{*}{(3)} & \multirow{2}{*}{$\overset{A_1}{\iff}$} & $x_{1,2} \wedge R_{1,1,1}$ & $\to$ & $R_{1,2,2}$ & \multirow{2}{*}{$\iff$} & $x_3 \wedge x_4$ & $\to$ & $R_{1,2,2}$ \\
&  & $x_{1.3} \wedge R_{1,2,1}$ & $\to$ & $R_{1,3,2}$ &  & $x_2 \wedge R_{1,2,1}$ & $\to$ & $R_{1,3,2}$ \\
&  &  &  &  &  &  &  & \\
(4) & $\overset{A_1}{\iff}$ & $\neg x_{1,2}$ & $\to$ & $\neg R_{1,2,2}$ & $\iff$ & $\neg x_3$ & $\to$ & $\neg R_{1,2,2}$ \\
&  &  &  &  &  &  &  &  \\
\multirow{2}{*}{(5)} & \multirow{2}{*}{$\overset{A_1}{\iff}$} & $\neg R_{1,1,1}$ & $\to$ & $\neg R_{1,2,2}$ & \multirow{2}{*}{$\iff$} & $\neg x_4$ & $\to$ & $\neg R_{1,2,2}$ \\
&  & $\neg R_{1,2,1}$ & $\to$ & $\neg R_{1,3,2}$ &  & $\neg R_{1,2,1}$ & $\to$ & $\neg R_{1,3,2}$ \\
&  &  &  &  &  &  &  &  \\
\multirow{3}{*}{(6)} & \multirow{3}{*}{$\overset{A_1}{\iff}$} & $\neg x_{1,2} \wedge \neg R_{1,1,1}$ & $\to$ & $\neg R_{1,2,1}$ & \multirow{3}{*}{$\iff$} & $\neg x_3 \wedge \neg x_4$ & $\to$ & $\neg R_{1,2,1}$ \\
&  & $\neg x_{1,3} \wedge \neg R_{1,2,1}$ & $\to$ & $\neg R_{1,3,1}$ &  & $\neg x_2 \wedge \neg R_{1,2,1}$ & $\to$ & $\neg R_{1,3,1}$ \\
&  & $\neg x_{1,3} \wedge \neg R_{1,2,2}$ & $\to$ & $\neg R_{1,3,2}$ &  & $\neg x_2 \wedge \neg R_{1,2,2}$ & $\to$ & $\neg R_{1,3,2}$ \\
&  &  &  &  &  &  &  &  \\
\multirow{2}{*}{(7)} & \multirow{2}{*}{$\overset{A_1}{\iff}$} & $x_{1,3}$ & $\to$ & $\neg R_{1,2,2}$ & \multirow{2}{*}{$\iff$} & $x_2$ & $\to$ & $\neg R_{1,2,2}$ \\
&  & $x_{1,4}$ & $\to$ & $\neg R_{1,3,2}$ &  & $x_1$ & $\to$ & $\neg R_{1,3,2}$ \\
&  &  &  &  &  &  &  &  \\
\end{tabular}
}


Encoding the block $A_2$ is the same as $A_1$, including the following clauses: 


\resizebox{0.8\textwidth}{!}{
\begin{tabular}{ccccccccc}
\multirow{2}{*}{(1)} & \multirow{2}{*}{$\overset{A_2}{\iff}$} & $x_{2,2}$ & $\to$ & $R_{2,2,1}$ & \multirow{2}{*}{$\iff$} & $x_6$ & $\to$ & $R_{2,2,1}$ \\
 &  & $x_{2,3}$ & $\to$ & $R_{2,3,1}$ &  & $x_7$ & $\to$ & $R_{2,3,1}$ \\
 &  &  &  &  &  &  &  &  \\
\multirow{3}{*}{(2)} & \multirow{3}{*}{$\overset{A_2}{\iff}$} & $R_{2,1,1}$ & $\to$ & $R_{2,2,1}$ & \multirow{3}{*}{$\iff$} & $x_5$ & $\to$ & $R_{2,2,1}$ \\
 &  & $R_{2,2,1}$ & $\to$ & $R_{2,3,1}$ &  & $R_{2,2,1}$ & $\to$ & $R_{2,3,1}$ \\
 &  & $R_{2,2,2}$ & $\to$ & $R_{2,3,2}$ &  & $R_{2,2,2}$ & $\to$ & $R_{2,3,2}$ \\
 &  &  &  &  &  &  &  &  \\
\multirow{2}{*}{(3)} & \multirow{2}{*}{$\overset{A_2}{\iff}$} & $x_{2,2} \wedge R_{2,1,1}$ & $\to$ & $R_{2,2,2}$ & \multirow{2}{*}{$\iff$} & $x_6 \wedge x_5$ & $\to$ & $R_{2,2,2}$ \\
 &  & $x_{2.3} \wedge R_{2,2,1}$ & $\to$ & $R_{2,3,2}$ &  & $x_7 \wedge R_{2,2,1}$ & $\to$ & $R_{2,3,2}$ \\
 &  &  &  &  &  &  &  &  \\
(4) & $\overset{A_2}{\iff}$ & $\neg x_{2,2}$ & $\to$ & $\neg R_{2,2,2}$ & $\iff$ & $\neg x_6$ & $\to$ & $\neg R_{2,2,2}$ \\
\end{tabular}
}

\resizebox{0.8\textwidth}{!}{
\begin{tabular}{ccccccccc}
\multirow{2}{*}{(5)} & \multirow{2}{*}{$\overset{A_2}{\iff}$} & $\neg R_{2,1,1}$ & $\to$ & $\neg R_{2,2,2}$ & \multirow{2}{*}{$\iff$} & $\neg x_5$ & $\to$ & $\neg R_{2,2,2}$ \\
 &  & $\neg R_{2,2,1}$ & $\to$ & $\neg R_{2,3,2}$ &  & $\neg R_{2,2,1}$ & $\to$ & $\neg R_{2,3,2}$ \\
 &  &  &  &  &  &  &  &  \\
\multirow{3}{*}{(6)} & \multirow{3}{*}{$\overset{A_2}{\iff}$} & $\neg x_{2,2} \wedge \neg R_{2,1,1}$ & $\to$ & $\neg R_{2,2,1}$ & \multirow{3}{*}{$\iff$} & $\neg x_6 \wedge \neg x_5$ & $\to$ & $\neg R_{2,2,1}$ \\
 &  & $\neg x_{2,3} \wedge \neg R_{2,2,1}$ & $\to$ & $\neg R_{2,3,1}$ &  & $\neg x_7 \wedge \neg R_{2,2,1}$ & $\to$ & $\neg R_{2,3,1}$ \\
 &  & $\neg x_{2,3} \wedge \neg R_{2,2,2}$ & $\to$ & $\neg R_{2,3,2}$ &  & $\neg x_7 \wedge \neg R_{2,2,2}$ & $\to$ & $\neg R_{2,3,2}$ \\
 &  &  &  &  &  &  &  &  \\
\multirow{2}{*}{(7)} & \multirow{2}{*}{$\overset{A_2}{\iff}$} & $x_{2,3}$ & $\to$ & $\neg R_{2,2,2}$ & \multirow{2}{*}{$\iff$} & $x_7$ & $\to$ & $\neg R_{2,2,2}$ \\
 &  & $x_{2,4}$ & $\to$ & $\neg R_{2,3,2}$ &  & $x_8$ & $\to$ & $\neg R_{2,3,2}$ \\
\end{tabular}
}

\vspace{0.25cm}

Since the block \( A_2 \) has already enforced the constraint \( x_5 + x_6 + x_7 + x_8 \leq 2 \), the block \( A_3 \) only leverages the formulas from equations \eqref{eq:ladder-amk-init-bit} to \eqref{eq:ladder-amk-derive-false-bit} to generate the necessary register bits for its connection to the adjacent block \( A_4 \). 
Finally, the block \(A_4\) is a special case, as it contains only two variables instead of four as in the other blocks. Consequently, it is encoded using seven equations~\eqref{eq:ladder-amk-init-bit} to \eqref{eq:ladder-amk-restrict-bit}, with \(w\) set to \(2\). As a result, the encoding of block $A_4$ consists of the following clauses: 
\[(x_9~\to~R_{4,2,1})~\wedge~(x_{10}~\to~R_{4,2,1})~\wedge~((x_9~\wedge~x_{10})~\to~R_{4,2,2})~\wedge~(\neg~x_9~\to~\neg~R_{4,2,2})\]
\[\wedge(\neg~x_{10}~\to~\neg~R_{4,2,2})~\wedge~((\neg~x_9~\wedge~\neg~x_{10})~\to~\neg~R_{4,2,1}).\]

\vspace{0.25cm}
\item \textbf{\textit{Clauses connecting two adjacent blocks}}
\vspace{0.1cm}

\noindent After encoding blocks, we construct connecting clauses to connect adjacent blocks, thereby reformulate the Ladder AMK constraint. This construction is supported by Proposition~\ref{prop:decompose_AMK_constraint}. For instance, connection between blocks $A_1$ and $A_2$ in Figure~\ref{fig:decompose_ladder_at_most_two} is formed as follows.

\begin{center}
    \begin{tabular}{ccc}
        \multirow{3}{*}{$(x_{2} + x_{3} + x_{4} + x_{5}\leq 2)$} & \multirow{3}{*}{$\overset{\text{Prop. \ref{prop:decompose_AMK_constraint}}}{\equiv}$} & $(x_{2} + x_{3} + x_{4} \leq 2) \wedge (x_{5} \leq 2)$\\
        &  & $\wedge (x_{2} + x_{3} + x_{4} \leq 1 \vee x_{5} \leq 0)$ \rule{0pt}{12pt} \\
        &  & $\wedge (x_{2} + x_{3} + x_{4} \leq 0 \vee x_{5} \leq 1)$ \rule{0pt}{12pt} \\
    \end{tabular}
\end{center}

\vspace{0.1cm}
\begin{center}
    \begin{tabular}{ccc}
    
        \multirow{3}{*}{$(x_{3} + x_{4} + x_{5} + x_{6}\leq 2)$} & \multirow{3}{*}{$\overset{\text{Prop. \ref{prop:decompose_AMK_constraint}}}{\equiv}$} & $(x_{3} + x_{4} \leq 2) \wedge (x_{5} + x_{6} \leq 2)$ \\
         &  & $\wedge (x_{3} + x_{4} \leq 1 \vee x_{5} + x_{6} \leq 0)$ \rule{0pt}{12pt}\\
         &  & $\wedge (x_{3} + x_{4} \leq 0 \vee x_{5} + x_{6} \leq 1)$ \rule{0pt}{12pt}\\
    \end{tabular}
\end{center}

\vspace{0.1cm}
\begin{center}
    \begin{tabular}{ccc}
         \multirow{3}{*}{$(x_{4} + x_{5} + x_{6} + x_{7} \leq 2)$} & \multirow{3}{*}{$\overset{\text{Prop. \ref{prop:decompose_AMK_constraint}}}{\equiv}$} & $(x_{4} \leq 2) \wedge (x_{5} + x_{6} + x_{7} \leq 2)$ \rule{0pt}{12pt}\\
         &  & $\wedge (x_{4} \leq 1 \vee x_{5} + x_{6} + x_{7} \leq 0)\rule{0pt}{12pt}$ \\
         &  & $\wedge (x_{4} \leq 0 \vee x_{5} + x_{6} + x_{7} \leq 1)\rule{0pt}{12pt}$ \\
    \end{tabular}
\end{center}
\vspace{0.5cm}

Since the constraint \(x_1 + x_2 + x_3 + x_4 \leq 2\) (resp. \(x_5 + x_6 + x_7 + x_8 \leq 2\)) is \textit{true}, as guaranteed by the equation~\eqref{eq:ladder-amk-restrict-bit}, it follows that all its sub-constraints are also \textit{true}, namely \(x_2 + x_3 + x_4 \leq 2\), \(x_3 + x_4 \leq 2\), and \(x_4 \leq 2\) (resp. \(x_5 + x_6 + x_7 \leq 2\), \(x_5 + x_6 \leq 2\), and \(x_5 \leq 2\)). Consequently, the connection between block \( A_1 \) and block \( A_2 \) can be represented as follows.
\vspace{-0.3cm}
\begin{align*}
    & (x_{2} + x_{3} + x_{4} \leq 1 \vee x_{5} \leq 0) \wedge (x_{2} + x_{3} + x_{4} \leq 0 \vee x_{5} \leq 1) \\
    \wedge & (x_{3} + x_{4} \leq 1 \vee x_{5} + x_{6} \leq 0) \wedge (x_{3} + x_{4} \leq 0 \vee x_{5} + x_{6} \leq 1) \\
    \wedge & (x_{4} \leq 1 \vee x_{5} + x_{6} + x_{7} \leq 0) \wedge (x_{4} \leq 0 \vee x_{5} + x_{6} + x_{7} \leq 1)
\end{align*}

We utilize the generated register bits to express the connection between \( A_1 \) and \( A_2 \). For instance, the expression \( x_2 + x_3 + x_4 \leq 1 \) serves as the negation of \( x_2 + x_3 + x_4 \geq 2 \) and can be replaced with \( \neg R_{1,3,2} \). Similarly, the expression \( x_2 + x_3 + x_4 \leq 0 \) is the negation of \( x_2 + x_3 + x_4 \geq 1 \) and can be represented as \( \neg R_{1,3,1} \). This substitution can be applied to the remaining expressions as well. Therefore, the connection between block \( A_1 \) and block \( A_2 \) can be expressed more straightforwardly as follows.

\vspace{-0.5cm}
\begin{align*}
    (\neg R_{1,3,2} \vee x_{5} \leq 0) &\wedge (\neg R_{1,3,1} \vee x_{5} \leq 1) \\
    \wedge (\neg R_{1,2,2} \vee \neg R_{2,2,1}) &\wedge (\neg R_{1,2,1} \vee \neg R_{2,2,2}) \\
    \wedge (x_{4} \leq 1 \vee \neg R_{2,3,1}) &\wedge (x_{4} \leq 0 \vee \neg R_{2,3,2})
\end{align*}

In summary, connecting two consecutive blocks \(i\) and \(i+1\) in the Ladder AMK constraint involves a series of disjunctions, as formally presented in equation \eqref{eq:connect_amk_blocks}. This equation can also be optimized by reusing register bits from the \textit{SC} encoding of the individual blocks, as presented in the example above.

\vspace{-0.5cm}

\begin{align}
\label{eq:connect_amk_blocks}
    \bigwedge_{j = 2}^{w}\bigwedge_{p =1}^k(\sum_{q=1}^{w-j+1}x_{i,q}\leq k - p \vee \sum_{q'=1}^{j-1}x_{i+1, q'}\leq p -1)
\end{align}

\end{enumerate}

\subsection{Comparison of Ladder AMK constraint encodings}
\label{sec:compare-ladder-amk}

In this section, we assess the performance of the \textit{SCL} encoding on the Ladder AMK constraint (\(SCL_{\leq k}\)) and compare it with other encodings that encode each AMK constraint individually. These encodings include \textit{Pairwise}, \textit{Sequential Counter} (\textit{Seq}) \cite{sinz2005towards}, \textit{Binary Decision Diagrams} (\textit{BDD}) \cite{abio2012new}, \textit{Cardinality Network} (\textit{Card}) \cite{asin2009cardinality, asin2011cardinality}, and \textit{Adder Network} (\textit{Adder}) \cite{een2006translating}.

Table \ref{tab:info_ladder_amk} summarizes the evaluation of different encodings used to represent the Ladder AMK constraints over \( n \) variables with width \( w \). The encodings are compared based on the number of required auxiliary variables, the number of generated clauses, and the complexity of the clause space in the worst case, specifically when \( w \) is approximately \( \frac{n}{2} \).

\begin{table*}[ht!]
\centering
\caption{Size of SAT encoding for a Ladder AMK constraint over $n$ variables with width $w$, $N = (n-w)+1$ and \( M = \left\lceil \frac{n}{w} \right\rceil \).}
\resizebox{\textwidth}{!}{
\begin{tabular}{|c|c|c|c|c|}
\hline
\multirow{2}{*}{Encoding} & \multirow{2}{*}{Technique} & \multirow{2}{*}{Auxiliary variables} & \multirow{2}{*}{Clauses} & \multirow{2}{*}{Complexity} \\
 &  &  &  &  \\ \hline
\multirow{2}{*}{\textit{Pairwise}} & \multirow{10}{*}{\begin{tabular}[c]{@{}c@{}}Encode each \\ AMK constraint \\ separately\end{tabular}} & \multirow{2}{*}{0} & \multirow{2}{*}{$C^w_{k+1} + (N-1)C^{w-1}_{k}$} & \multirow{2}{*}{$\mathcal{O}(n^{k+1})$} \\
 &  &  &  &  \\ \cline{1-1} \cline{3-5} 
\multirow{2}{*}{\textit{Seq}} &  & \multirow{2}{*}{$Nk(w-2)$} & \multirow{2}{*}{$N(3kw -7k + w)$} & \multirow{2}{*}{$\mathcal{O}(n^2k)$} \\
 &  &  &  &  \\ \cline{1-1} \cline{3-5} 
\multirow{2}{*}{\textit{BDD}} &  & \multirow{2}{*}{$N(w - k)(k + 1)$} & \multirow{2}{*}{$2N(w - k)(k + 1)$} & \multirow{2}{*}{$\mathcal{O}(n^2k)$} \\
 &  &  &  &  \\ \cline{1-1} \cline{3-5} 
\multirow{2}{*}{\textit{Card}} &  & \multirow{2}{*}{$\mathcal{O}(n^2\text{log}_2^2k)$} & \multirow{2}{*}{$\mathcal{O}(n^2\text{log}_2^2k)$} & \multirow{2}{*}{$\mathcal{O}(n^2\text{log}_2^2k)$} \\
 &  &  &  &  \\ \cline{1-1} \cline{3-5}
 \multirow{2}{*}{\textit{Adder}} &  & \multirow{2}{*}{$\mathcal{O}(n^2)$} & \multirow{2}{*}{$\mathcal{O}(n^2)$} & \multirow{2}{*}{$\mathcal{O}(n^2)$} \\
 &  &  &  &  \\ \hline
\multirow{2}{*}{$SCL_{\leq k}$} & \multirow{2}{*}{\begin{tabular}[c]{@{}c@{}}Encode set of \\ AMK constraints\end{tabular}} & \multirow{2}{*}{$(2M - 2)(wk - \frac{k^2 + k}{2} - 1)$} & \multirow{1}{*}{$9Mkw - 5Mk^2 - Mw - 7Mk - 9kw $} & \multirow{2}{*}{$\mathcal{O}(nk)$} \rule{0pt}{10pt}\\
 &  &  & $+ \text{ } 5k^2 - 2M + 2w + 6k + 2$ & \rule{0pt}{10pt} \\ \hline
\end{tabular}
}
\label{tab:info_ladder_amk}
\end{table*}

\textit{Pairwise} encoding generates a clause for every set of \( k + 1 \) variables among the \( w \) variables in each AMK constraint. These clauses ensure that at least one variable in any set of \( k + 1 \) variables is assigned a value of \( \text{false} \). This means that no more than \( k \) variables can be assigned the value \( \text{true} \). By eliminating redundant constraints, \textit{Pairwise} encoding uses \( C^w_{k + 1} \) clauses to encode the first AMK constraint and \( C^{w-1}_{k} \) clauses to encode each of the remaining \( N - 1 \) AMK constraints. As a result, the total number of clauses is \( C^w_{k + 1} + (N - 1)C^{w-1}_{k} \), leading to a complexity of \( \mathcal{O}(n^{k + 1}) \).

\textit{Sequential Counter} (\textit{Seq}) encoding \cite{sinz2005towards} employs a sequential counting technique. In the context of the AMK constraint, this encoding ensures that in any sequence of $w$ consecutive variables, no more than $k$ variables can be assigned $true$, thereby maintaining the AMK property. Each \textit{Seq} encoding for an AMK constraint over $w$ variables requires at most $k(w-2)$ additional variables and generates approximately $3kw -7k + w$ clauses, leading to a total of $Nk(w-2)$ auxiliary variables, $N(3kw -7k + w)$ clauses, and a complexity of $\mathcal{O}(n^2k)$.

\textit{Binary Decision Diagrams} (\textit{BDD}) encoding \cite{abio2012new} represents the AMK constraint as a circuit of \textit{if-then-else} (\textit{ITE}) gates. The goal of this encoding is to constrain the circuit's inputs to ensure that signal propagation through the gates ultimately reaches the final gate. For an AMK constraint on $w$ variables, \textit{BDD} encoding constructs a circuit with approximately $(w - k)(k + 1)$ gates, corresponding to the same number of auxiliary variables. Each gate in the circuit has two outputs, representing the boolean values $true$ and $false$, leading to a total of approximately $2(w - k)(k + 1)$ clauses. In summary, encoding a Ladder AMK constraint using the \textit{BDD} encoding requires approximately $N(w - k)(k + 1)$ auxiliary variables, $2N(w - k)(k + 1)$ clauses, and has a complexity of $\mathcal{O}(n^2k)$.

Leveraging the odd-even merging technique in \textit{Sorting Network} encoding \cite{batcher1968sorting}, Asín et al. introduced the \textit{Half Sorting Network} and \textit{Half Merging Network} encodings \cite{asin2011cardinality}, which reduce the number of clauses by half. Building on these approaches, Asín et al. further proposed the \textit{Simplified Half-Merging Network}, a streamlined version of the \textit{Half Merging Network}. This encoding later served as the foundation for the development of \textit{Cardinality Network} (\textit{Card}) encoding \cite{asin2011cardinality}. For an AMK constraint of $w$ variables, \textit{Card} encoding uses $\mathcal{O}(n\text{log}_2^2k)$ auxiliary variables and $\mathcal{O}(n\text{log}_2^2k)$ clauses. When applied to Ladder AMK constraints, this encoding requires a total of $\mathcal{O}(n^2\text{log}_2^2k)$ auxiliary variables, $\mathcal{O}(n^2\text{log}_2^2k)$ clauses, and has a complexity of $\mathcal{O}(n^2\text{log}_2^2k)$.

\textit{Adder Network} (\textit{Adder}) encoding \cite{een2006translating} translates each AMK constraint into a boolean circuit composed of half-adders and full-adders arranged as a binary adder tree. Through this circuit, a binary representation of the sum of input variables is produced, and a binary comparator circuit is used to ensure that this sum does not exceed $k$. For an AMK constraint over $w$ variables, \textit{Adder} encoding introduces a $\mathcal{O}(n)$ sized translation \cite{een2006translating}. Consequently, encoding a Ladder AMK constraint using \textit{Adder} requires $\mathcal{O}(n^2)$ auxiliary variables and clauses.

Our analysis reveals that the \(SCL_{\leq k}\) encoding results in fewer clauses than other encodings. For simplicity, we assume that the size of the last subset is equal to that of the others, although it is typically smaller. Through block decomposition, one AMK block is generated for the first subset and one for the last subset. For the remaining $(M - 2)$ subsets, a total of $(M - 2)$ AMK blocks and $(M - 2)$ connection blocks are generated. Each AMK block, which is encoded using equations~\eqref{eq:ladder-amk-init-bit} to \eqref{eq:ladder-amk-restrict-bit}, requires $4kw - 2k^2 - 4k -1$ clauses, whereas each connection block, which is encoded using equations~\eqref{eq:ladder-amk-init-bit} to \eqref{eq:ladder-amk-derive-false-bit}, uses $(w - k)$ fewer clauses than the AMK blocks. Furthermore, connecting these $2 + (M -2) + (M-2) = 2M - 2$ blocks necessitates $M-1$ connections, with each connection requiring $(w-k)k$ clauses. In total, the number of clauses employed by the \(SCL_{\leq k}\) encoding is given by:

\vspace{0.5cm}
\resizebox{0.95\textwidth}{!}{
\begin{tabular}{lcccc}
AMK-blocks-clauses & $=$ & $2 + (M - 2)$ & x & $(4kw - 2k^2 - 4k -1)$ \\
Connection-blocks-clauses & $=$ & $(M-2)$ & x & $(4kw -2k^2 - 4k - 1 - w + k)$ \\
Connecting-clauses & $=$ & $(M-1)$ & x & $(w-k)k$ \\
~ & ~ & ~ & ~ & ~ \\ \hline
~ & ~ & ~ & ~ & ~ \\
Total-clauses & $=$ & \multicolumn{3}{c}{$9Mkw - 5Mk^2 - Mw - 7Mk - 9kw + 5k^2 - 2M + 2w + 6k + 2$}
\end{tabular}
}
\vspace{0.5cm}

In the calculation of auxiliary variable usage, \(SCL_{\leq k}\) encoding employs $wk - \frac{k^2 + k}{2}$ register bits per block, including both AMK blocks and connection blocks. By referring to Figure \ref{fig:register_bit_constructing_ladder_at_most_two}, we also observe that the first register bit is equivalent to the first variable, i.e., $R_{1,1,1} \equiv x_4$. Leveraging this insight, we can substitute that register bit with its corresponding variable to further decrease the number of auxiliary variables required by each block by one, resulting in $wk - \frac{k^2 + k}{2} - 1$. Consequently, the total number of auxiliary variables needed for the entire Ladder constraint becomes $(2M - 2)(wk - \frac{k^2 + k}{2} - 1)$.

The analysis above demonstrates the efficiency of our proposed encoding. In terms of the required auxiliary variables and the number of generated clauses, the \(SCL_{\leq k}\) encoding outperforms the \textit{Seq}, \textit{BDD}, \textit{Card}, and \textit{Adder} encodings. Specifically, while these four encodings require at least \(\mathcal{O}(n^2)\) auxiliary variables and clause space, \(SCL_{\leq k}\) only needs \(\mathcal{O}(Mwk)\), which is approximately \(\mathcal{O}(nk)\). 
Additionally, although the \textit{Pairwise} encoding does not utilize auxiliary variables, its complexity of \(\mathcal{O}(n^{k+1})\) is significantly higher than our linear complexity of \(\mathcal{O}(nk)\). We provide a more comprehensive demonstration in the experimental evaluation (Section \ref{sec:eval-ladder-amk}), which clarifies the comparative analysis and highlights the superiority of \(SCL_{\leq k}\) over \textit{Pairwise}, \textit{Seq}, \textit{BDD}, \textit{Card} and \textit{Adder} encodings.

\section{Encodings for Ladder AMO constraint}

In this section, we introduce the \textit{SCL} encoding formulas for the Ladder AMO constraint $(SCL_{\leq 1})$, derived from those for the Ladder AMK constraint ($SCL_{\leq k}$). Additionally, we present \textit{Duplex} encoding \cite{fazekas2020duplex}, a decomposition-based encoding that has proven to be an effective encoding for the Ladder AMO constraint.

\subsection{$SCL_{\leq 1}$ encoding}
\label{sec:scl-encoding-ladder-amo}

\vspace{0.25cm}
\begin{enumerate}[label=\textbf{\alph*.}]
\item \textbf{\textit{Encoding each block in the Ladder AMO constraint}}
\vspace{0.1cm}

\noindent The Ladder AMO constraint is essentially a specific instance of the Ladder AMK constraint. Therefore, its block encoding can be directly derived from that of the Ladder AMK. When \( k=1 \), the formulas from \eqref{eq:ladder-amo-init-bit} to \eqref{eq:ladder-amo-restrict-bit} below are utilized to encode the $i^{th}$ block in the Ladder AMO constraint.

{ \small
\begin{align}
    \label{eq:ladder-amo-init-bit}
    \bigwedge_{j = 1}^{w - 1} x_{i,j} \to R_{i, j, 1}  & \overset{k=1}{\Rightarrow}  \bigwedge_{j=1}^{w-1}x_{i,j}\to R_{i,j,1} \\
    \label{eq:ladder-amo-copy-bit}
    \bigwedge_{j = 2}^{w - 1} \bigwedge_{s = 1}^{\min(j-1, k)} R_{i, j - 1, s} \to R_{i,j, s}  & \overset{k=1}{\Rightarrow}  \bigwedge_{j=2}^{w-1} R_{i,j-1,1} \to R_{i,j,1} \\
    \bigwedge_{j = 2}^{w - 1} \bigwedge_{s = 2}^{\min(j, k)} x_{i,j} \wedge R_{i, j-1, s - 1} \to R_{i,j,s}  & \overset{k=1}{\Rightarrow}  \emptyset \notag\\
    \label{eq:ladder-amo-init-false-bit}
    \bigwedge_{j = 1}^{k} \neg x_{i,j} \to \neg R_{i,j,j}  & \overset{k=1}{\Rightarrow}  \neg x_{i,1} \to \neg R_{i,1,1}\\
    \bigwedge_{j = 2}^{w - 1} \bigwedge_{s = 2}^{min(j, k)} \neg R_{i, j-1, s-1} \to \neg R_{i,j,s}  & \overset{k=1}{\Rightarrow}  \emptyset \notag\\
    \label{eq:ladder-amo-derive-false-bit}
    \bigwedge_{j = 2}^{w - 1} \bigwedge_{s = 1}^{min(j-1, k)} \neg x_{i,j} \wedge \neg R_{i, j-1, s} \to \neg R_{i,j,s}  & \overset{k=1}{\Rightarrow}  \bigwedge_{j=2}^{w-1}\neg x_{i,j}\wedge\neg R_{i,j-1,1}\to \neg R_{i,j,1} \\
    \label{eq:ladder-amo-restrict-bit}
    \bigwedge_{j=k+1}^w x_{i,j} \to \neg R_{i, j - 1, k}  & \overset{k=1}{\Rightarrow}  \bigwedge_{j=2}^{w} x_{i,j} \to \neg R_{i,j-1,1}
\end{align}
}

\vspace{0.25cm}
\item \textbf{\textit{Clauses connecting two adjacent blocks}}
\vspace{0.1cm}

\noindent Similar to the encoding of individual blocks, the connection between blocks in the Ladder AMO constraint can also be derived from the block connection in the Ladder AMK constraint (refer to equation \eqref{eq:connect_amk_blocks}) by setting $k = 1$. Specifically,

\vspace{-0.5cm}
\begin{align*}
    \bigwedge_{j = 2}^{w}\bigwedge_{p =1}^k(\sum_{q=1}^{w-j+1}x_{i,q}\leq k - p \vee \sum_{q'=1}^{j-1}x_{i+1, q'}\leq p -1)
\end{align*}

\[\Big\Downarrow k=1\]

\begin{align}
    \bigwedge_{j = 2}^{w}(\sum_{q=1}^{w-j+1}x_{i,q}\leq 0 \vee \sum_{q'=1}^{j-1}x_{i+1, q'}\leq 0)
\end{align}

\end{enumerate}

\subsection{\textit{Duplex} encoding}
\label{sec:duplex_encoding}

Fazekas et al. introduced the \textit{Duplex} \cite{fazekas2020duplex} for encoding Ladder AMO constraints. In this approach, the AMO and At-Most-Zero (AMZ) sub-expressions obtained through the decomposition of Ladder AMO constraints are encoded by applying \textit{Binary Decision Diagrams} (\textit{BDD}) encoding \cite{bryant1986graph, akers1978binary} for corresponding windows of variables (note that the term ``window'' is equivalent to our term ``subset''). Each window is encoded using two components: a forward \textit{BDD} and a backward \textit{BDD}. The forward \textit{BDD} is constructed from the leftmost to the rightmost variable, while the backward \textit{BDD} is constructed in the reverse direction, from right to left.

\begin{figure*}[h!]
    \centering\
    \resizebox{\textwidth}{!}{%
    \includegraphics{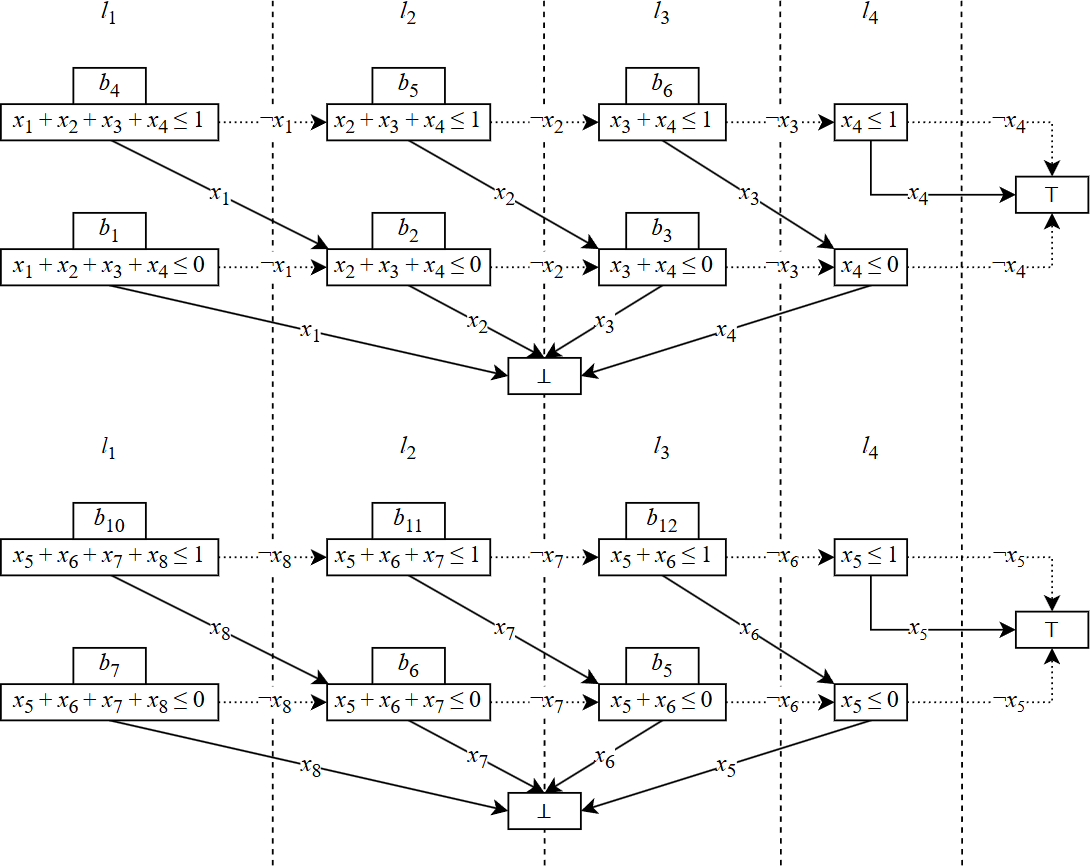}
    }   
    \caption{Forward and backward \textit{BDDs} for Ladder AMO constraint.}
    \begin{center}
        Source: Fazekas et al. \cite{fazekas2020duplex}
    \end{center}
    \label{fig:forward_backward_BDD}
\end{figure*}

Figure \ref{fig:forward_backward_BDD} illustrates the construction of two \textit{BDDs} with the variable arrangements of $x_1 \prec x_2 \prec x_3 \prec x_4$ and $x_8 \prec x_7 \prec x_6 \prec x_5$, respectively. This construction involves making decisions for each variable, using their values to branch the necessary conditions of the original constraints. For instance, to maintain the condition $x_1 + x_2 + x_3 + x_4 \leq 1$, if $x_1$ is $true$, then $x_2 + x_3 + x_4 \leq 0$ must hold; conversely, if $x_1$ is $false$, the condition $x_2 + x_3 + x_4 \leq 1$ must be satisfied.

To ensure there is at most one variable assigned $true$ within the AMO constraints of Ladder AMO constraint, \textit{Duplex} encoding bonds the constructed \textit{BDDs} together. Specifically, for each successive pair of windows $\{\omega_i, \omega_{i+1}\}$, where $\omega^f_i$ is the forward \textit{BDD} of $\omega_i$ and $\omega_{i+1}^b$ is the backward \textit{BDD} of $\omega_{i+1}$, the $j^{th}$ layer of $\omega^f_i$ is connected to the $(w-j+2)^{th}$ layer of $\omega_{i+1}^b$, where $2 \leq j \leq w$. For instance, bonding two \textit{BDDs} as shown in Figure \ref{fig:forward_backward_BDD} requires the following clauses.

\[\wedge (x_2+x_3+x_4 \leq 1) \wedge (x_5 \leq 1) \wedge ((x_2+x_3+x_4 \leq 0) \vee (x_5 \leq 0))\]
\[\wedge (x_3+x_4 \leq 1) \wedge (x_5+x_6 \leq 1) \wedge ((x_3+x_4 \leq 0) \vee (x_5+x_6 \leq 0))\]
\[\wedge (x_4 \leq 1) \wedge (x_5+x_6+x_7 \leq 1) \wedge ((x_4 \leq 0) \vee (x_5+x_6+x_7 \leq 0))\]
\vspace{0.1cm}

Leveraging the benefits of reusing shared sub-expressions, the \textit{Duplex} encoding achieves significantly enhanced complexity compared to non-decompose encodings. As shown in the work of Fazekas et al., the \textit{Duplex} encoding has a complexity of $\mathcal{O}\left(n\right)$ \cite{fazekas2020duplex}, whereas all non-decompose encodings exhibit complexity of $\mathcal{O}\left(n^2\right)$ (refer to Table \ref{tab:info_ladder_amo}). For a Ladder AMO constraint involving $n$ variables and considering $M = \lceil \frac{n}{w} \rceil$ windows of width $w$, with $N = (n-w) +1$ being the number of AMO constraints, this encoding generates around $13Mw -14M -3w + 2$ clauses and introduces additional $4M(w - 1)$ auxiliary variables.

\subsection{Encoding comparison for Ladder AMO constraints}\label{sec:comparison_Ladder_AMO}

In addition to the previously mentioned encodings, which are \textit{Pairwise}, \textit{Sequential Counter} (\textit{Seq}), \textit{Binary Decision Diagrams} (\textit{BDD}), \textit{Cardinality Network} (\textit{Card}), and \textit{Adder Network} (\textit{Adder}), we compare $SCL_{\leq 1}$ with the \textit{Duplex} encoding \cite{fazekas2020duplex}. Moreover, among the group of encodings that individually handle each AMO constraint, we also include the \textit{Product} encoding \cite{chen2010new}.

\begin{table*}[ht!]
\centering
\caption{Size of SAT encoding for a Ladder AMO constraint over $n$ variables with width $w$, $N = (n-w)+1$ and \( M = \left\lceil \frac{n}{w} \right\rceil \).}
\resizebox{\textwidth}{!}{
\begin{tabular}{|c|c|c|c|c|}
\hline
\multirow{2}{*}{Encoding} & \multirow{2}{*}{Technique} & \multirow{2}{*}{Auxiliary variables} & \multirow{2}{*}{Clauses} & \multirow{2}{*}{Complexity} \\
 &  &  &  &  \\ \hline
\multirow{2}{*}{\textit{Pairwise}} & \multirow{12}{*}{\begin{tabular}[c]{@{}c@{}}Encode each \\ AMO constraint \\ separately\end{tabular}} & \multirow{2}{*}{0} & \multirow{2}{*}{$\frac{(w-1)w}{2} + (N-1)(w-1)$} & \multirow{2}{*}{$\mathcal{O}(n^2)$} \\
 &  &  &  &  \\ \cline{1-1} \cline{3-5} 
\multirow{2}{*}{\textit{Seq}} &  & \multirow{2}{*}{$N(w-2)$} & \multirow{2}{*}{$N(4w-7)$} & \multirow{2}{*}{$\mathcal{O}(n^2)$} \\
 &  &  &  &  \\ \cline{1-1} \cline{3-5} 
\multirow{2}{*}{\textit{BDD}} &  & \multirow{2}{*}{$2N(w-1)$} & \multirow{2}{*}{$4N(w-1)$} & \multirow{2}{*}{$\mathcal{O}(n^2)$} \\
 &  &  &  &  \\ \cline{1-1} \cline{3-5} 
\multirow{2}{*}{\textit{Card}} &  & \multirow{2}{*}{$\mathcal{O}(n^2)$} & \multirow{2}{*}{$\mathcal{O}(n^2)$} & \multirow{2}{*}{$\mathcal{O}(n^2)$} \\
 &  &  &  &  \\ \cline{1-1} \cline{3-5} 
\multirow{2}{*}{\textit{Adder}} &  & \multirow{2}{*}{$\mathcal{O}(n^2)$} & \multirow{2}{*}{$\mathcal{O}(n^2)$} & \multirow{2}{*}{$\mathcal{O}(n^2)$} \\
 &  &  &  &  \\ \cline{1-1} \cline{3-5} 
\multirow{2}{*}{\textit{Product}} &  & \multirow{2}{*}{$N(2\sqrt{w} + \mathcal{O}\sqrt[4]{w})$} & \multirow{2}{*}{$N(2w+4\sqrt{w}+\mathcal{O}\sqrt[4]{w})$} & \multirow{2}{*}{$\mathcal{O}(n^2)$} \\
 &  &  &  &  \\ \hline
\multirow{2}{*}{\textit{Duplex}} & \multirow{4}{*}{\begin{tabular}[c]{@{}c@{}}Encode set of \\ AMO constraints\end{tabular}} & \multirow{2}{*}{$4M(w-1)$} & \multirow{2}{*}{$13Mw-14M-3w+2$} & \multirow{2}{*}{$\mathcal{O}(n)$} \\
 &  &  &  &  \\ \cline{1-1} \cline{3-5} 
\multirow{2}{*}{$SCL_{\leq 1}$} &  & \multirow{2}{*}{$(2M - 2)(w - 2)$} & \multirow{2}{*}{$8Mw - 14M - 7w + 13$} & \multirow{2}{*}{$\mathcal{O}(n)$} \\
 &  &  &  &  \\ \hline
\end{tabular}
}
\label{tab:info_ladder_amo}
\end{table*}

Table \ref{tab:info_ladder_amo} presents a summary of the evaluation results for various encoding techniques used to encode Ladder AMO constraints across $n$ variables with a width of $w$. The comparison is based on several criteria, including the number of required auxiliary variables needed, the total number of generated clauses, and the complexity of the clause space. This evaluation incorporates the study by Fazekas et al. \cite{fazekas2020duplex} as well as our proposed \textit{SCL$_{\leq k}$} encoding for the case when $k = 1$.

To represent AMO constraints, the \textit{Cardinality Network} (\textit{Card}) encoding selects $k = 2$ and generates two outputs, i.e. $c_1$ and $c_2$. Then, the output $c_2$ is imposed to be \textit{false}, ensuring that no more than one input variable can be assigned \textit{true}. As a result, the number of auxiliary variables, required clauses, and the overall clause complexity of the \textit{Card} encoding is $\mathcal{O}(n^2 \log_2^2 2) = \mathcal{O}(n^2)$.

The \textit{Product} encoding is based on arranging variables in a two-dimensional coordinate system. The values along each axis are treated as additional variables and are then organized into a smaller two-dimensional coordinate system. This process is recursively applied until further decomposition into smaller coordinate systems is no longer possible. Through this representation technique, \textit{Product} encoding utilizes $N(2\sqrt{w} + \mathcal{O}\sqrt[4]{w})$ auxiliary variables and $N(2w + 4\sqrt{w} + \mathcal{O}\sqrt[4]{w})$ clauses to encode a Ladder AMO constraint.


The \textit{Duplex} encoding employs the decomposition technique and encodes each window separately by generating two \textit{BDDs} with distinct variable arrangements. After that, these constructed windows are bonded together to maintain the AMO property of the original constraints. As presented in Section \ref{sec:duplex_encoding}, the \textit{Duplex} encoding requires approximately \(13Mw - 14M - 3w + 2\) clauses and \(4M(w-1)\) auxiliary variables.

\begin{figure*}[ht!]
    \centering
    \begin{minipage}[t]{0.48\textwidth}
        \centering
        \vspace{0pt}
        \includegraphics[width=\linewidth]{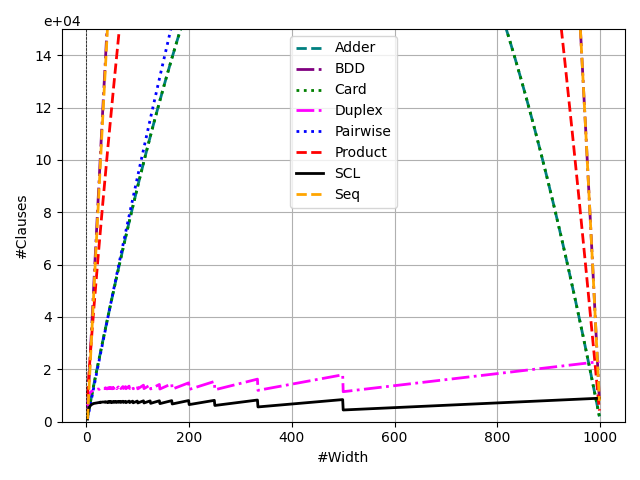}
        \caption{Comparison of the number of clauses of a Ladder AMO constraint with parameters set to $n=1000$ and $w\in(1,1000)$.}
        \label{fig:comparision_clauses_ladder_amo}
    \end{minipage}
    \hfill
    \begin{minipage}[t]{0.48\textwidth}
        \centering
        \vspace{0pt}
        \includegraphics[width=\linewidth]{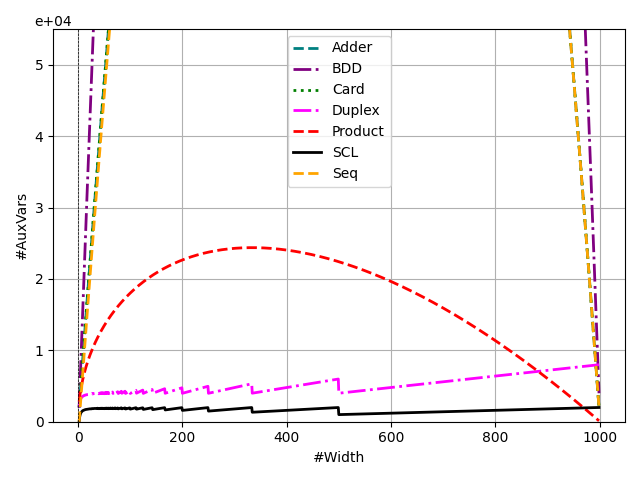}
        \caption{Comparison of the number of auxiliary variables of a Ladder AMO constraint with parameters set to $n=1000$ and $w\in(1,1000)$.}
        \label{fig:comparision_aux_vars_ladder_amo}
    \end{minipage}
\end{figure*}

The analysis presented above underscores the efficiency of our proposed encoding. Like the $SCL_{\leq k}$ encoding, the $SCL_{\leq 1}$ encoding outperforms several alternatives, including \textit{Pairwise}, \textit{Seq}, \textit{BDD}, \textit{Card}, and \textit{Adder} encodings. As detailed in Table \ref{tab:info_ladder_amo}, the complexity of $SCL_{\leq 1}$ is $\mathcal{O}(n)$, which is also superior to $\mathcal{O}(n^2)$ of the \textit{Product} encoding. Furthermore, among linear complexity encodings, $SCL_{\leq 1}$ uses roughly half the number of variables and clauses compared to the \textit{Duplex} encoding.
To distinctly showcase the outstanding performance of our encoding, we present Figures \ref{fig:comparision_clauses_ladder_amo} and \ref{fig:comparision_aux_vars_ladder_amo}, which illustrate the performance efficiency of the $SCL_{\leq 1}$ encoding. Figure \ref{fig:comparision_clauses_ladder_amo} depicts the number of clauses, while Figure \ref{fig:comparision_aux_vars_ladder_amo} illustrates the number of auxiliary variables required to encode a set of Ladder AMO constraints for $n=1000$ and $w\in(1,1000)$, based on the calculations presented in Table \ref{tab:info_ladder_amo}.

\section{SAT solving for the Anti-bandwidth problem}

The Anti-bandwidth problem (ABP) is an NP-hard problem with applications in various scheduling scenarios \cite{leung1984some}, including radio frequency assignment \cite{hale1980frequency}, obnoxious facility location \cite{cappanera1999survey}, and map coloring \cite{hu2010visualizing}. In this section, we present SAT formulae, including the Ladder AMO constraints, for solving the ABP. 

\subsection{Problem definition}
Let $G = (V, E)$ be a graph where $V$ and $E$ denote the sets of vertices and edges, respectively. A labeling function $f$ is a bijection from $V$ to $\{1, 2, \ldots, |V|\}$, assigning a unique label to each vertex in $V$. The anti-bandwidth $AB_f(i)$ of a vertex $i \in V$ under the labeling function $f$ is defined as the minimum absolute difference between the label of $i$ and the labels of its adjacent vertices:

\[
    AB_{f}(i) = min\{|f(i) - f(i')| : \{i, i'\} \in E\}
\]

\noindent The anti-bandwidth of $G$ under the labeling function $f$, denoted as $AB_{f}(G)$, is the minimal value among the $AB_{f}(i)$ values:

\[
    AB_{f}(G) = min\{AB_{f}(i) : i \in V\}
\]

\noindent The Anti-bandwidth Problem aims to find the maximum anti-bandwidth value of  $G$, denoted as $AB(G)$, over all feasible labeling functions $f$.

\[
    AB(G) = \max AB_{f}(G)
\]

\subsection{ILP model for ABP}
\label{sec:iterativeMIPapproach}
The ABP is expressible as an Integer Linear Programming (ILP) \cite{sinnl2021note} model with binary (0–1) variables. 
By defining $x_{i}^{l}$ to be a boolean variable that holds $true$ if vertex $i$ is assigned label $l$, and $false$ otherwise, this model employs two formulas, \eqref{eq:vertices} and \eqref{eq:labels}, to guarantee that every vertex in the graph receives a unique label. Specifically, equation \eqref{eq:vertices} ensures that each label is allocated to exactly one vertex, while equation \eqref{eq:labels} guarantees that each vertex is given precisely one label. Combining these two equations preserves the one-to-one relationship between vertices and labels.

\begin{align}
&\sum_{i\in V}^{}x_{i}^{l} = 1 && \forall l\in \{1,\ldots,|V|\} \tag{VERTICES} \label{eq:vertices}
\end{align}
\begin{align}
&\sum_{l=1}^{|V|}x_{i}^{l} = 1  && \forall i \in V \tag{LABELS} \label{eq:labels}
\end{align}

Based on this approach, Markus Sinnl \cite{sinnl2021note} proposed an iterative formulation, framed as a feasibility problem, to answer the question: ``Does there exist a solution with $AB(G) \geq k+1$?''. This formulation includes both formulas \eqref{eq:vertices} and \eqref{eq:labels}, along with equation \eqref{eq:obj-k}, which ensures that the difference between the labels of adjacent vertices $i$ and $i'$ is strictly greater than $k$ whenever $k + 1$ is a feasible anti-bandwidth.

\begin{align}
&\sum_{l'=l_{2}}^{l_{2+k}}(x_{i}^{l'} + x_{i'}^{l'})\leq 1  && \forall \{i, i'\} \in E, \forall 1 \leq l_2 \leq |V| - k \tag{OBJ-k} \label{eq:obj-k} 
\end{align}

For the ABP problem with 10 vertices (\(|V| = 10\)) and \(k = 3\), the constraints given by equation \eqref{eq:obj-k} for any pair of vertices \(i\) and \(i'\) are illustrated in Figure \ref{fig:obj-k-10n-3k}. These constraints ensure that the labels \(l_i\) and \(l_{i'}\) of vertices \(i\) and \(i'\), respectively, always satisfy the condition \(|l_i - l_{i'}| \leq k + 1 = 4\). As an example, if \(l_i = 4\), meaning \(x_i^4\) holds \textit{true}, then to prevent the set of AMO constraints (in Figure \ref{fig:obj-k-10n-3k}) from being violated, \(x_{i'}^1, x_{i'}^2, x_{i'}^3, x_{i'}^4, x_{i'}^5, x_{i'}^6\), and \(x_{i'}^7\) must all be \textit{false}, which implies that \(l_{i'} \notin \{1, 2, 3, 4, 5, 6, 7\}\). The remaining possible labels for vertex \(i'\) are \(\{8, 9, 10\}\), ensuring that the label distance condition between vertices is maintained.

\begin{figure}[ht!]
    \centering
    \resizebox{!}{!}{
    \begin{tabular}{cccccccccccccccc}
    $x_{i}^{1}$ & $+$ & $x_{i}^{2}$ & $+$ & $x_{i}^{3}$ & $+$ & $x_{i}^{4}$ & $+$ & $x_{i'}^{1}$ & $+$ & $x_{i'}^{2}$ & $+$ & $x_{i'}^{3}$ & $+$ & $x_{i'}^{4}$ & $\leq 1$ \rule{0pt}{15pt} \\
    $x_{i}^{2}$ & $+$ & $x_{i}^{3}$ & $+$ & $x_{i}^{4}$ & $+$ & $x_{i}^{5}$ & $+$ & $x_{i'}^{2}$ & $+$ & $x_{i'}^{3}$ & $+$ & $x_{i'}^{4}$ & $+$ & $x_{i'}^{5}$ & $\leq 1$ \rule{0pt}{15pt} \\
    $x_{i}^{3}$ & $+$ & $x_{i}^{4}$ & $+$ & $x_{i}^{5}$ & $+$ & $x_{i}^{6}$ & $+$ & $x_{i'}^{3}$ & $+$ & $x_{i'}^{4}$ & $+$ & $x_{i'}^{5}$ & $+$ & $x_{i'}^{6}$ & $\leq 1$ \rule{0pt}{15pt} \\
    $x_{i}^{4}$ & $+$ & $x_{i}^{5}$ & $+$ & $x_{i}^{6}$ & $+$ & $x_{i}^{7}$ & $+$ & $x_{i'}^{4}$ & $+$ & $x_{i'}^{5}$ & $+$ & $x_{i'}^{6}$ & $+$ & $x_{i'}^{7}$ & $\leq 1$ \rule{0pt}{15pt} \\
    $x_{i}^{5}$ & $+$ & $x_{i}^{6}$ & $+$ & $x_{i}^{7}$ & $+$ & $x_{i}^{8}$ & $+$ & $x_{i'}^{5}$ & $+$ & $x_{i'}^{6}$ & $+$ & $x_{i'}^{7}$ & $+$ & $x_{i'}^{8}$ & $\leq 1$ \rule{0pt}{15pt} \\
    $x_{i}^{6}$ & $+$ & $x_{i}^{7}$ & $+$ & $x_{i}^{8}$ & $+$ & $x_{i}^{9}$ & $+$ & $x_{i'}^{6}$ & $+$ & $x_{i'}^{7}$ & $+$ & $x_{i'}^{8}$ & $+$ & $x_{i'}^{9}$ & $\leq 1$ \rule{0pt}{15pt} \\
    $x_{i}^{7}$ & $+$ & $x_{i}^{8}$ & $+$ & $x_{i}^{9}$ & $+$ & $x_{i}^{10}$ & $+$ & $x_{i'}^{7}$ & $+$ & $x_{i'}^{8}$ & $+$ & $x_{i'}^{9}$ & $+$ & $x_{i'}^{10}$ & $\leq 1$ \rule{0pt}{15pt}
    \end{tabular}
    }
    \caption{OBJ-k constraint of \{$i$, $i'$\} with $|V| = 10$ and $k=3$.}
    \label{fig:obj-k-10n-3k}
\end{figure}

\subsection{SAT implementation of ABP constraints}

\subsubsection{Decomposition of $OBJ-k$ constraints}

\noindent Encoding the equation \eqref{eq:obj-k} requires a total of $|E|$ separate constraints, each corresponding to an edge in the set \(E\). However, in most graphs, the number of edges is significantly greater than the number of vertices. To optimize the encoding of \eqref{eq:obj-k}, Fazekas et al. \cite{fazekas2020duplex} employ Lemma \ref{lemma:decompose_AMO_constraint} to decompose the constraints of \eqref{eq:obj-k} as follows:

\[\sum_{l'=l_{2}}^{l_{2+k}}(x_{i}^{l'} + x_{i'}^{l'})\leq 1\]
\[\overset{\text{Lemma \ref{lemma:decompose_AMO_constraint}}}{\equiv}\]
\[\sum_{l'=l_{2}}^{l_{2+k}}x_{i}^{l'}\leq 1 \wedge\sum_{l'=l_{2}}^{l_{2+k}}x_{i'}^{l'}\leq 1 \wedge (\sum_{l'=l_{2}}^{l_{2+k}}x_{i}^{l'}\leq 0 \vee\sum_{l'=l_{2}}^{l_{2+k}}x_{i'}^{l'}\leq 0)\]

Using this decomposition technique, the constraints from equation \eqref{eq:obj-k} for any edge \( \{x_i, x_{i'}\} \) can be represented through the corresponding vertex-level constraints of \( x_i \) and \( x_{i'} \). These vertex-level constraints can be reused to represent multiple edges, which reduces the total number of constraints needed from \( |E| \) to \( |V| \).

\begin{figure*}[ht!]
    \centering
    \begin{minipage}[t]{\textwidth}
        \centering
        \vspace{0pt}
        \resizebox{!}{!}{
        \begin{tabular}{c}
        $x_{i}^{2} + x_{i}^{3} + x_{i}^{4} + x_{i}^{5} + x_{i'}^{2} + x_{i'}^{3} + x_{i'}^{4} + x_{i'}^{5} \leq 1$ \rule{0pt}{15pt} \\
        $\equiv$ \rule{0pt}{15pt} \\
        $(x_{i}^{2} + x_{i}^{3} + x_{i}^{4} + x_{i}^{5} \leq 1) \wedge (x_{i'}^{2} + x_{i'}^{3} + x_{i'}^{4} + x_{i'}^{5} \leq 1)$ \rule{0pt}{15pt} \\
        $\wedge  (x_{i}^{2} + x_{i}^{3} + x_{i}^{4} + x_{i}^{5} \leq 0 \vee x_{i'}^{2} + x_{i'}^{3} + x_{i'}^{4} + x_{i'}^{5} \leq 0)$ \rule{0pt}{15pt}
        \end{tabular}
        } \\
        \caption{\small Decomposition of the second AMO constraint in \eqref{eq:obj-k} constraint of edge $\{i, i'\}$ with $|V| = 10$ and $k = 3$.}
        \label{fig:decomposition-of-the-second-AMO-constraint}
    \end{minipage}
    \hfill
    \begin{minipage}[t]{0.48\textwidth}
        \centering
        \vspace{0pt}
        \resizebox{\textwidth}{!}{
        \begin{tabular}{cccccccc}
        $x_{i}^{1}$ & $+$ & $x_{i}^{2}$ & $+$ & $x_{i}^{3}$ & $+$ & $x_{i}^{4}$ & $\leq 1$  \rule{0pt}{15pt} \\
        $x_{i}^{2}$ & $+$ & $x_{i}^{3}$ & $+$ & $x_{i}^{4}$ & $+$ & $x_{i}^{5}$ & $\leq 1$  \rule{0pt}{15pt} \\
        $x_{i}^{3}$ & $+$ & $x_{i}^{4}$ & $+$ & $x_{i}^{5}$ & $+$ & $x_{i}^{6}$ & $\leq 1$  \rule{0pt}{15pt} \\
        $x_{i}^{4}$ & $+$ & $x_{i}^{5}$ & $+$ & $x_{i}^{6}$ & $+$ & $x_{i}^{7}$ & $\leq 1$  \rule{0pt}{15pt} \\
        $x_{i}^{5}$ & $+$ & $x_{i}^{6}$ & $+$ & $x_{i}^{7}$ & $+$ & $x_{i}^{8}$ & $\leq 1$  \rule{0pt}{15pt} \\
        $x_{i}^{6}$ & $+$ & $x_{i}^{7}$ & $+$ & $x_{i}^{8}$ & $+$ & $x_{i}^{9}$ & $\leq 1$  \rule{0pt}{15pt} \\
        $x_{i}^{7}$ & $+$ & $x_{i}^{8}$ & $+$ & $x_{i}^{9}$ & $+$ & $x_{i}^{10}$ & $\leq 1$  \rule{0pt}{15pt}
        \end{tabular}
        }
        \caption{\small Ladder AMO constraint associated with vertex $i$ with $|V| = 10$ and $k = 3$.}
        \label{fig:Ladder-AMO-constraint-of-vertex-i}
    \end{minipage}
    \hfill
    \begin{minipage}[t]{0.49\textwidth}
        \centering
        \vspace{0pt}
        \resizebox{\textwidth}{!}{
        \begin{tabular}{cccccccc}
        $x_{i'}^{1}$ & $+$ & $x_{i'}^{2}$ & $+$ & $x_{i'}^{3}$ & $+$ & $x_{i'}^{4}$ & $\leq 1$  \rule{0pt}{15pt} \\
        $x_{i'}^{2}$ & $+$ & $x_{i'}^{3}$ & $+$ & $x_{i'}^{4}$ & $+$ & $x_{i'}^{5}$ & $\leq 1$  \rule{0pt}{15pt} \\
        $x_{i'}^{3}$ & $+$ & $x_{i'}^{4}$ & $+$ & $x_{i'}^{5}$ & $+$ & $x_{i'}^{6}$ & $\leq 1$  \rule{0pt}{15pt} \\
        $x_{i'}^{4}$ & $+$ & $x_{i'}^{5}$ & $+$ & $x_{i'}^{6}$ & $+$ & $x_{i'}^{7}$ & $\leq 1$  \rule{0pt}{15pt} \\
        $x_{i'}^{5}$ & $+$ & $x_{i'}^{6}$ & $+$ & $x_{i'}^{7}$ & $+$ & $x_{i'}^{8}$ & $\leq 1$  \rule{0pt}{15pt} \\
        $x_{i'}^{6}$ & $+$ & $x_{i'}^{7}$ & $+$ & $x_{i'}^{8}$ & $+$ & $x_{i'}^{9}$ & $\leq 1$  \rule{0pt}{15pt} \\
        $x_{i'}^{7}$ & $+$ & $x_{i'}^{8}$ & $+$ & $x_{i'}^{9}$ & $+$ & $x_{i'}^{10}$ & $\leq 1$  \rule{0pt}{15pt}
        \end{tabular}
        }
        \caption{\small Ladder AMO constraint associated with vertex $i'$ with $|V| = 10$ and $k = 3$.}
        \label{fig:Ladder-AMO-constraint-of-vertex-i'}
    \end{minipage}
    \hfill
    \begin{minipage}[t]{\textwidth}
        \centering
        \vspace{0pt}
        \resizebox{\textwidth}{!}{
        \begin{tabular}{ccccccccccccccccc}
        $x_{i}^{1}$ & $+$ & $x_{i}^{2}$ & $+$ & $x_{i}^{3}$ & $+$ & $x_{i}^{4}$ & $\leq 0$ & $\vee$ & $x_{i'}^{1}$ & $+$ & $x_{i'}^{2}$ & $+$ & $x_{i'}^{3}$ & $+$ & $x_{i'}^{4}$ & $\leq 0$  \rule{0pt}{15pt} \\
        $x_{i}^{2}$ & $+$ & $x_{i}^{3}$ & $+$ & $x_{i}^{4}$ & $+$ & $x_{i}^{5}$ & $\leq 0$ & $\vee$ & $x_{i'}^{2}$ & $+$ & $x_{i'}^{3}$ & $+$ & $x_{i'}^{4}$ & $+$ & $x_{i'}^{5}$ & $\leq 0$  \rule{0pt}{15pt} \\
        $x_{i}^{3}$ & $+$ & $x_{i}^{4}$ & $+$ & $x_{i}^{5}$ & $+$ & $x_{i}^{6}$ & $\leq 0$ & $\vee$ & $x_{i'}^{3}$ & $+$ & $x_{i'}^{4}$ & $+$ & $x_{i'}^{5}$ & $+$ & $x_{i'}^{6}$ & $\leq 0$  \rule{0pt}{15pt} \\
        $x_{i}^{4}$ & $+$ & $x_{i}^{5}$ & $+$ & $x_{i}^{6}$ & $+$ & $x_{i}^{7}$ & $\leq 0$ & $\vee$ & $x_{i'}^{4}$ & $+$ & $x_{i'}^{5}$ & $+$ & $x_{i'}^{6}$ & $+$ & $x_{i'}^{7}$ & $\leq 0$  \rule{0pt}{15pt} \\
        $x_{i}^{5}$ & $+$ & $x_{i}^{6}$ & $+$ & $x_{i}^{7}$ & $+$ & $x_{i}^{8}$ & $\leq 0$ & $\vee$ & $x_{i'}^{5}$ & $+$ & $x_{i'}^{6}$ & $+$ & $x_{i'}^{7}$ & $+$ & $x_{i'}^{8}$ & $\leq 0$  \rule{0pt}{15pt} \\
        $x_{i}^{6}$ & $+$ & $x_{i}^{7}$ & $+$ & $x_{i}^{8}$ & $+$ & $x_{i}^{9}$ & $\leq 0$ & $\vee$ & $x_{i'}^{6}$ & $+$ & $x_{i'}^{7}$ & $+$ & $x_{i'}^{8}$ & $+$ & $x_{i'}^{9}$ & $\leq 0$  \rule{0pt}{15pt} \\
        $x_{i}^{7}$ & $+$ & $x_{i}^{8}$ & $+$ & $x_{i}^{9}$ & $+$ & $x_{i}^{10}$ & $\leq 0$ & $\vee$ & $x_{i'}^{7}$ & $+$ & $x_{i'}^{8}$ & $+$ & $x_{i'}^{9}$ & $+$ & $x_{i'}^{10}$ & $\leq 0$  \rule{0pt}{15pt}
        \end{tabular}
        }
        \caption{\small Connecting clauses obtained from the decomposition of \eqref{eq:obj-k} constraint of edge $\{i, i'\}$ with $|V| = 10$ and $k = 3$.}
        \label{fig:or-connected-clauses-from-decomposition-OBJ-k}
    \end{minipage}
\end{figure*}

In particular, each AMO constraint in equation~\eqref{eq:obj-k} is decomposed into two sub-AMO constraints, each involving only the variables associated with vertex~$i$ or vertex~$i'$ (see Figure~\ref{fig:decomposition-of-the-second-AMO-constraint} for an illustration). By decomposing all AMO constraints in \eqref{eq:obj-k}, the obtained sub-AMO constraints form two Ladder AMO constraints, as presented in Figures~\ref{fig:Ladder-AMO-constraint-of-vertex-i} and \ref{fig:Ladder-AMO-constraint-of-vertex-i'}. These Ladder constraints can be directly encoded using the proposed $SCL_{\leq 1}$ encoding.

Additionally, the decomposition of \eqref{eq:obj-k} uses disjunctions as connecting clauses to connect vertices $i$ and $i'$, thereby ensuring that the AMO property is maintained for the \eqref{eq:obj-k} constraint over the edge $\{i, i'\}$ (see Figure~\ref{fig:or-connected-clauses-from-decomposition-OBJ-k}). These connecting clauses can be encoded using register bits required by the \( SCL_{\leq 1} \) encoding of the corresponding \eqref{eq:obj-k} constraints, in accordance with Proposition \ref{prop:decomposeAMZconstraint}.

\vspace{0.25cm}

\begin{proposition}
    For any integer number $i$ s.t. $1 \leq i < n$, the constraint
    \[x_{1} + x_{2} + \ldots + x_{n} \leq 0\]
    holds if and only if the following constraint holds:
    \[(x_{1} + \ldots + x_{i}\leq 0)\wedge (x_{i+1} + \ldots + x_{n}\leq 0).\]
    \label{prop:decomposeAMZconstraint}
\end{proposition}

For instance, the decomposition of \( x_{i}^{2} + x_{i}^{3} + x_{i}^{4} + x_{i}^{5} + x_{i'}^{2} + x_{i'}^{3} + x_{i'}^{4} + x_{i'}^{5} \leq 1 \) utilizes \(x_{i}^{2} + x_{i}^{3} + x_{i}^{4} + x_{i}^{5} \leq 0 \vee x_{i'}^{2} + x_{i'}^{3} + x_{i'}^{4} + x_{i'}^{5} \leq 0\), as presented in Figure~\ref{fig:decomposition-of-the-second-AMO-constraint}. Applying Proposition \ref{prop:decomposeAMZconstraint}, \( x_{i}^{2} + x_{i}^{3} + x_{i}^{4} + x_{i}^{5} \leq 0 \) (resp. \( x_{i'}^{2} + x_{i'}^{3} + x_{i'}^{4} + x_{i'}^{5} \leq 0 \)) can be further decomposed into \(x_{i}^{2} + x_{i}^{3} + x_{i}^{4} \leq 0 \wedge x_{i}^{5} \leq 0\) (resp. \(x_{i'}^{2} + x_{i'}^{3} + x_{i'}^{4} \leq 0 \wedge x_{i'}^{5} \leq 0\)). These clauses can then be substituted with the corresponding \textit{SC} register bits, as discussed in Section~\ref{sec:encode-ladder-amk-constraint}.

To enhance our approach, we implement the constraint \eqref{eq:labels} by utilizing auxiliary variables when constructing \( SCL_{\leq 1} \) for each vertex in the graph. This strategy improves the optimization of the generated clauses, providing a more efficient alternative to encoding with \( x_i^l \) for the constraint \eqref{eq:labels}. The remaining \eqref{eq:vertices} constraint is applied with exactly one encoding in SAT using a combination of \textit{Product} and \textit{Seq} encodings.

\subsubsection{Symmetry breaking}

\noindent Referring to Markus Sinnl \cite{sinnl2021note}, for every labeling $f$ of an n-vertex graph, there exists a reversed labeling $f'$, where each label in $f'$ is computed from its corresponding label in $f$ using the following linear transformation:
\[f' = n + 1 - f.\]
This transformation ensures a one-to-one correspondence between the values in $f$ and those in $f'$. Consequently, if $f$ satisfies $(VERTICES)$ and $(LABELS)$ constraints, then $f'$ also satisfies them. Moreover, this transformation also ensures that $f'$ maintains the same anti-bandwidth as $f$, which is proven as follows. 

\vspace{0.25cm}
\begin{center}
\resizebox{!}{!}{
\begin{tabular}{rcl}
     $\left|f'(i) -f'(i')\right|$ & $=$ & $\left|(n + 1 - f(i)) - (n + 1 - f(i'))\right|$ \\
     & $=$ & $\left| f(i') - f(i)\right|$
\end{tabular}
}
\end{center}
\vspace{0.25cm}

Based on this observation, we apply the symmetry breaking technique \cite{gent2006symmetry} to reduce the search space. In our implementation of ABP, we employ symmetry breaking at one selected node using two different configurations: the first node and the highest-degree node.

\section{SAT solving for the Nurse Rostering problem}
\label{sec:sat-nrp}

\subsection{Problem definition and ILP model}

The Nurse Rostering Problem (NRP) aims to construct a work schedule for one or more nurses over a planning horizon of $d$ days. On each day, a nurse may be assigned to a day shift (D), an evening shift (E), or a night shift (N). If the nurse is not assigned to any shift (D, E, or N) on a day, the nurse has an off-day (O). The schedule must satisfy a set of constraints related to work frequency and rest periods in order to ensure nurses' well-being and workability. 

This study considers the NRP variant used by Bergman et al. \cite{bergman2014mdd}, which focuses on designing a schedule for a single nurse, subject to the scheduling requirements listed in Table~\ref{tab:nrp-requirements}. Although this variant focuses on a single nurse, it captures the essential workload distributions and restrictions that also arise in the multi-nurse problem.

Let $x_{i,j}$ be a boolean variable that denotes whether the nurse is assigned to shift $j$ on day $i$. More precisely, if the nurse is assigned to shift $j$ on day $i$, then $x_{i,j} = true$. Otherwise, if the nurse is not assigned to shift $j$ on day $i$, then $x_{i,j} = false$. Additionally, $x_{i,O} = true$ indicates that the nurse has an off-day on day $i$. Otherwise, the nurse must be assigned to exactly one of the three working shifts on day $i$. Since the nurse either has to work exactly one shift per day or has an off-day, the assignment variables must satisfy
\[
\sum_{j \in \{D,E,N,O\}} x_{i,j} = 1 
\qquad \forall i = 1,\dots,d .
\]

Table~\ref{tab:nrp-ilp-model} presents the ILP model to construct the schedule that satisfies the requirements outlined in Table~\ref{tab:nrp-requirements}. In this table, each schedule requirement, except the first one, is expressed as a sequence of cardinality constraints.

\noindent
\begin{minipage}[b]{0.48\textwidth}
    \centering
    \captionof{table}{\small Schedule requirements of NRP \cite{bergman2014mdd}.}
    \resizebox{\textwidth}{!}{
        \begin{tabular}{cl}
            \hline
            1. & At most 1 work shift every day \rule{0pt}{15pt} \\
            2. & At least 20 work shifts every 28 days \rule{0pt}{15pt} \\
            3. & At least 4 off-days every 14 days \rule{0pt}{15pt} \\
            4. & At least 1 night shift every 14 days \rule{0pt}{15pt} \\
            5. & At most 4 night shifts every 14 days \rule{0pt}{15pt} \\
            6. & At least 4 evening shifts every 14 days \rule{0pt}{15pt} \\
            7. & At most 8 evening shifts every 14 days \rule{0pt}{15pt} \\
            8. & Nights shifts is not on consecutive days \rule{0pt}{15pt} \\
            9. & At least 2 evening/night shifts every 7 days \rule{0pt}{15pt} \\
            10. & At most 4 evening/night shifts every 7 days \rule{0pt}{15pt} \\
            11. & At most 6 work shifts every 7 days \rule{0pt}{15pt} \\ 
            \hline \\
        \end{tabular}
    }
    \label{tab:nrp-requirements}
\end{minipage}
\hfill
\begin{minipage}[b]{0.49\textwidth}
    \centering
    \captionof{table}{\small ILP model of NRP.}
    \resizebox{\textwidth}{!}{
        \begin{tabular}{cll}
         \hline
         1. & \(\sum_{j \in \{D, E, N, O\}} x_{i,j} = 1\) & \(\forall i \in 1,\ldots,d\) \rule{0pt}{14.3pt}\\
         2. & \(\sum_{i' = i}^{i + 27} \neg x_{i', O} \geq 20\) & \(\forall i \in 1,\ldots,d - 27\) \rule{0pt}{14.3pt}\\
         3. & \(\sum_{i' = i}^{i + 13} x_{i', O} \geq 4\) & \(\forall i \in 1,\ldots,d - 13\) \rule{0pt}{14.3pt}\\
         4. & \(\sum_{i' = i}^{i + 13} x_{i', N} \geq 1\) & \(\forall i \in 1,\ldots,d - 13\) \rule{0pt}{14.3pt}\\
         5. & \(\sum_{i' = i}^{i + 13} x_{i', N} \leq 4\) & \(\forall i \in 1,\ldots,d - 13\) \rule{0pt}{14.3pt}\\
         6. & \(\sum_{i' = i}^{i + 13} x_{i', E} \geq 4\) & \(\forall i \in 1,\ldots,d - 13\) \rule{0pt}{14.3pt}\\
         7. & \(\sum_{i' = i}^{i + 13} x_{i', E} \leq 8\) & \(\forall i \in 1,\ldots,d - 13\) \rule{0pt}{14.3pt}\\
         8. & \(\sum_{i' = i}^{i + 1} x_{i', N} \leq 1\) & \(\forall i \in 1,\ldots,d - 1\) \rule{0pt}{14.3pt}\\ 
         9. & \(\sum_{i' = i}^{i + 6} x_{i', E} + x_{i', N} \geq 2\) & \(\forall i \in 1,\ldots,d - 6\) \rule{0pt}{14.3pt}\\
         10. & \(\sum_{i' = i}^{i + 6} x_{i', E} + x_{i', N} \leq 4\) & \(\forall i \in 1,\ldots,d - 6\) \rule{0pt}{14.3pt}\\
         11. & \(\sum_{i' = i}^{i + 6} \neg x_{i', O} \leq 6\) & \(\forall i \in 1,\ldots,d - 6\) \rule{0pt}{14.3pt}\\ \hline \\
        \end{tabular}
    }
    \label{tab:nrp-ilp-model}
\end{minipage}


\subsection{SAT implementation of NRP constraints}
\label{sec:nrp-sat-implementation}

The ILP model in Table~\ref{tab:nrp-ilp-model} primarily consists of sequence constraints with ladder structure over the planning horizon. 
Some of them are naturally the Ladder AMK constraint with $k \geq 2$, which can be encoded by applying the proposed encoding directly.
However, several of these constraints are of the At-Least-K type with $k \geq 2$, such as constraints (2), (3), (6), and (9). 
Therefore, to leverage the advantages of our proposed encoding, we systematically transform these constraints into equivalent Ladder At-Most-K constraints.

In particular, constraint (2) requires at least 20 working days in every 28 consecutive days. 
Since each day is either a working day or an off-day, this constraint is equivalent to limiting the number of off-days to at most 8 in every such consecutive days. 
\[\sum_{i' = i}^{i + 27} \neg x_{i', O} \geq 20 \equiv\sum_{i' = i}^{i + 27} x_{i', O} \leq 8 \]
Similarly, constraint (3), which requires at least 4 off-days every 14 days, can be reformulated as restricting the number of working days to at most 10.
\[\sum_{i' = i}^{i + 13} x_{i', O} \geq 4 \equiv \sum_{i' = i}^{i + 13} \neg x_{i', O} \leq 10\]
Constraint (6), imposing at least 4 evening shifts every 14 days, can be reformulated as restricting the number of days the nurse does not work in the evening to at most 10.
\[\sum_{i' = i}^{i + 13} x_{i', E} \geq 4 \equiv \sum_{i' = i}^{i + 13} \neg x_{i', E} \leq 10\]
Since the nurse has to work at most 1 shift every day, constraint (9), which states that the nurse has to work at least 2 evening or night shifts every 7 days, is equivalent to imposing that the total number of day shifts and off-days is at most 5 every 7 days.
\[\sum_{i' = i}^{i + 6} x_{i', E} + x_{i', N} \geq 2 \equiv \sum_{i' = i}^{i + 6} x_{i', D} + x_{i', O} \leq 5\]

\section{Experimental Evaluation}
\label{sec:experiments}

We conducted three experiments to evaluate the effectiveness of our proposed SAT encoding \(SCL\). The first experiment compares the solving times of \(SCL\) against other SAT encodings when addressing single Ladder AMO constraints (\(k=1\)) and Ladder AMK constraints (\(k>1\)). The second and third experiments, respectively, focus on evaluating \(SCL_{\leq 1}\) and \(SCL_{\leq k}\) in solving the Anti-bandwidth Problem and Nurse Rostering Problem. 
These three experiments were performed on a cluster provided by the \textit{Google Cloud Platform}\footnote{\url{https://console.cloud.google.com/compute}} featuring \textit{8 vCPUs, 4 cores, and 64GB of memory}, running the \textit{Debian GNU/Linux 12 operating system}. The SAT solver used was CaDiCaL version 1.2.1 \cite{BiereSATRace2019}. 

\subsection{Single Ladder AMK constraint}
The number of variables and clauses generated by the $SCL_{\leq 1}$ and $SCL_{\leq k}$ encodings, compared to other SAT encodings, is detailed in Tables \ref{tab:info_ladder_amk} and \ref{tab:info_ladder_amo} of the paper. In this experiment, we aim to evaluate the solving time of $SCL$ relative to other SAT encodings when given a Ladder AMK constraint as input, while varying the number of variables and the width $w$. 

\subsubsection{Solving time of the Ladder AMO constraint}
\label{sec:eval-ladder-amo}

\begin{figure*}[ht!]
    \centering
    \begin{minipage}{0.49\textwidth}
        \centering
        \includegraphics[width=\linewidth]{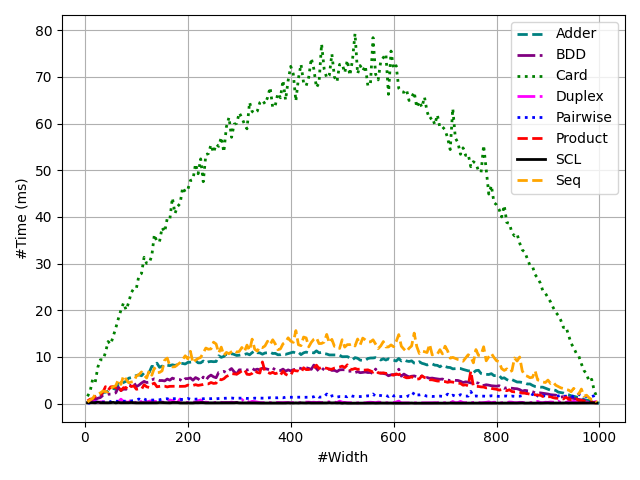}
        \caption{Solving time of various encodings on the Ladder AMO constraints with $n=1000$ and $w\in \{5, 10, 15,...,995\}$.}
        \label{fig:ladder_amo_runtime}
    \end{minipage}%
    \hfill
    \begin{minipage}{0.49\textwidth}
        \centering
        \includegraphics[width=\linewidth]{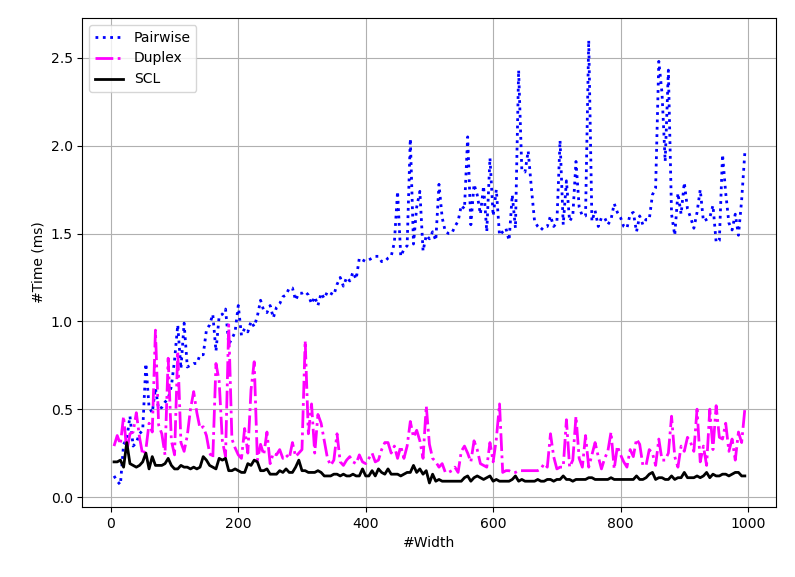}
        \caption{Solving time of the top three encodings on the Ladder AMO constraints with $n=1000$ and $w\in \{5, 10, 15,...,995\}$.}
        \label{fig:ladder_amo_runtime_best_3}
    \end{minipage}
\end{figure*}

The experimental dataset for the Ladder AMO constraint consists of a fixed number of variables, specifically $n = 1000$. The width $w$ ranges from $5$ to $995$ in increments of $5$ (i.e., $5, 10, 15, ..., 995$). The different SAT encodings compared to the $SCL_{\leq 1}$ are detailed in Table \ref{tab:info_ladder_amo}. These encodings include \textit{Pairwise}, \textit{Sequential Counter (Seq)}, \textit{Binary Decision Diagrams} (\textit{BDD}), \textit{Cardinality Network} (\textit{Card}), \textit{Adder Network} (\textit{Adder}), \textit{Product}, and \textit{Duplex}.
Figures \ref{fig:ladder_amo_runtime} and \ref{fig:ladder_amo_runtime_best_3} illustrate the solving times for the various encodings. 

Among the top three encodings, as shown in Figure \ref{fig:ladder_amo_runtime_best_3}, \( SCL_{\leq 1} \) emerged as the most effective, surpassing \( Duplex \), which was previously recognized as an efficient encoding \cite{fazekas2020duplex}. Overall, the results indicate that, in most cases, the \( SCL_{\leq 1} \) encoding significantly outperforms all other SAT encodings in terms of time efficiency. Additionally, the effectiveness of the generated clauses and the necessary auxiliary variables for \( SCL_{\leq 1} \) are discussed earlier in Section \ref{sec:comparison_Ladder_AMO}.

\subsubsection{Solving time of the Ladder AMK constraint}
\label{sec:eval-ladder-amk}
\noindent 
To compare the solving times of \( SCL_{\leq k} \) with other SAT encodings, we employed a dataset of Ladder AMK constraints that included \( n \) variables and a width \( w \). The values for \( n \) were chosen from the range \{100, 200, 300, \ldots, 5000\}, while the values for \( w \) were taken from the set \{10, 20, 50, 100\}. We define \( k \) as \( k = \frac{1}{5}w \) (which means \( k \in \{2, 4, 10, 20\} \)). Each instance was allowed to run for a maximum of 600 seconds. The SAT encodings compared to \( SCL_{\leq k} \) are detailed in Table \ref{tab:info_ladder_amk}, which includes \textit{Pairwise}, \textit{Seq}, \textit{BDD}, \textit{Adder}, and \textit{Card}. \textit{Product} encoding is excluded from this experiment, as it is less efficient for solving AMK constraints. 

\begin{figure*}[ht!]
    \centering
    \begin{minipage}{0.49\textwidth}
        \centering
        \includegraphics[width=\linewidth]{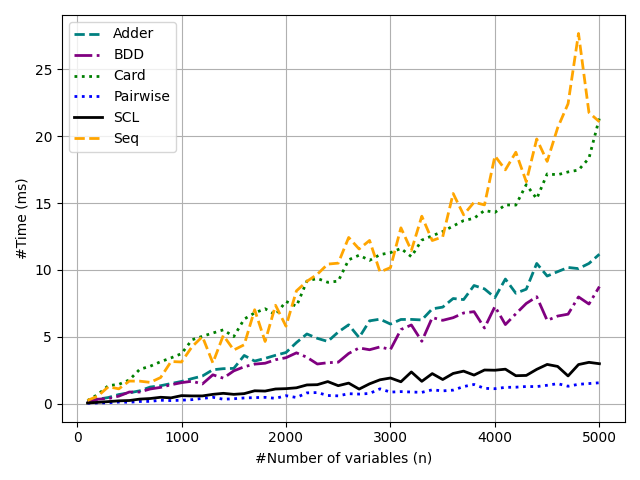}
        \caption{Solving time of Ladder AMK constraints with $w = 10$ and $k = 2$.}
        \label{fig:ladder_amk_runtime_10_2}
    \end{minipage}%
    \hfill
    \begin{minipage}{0.49\textwidth}
        \centering
        \includegraphics[width=\linewidth]{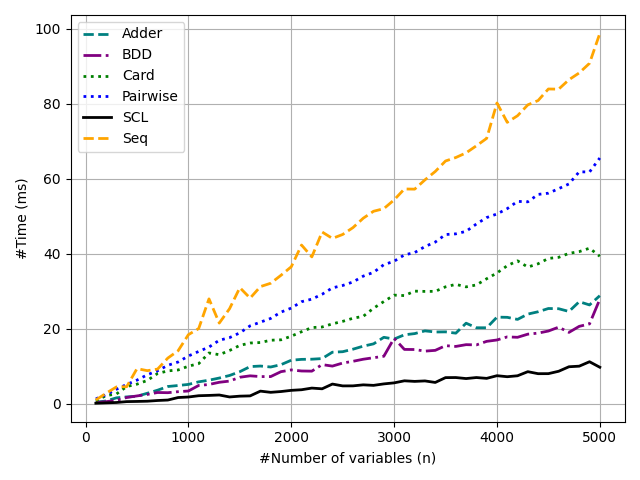}
        \caption{Solving time of Ladder AMK constraints with $w = 20$ and $k = 4$.}
        \label{fig:ladder_amk_runtime_20_4}
    \end{minipage}
\end{figure*}

\begin{figure*}[ht!]
    \centering
    \begin{minipage}{0.49\textwidth}
        \centering
        \includegraphics[width=\linewidth]{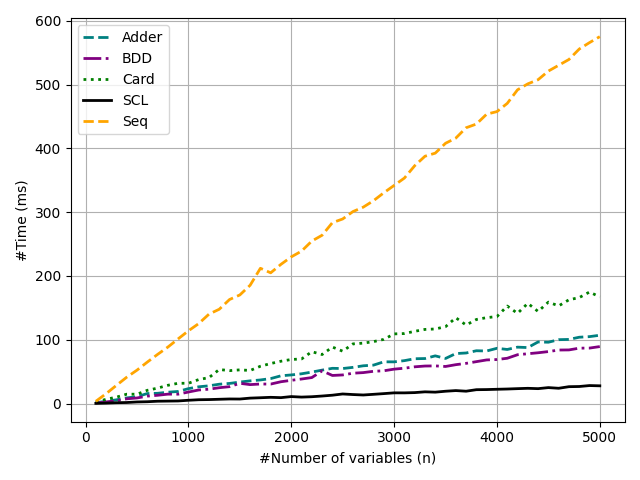}
        \caption{Solving time of Ladder AMK constraints\protect\footref{fn:example} with $w = 50$ and $k = 10$.}
        \label{fig:ladder_amk_runtime_50_10}
    \end{minipage}
    \hfill
    \begin{minipage}{0.49\textwidth}
        \centering
        \includegraphics[width=\linewidth]{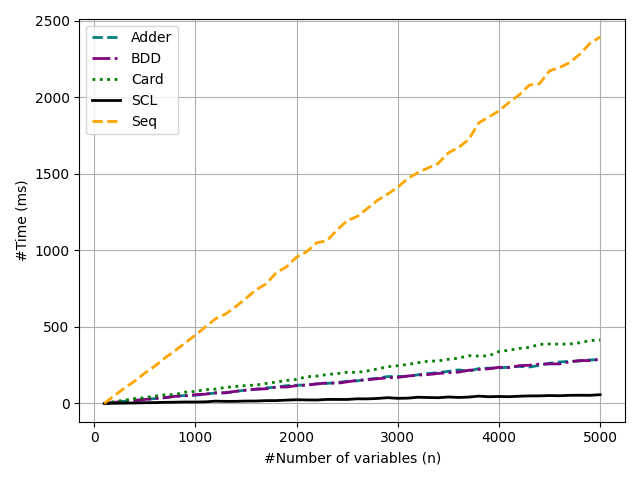}
        \caption{Solving time of Ladder AMK constraints\protect\footref{fn:example} with $w = 100$ and $k = 20$.}
        \label{fig:ladder_amk_runtime_100_20}
    \end{minipage}
\end{figure*}

The solving times of SAT encodings for four values of \( k \) are illustrated in Figures \ref{fig:ladder_amk_runtime_10_2}, \ref{fig:ladder_amk_runtime_20_4}, \ref{fig:ladder_amk_runtime_50_10}, and \ref{fig:ladder_amk_runtime_100_20}. The encoding \( SCL_{\leq k} \) achieves the fastest solving times for \( \{w, k\} = \{20, 4\} , \{50, 10\}, \) and \( \{100, 20\} \), and it ranks second after the \textit{Pairwise} encoding for \( \{w, k\} = \{10, 2\} \).  Compared to \textit{Seq}, \textit{BDD}, \textit{Card}, and \textit{Adder}, \( SCL_{\leq k} \) consistently demonstrates superior performance regarding solving time.

For instance, when \( n = 5000 \), the solving time for \( SCL_{\leq k} \) is approximately one-third of the time required by the second fastest encoding, \textit{BDD}. Specifically, for \( \{w, k\} = \{20,4\} \), \( SCL_{\leq k} \) solves the problem in \( 6.6 \) milliseconds, while \textit{BDD} takes \( 18.1 \) milliseconds. For \( \{w, k\} = \{50,10\} \), \( SCL_{\leq k} \) takes \( 23 \) milliseconds compared to \textit{BDD}'s \( 106 \) milliseconds. For \( \{w, k\} = \{100,20\} \), \( SCL_{\leq k} \) finishes in \( 46.6 \) milliseconds, while \textit{BDD} takes \( 298.4 \) milliseconds. The \textit{Pairwise} encoding is the fastest for \( \{w,k\} = \{10,2\} \) but begins to slow down at \( \{w, k\} = \{20,4\} \). It times out for all input sizes \( n \) within the 600-second limit when solving for \( \{w, k\} = \{50, 10\} \) and \( \{100, 20\} \).

\footnotetext[2]{\label{fn:example}\textit{Pairwise} encoding times out for all input sizes \( n \) within the 600-second limit.}
\setcounter{footnote}{2}

\subsection{Computation Results for the Anti-bandwidth Problem}
\subsubsection{Datasets and Implementation Details}
\label{sec:dataset-implementation-details}

The second framework evaluates the effectiveness of the proposed encoding in addressing the Anti-bandwidth Problem. It utilizes a dataset consisting of 24 matrices from the Harwell-Boeing Sparse Matrix Collection \cite{dataset}. This collection includes 12 small instances and 12 large instances. Detailed information about these 24 matrices can be found in Table \ref{tab:benchmarkData}, which lists the names of the matrices, the number of vertices \(|V|\), and the number of edges \(|E|\). Additionally, the corresponding lower and upper bounds (LB and UB) are provided by Fazekas et al. \cite{fazekas2020duplex}.

\begin{table}[ht!]
\centering
\caption{Harwell-Boeing Sparse Matrix benchmark.}
\begin{tabular}{lcccclllllcccc}
\multicolumn{5}{c}{\textbf{Small instances}} &  &  &  &  & \multicolumn{5}{c}{\textbf{Large instances}} \rule{0pt}{10pt} \\ \cline{1-5} \cline{10-14} 
\textbf{Instance} & \textbf{$|$V$|$} & \textbf{$|$E$|$} & \textbf{LB} & \textbf{UB} &  &  &  &  & \textbf{Instance} & \textbf{$|$V$|$} & \textbf{$|$E$|$} & \textbf{LB} & \textbf{UB} \rule{0pt}{10pt} \\ \cline{1-5} \cline{10-14} 
A-pores\_1 & 30 & 103 & 6 & 8 &  &  &  &  & M-bcsstk06 & 420 & 3720 & 28 & 72 \rule{0pt}{10pt} \\ \cline{1-5} \cline{10-14} 
B-ibm32 & 32 & 90 & 9 & 9 &  &  &  &  & N-bcsstk07 & 420 & 3720 & 28 & 72 \rule{0pt}{10pt} \\ \cline{1-5} \cline{10-14} 
C-bcspwr01 & 39 & 46 & 16 & 17 &  &  &  &  & O-impcol\_d & 425 & 1267 & 91 & 173 \rule{0pt}{10pt} \\ \cline{1-5} \cline{10-14} 
D-bcsstk01 & 48 & 176 & 8 & 9 &  &  &  &  & P-can\_\_445 & 445 & 1682 & 78 & 120 \rule{0pt}{10pt} \\ \cline{1-5} \cline{10-14} 
E-bcspwr02 & 49 & 59 & 21 & 22 &  &  &  &  & Q-494\_bus & 494 & 586 & 219 & 246 \rule{0pt}{10pt} \\ \cline{1-5} \cline{10-14} 
F-curtis54 & 54 & 124 & 12 & 13 &  &  &  &  & R-dwt\_\_503 & 503 & 2762 & 46 & 71 \rule{0pt}{10pt} \\ \cline{1-5} \cline{10-14} 
G-will57 & 57 & 127 & 12 & 14 &  &  &  &  & S-sherman4 & 546 & 1341 & 256 & 272 \rule{0pt}{10pt} \\ \cline{1-5} \cline{10-14} 
H-impcol\_b & 59 & 281 & 8 & 8 &  &  &  &  & T-dwt\_\_592 & 592 & 2256 & 103 & 150 \rule{0pt}{10pt} \\ \cline{1-5} \cline{10-14} 
I-ash85 & 85 & 219 & 19 & 27 &  &  &  &  & U-662\_bus & 662 & 906 & 219 & 220 \rule{0pt}{10pt} \\ \cline{1-5} \cline{10-14} 
J-nos4 & 100 & 247 & 32 & 40 &  &  &  &  & V-nos6 & 675 & 1290 & 326 & 337 \rule{0pt}{10pt} \\ \cline{1-5} \cline{10-14} 
K-dwt\_\_234 & 117 & 162 & 46 & 58 &  &  &  &  & W-685\_bus & 685 & 1282 & 136 & 136 \rule{0pt}{10pt} \\ \cline{1-5} \cline{10-14} 
L-bcspwr03 & 118 & 179 & 39 & 39 &  &  &  &  & X-can\_\_715 & 715 & 2975 & 112 & 142 \rule{0pt}{10pt} \\ \cline{1-5} \cline{10-14} 
\end{tabular}
\label{tab:benchmarkData}
\end{table}

In this framework, we implemented the \( SCL_{\leq 1} \) encoding alongside four SAT-based encodings: \textit{Product}, \textit{Pairwise}, \textit{Sequential Counter}, and \textit{Duplex}. Throughout the implementation, we followed a consistent sequential process. We began by setting the lower bound (LB) as the starting width for the Ladder AMO constraints. If the SAT solver returns a satisfiable (SAT) result, we increase the width by 1 and restart the process. Conversely, if the solver returns unsatisfiable (UNSAT) or reaches the upper bound (UB), it indicates that we have found the optimal solution, and the iteration process concludes.

Constraint Programming (CP) and Mixed Integer Programming (MIP) are included in this experiment, which are \textit{$F_e(k)$}, \textit{CPLEX}, \textit{CP-MZ-Chuffed}, \textit{CP-SAT}, and \textit{Gurobi}. In our experiment, the \textit{$F_e(k)$} encoding leverages the MIP APIs from IBM ILOG CPLEX Optimization Studio\footnote{\url{https://www.ibm.com/products/ilog-cplex-optimization-studio/cplex-optimizer}} version $20.1$, a robust framework that integrates constraint propagation with advanced optimization techniques for efficient problem-solving. The \textit{CPLEX} approach utilizes the CP APIs of IBM ILOG CPLEX Optimization Studio version $22.1.1$. The \textit{CP-MZ-Chuffed} method employs the MiniZinc language \cite{nethercote2007minizinc} version $2.8.6$ and incorporates Chuffed solver\footnote{\url{https://github.com/chuffed/chuffed.git}} version $0.13.2$, which excels in lazy clause generation and is particularly effective for combinatorial problems. The \textit{CP-SAT} approach relies on the CP-SAT Solver, a versatile solver designed for constraint satisfaction and optimization, leveraging multi-processing and SAT-based techniques to handle complex constraints, which is provided by Google OR-Tools\footnote{\url{https://developers.google.com/optimization/cp/cp_solver}} version 9.11.4210. Finally, the \textit{Gurobi} encoding harnesses the Gurobi Optimizer\footnote{\url{https://www.gurobi.com/}} version 11.0.3, a powerful tool for mathematical optimization, known for its efficiency in handling large-scale constraint-based problems. 
Note that \textit{CPLEX}, \textit{Gurobi}, and \textit{CP-SAT} are powerful CP and MIP solvers that have been under development for a long time.

In addition to the sequential implementation, referred to as $SCL_{seq}$, which incrementally solves for anti-bandwidth values from LB to UB, we also implemented a multi-process approach, referred to as \( SCL_{par} \), with the aim of providing a fair comparison with state-of-the-art CP and MIP solvers that exploit multi-core architectures, such as \textit{CPLEX}, \textit{Gurobi}, and \textit{CP-SAT}. This implementation initializes multiple processes, with each process invoking a SAT solver to address a different anti-bandwidth. The multi-process approach is more efficient than the sequential method because if one process finds a solution for a higher anti-bandwidth, it can terminate other processes working on smaller anti-bandwidths, hence allows the search to concentrate on better values. In this experiment, \( SCL_{par} \) is executed with up to four parallel processes invoking the SAT solver.

\subsubsection{Performance Evaluation}

\afterpage{
\newpage
\newgeometry{top=3cm, bottom=3cm, left=3cm, right=3cm}
\thispagestyle{empty}
\begin{landscape}
%
\begin{table*}[ht!] 
\centering
\caption{ABP solving results of different approaches with TO = 1800s and MO = 30GB.}
\resizebox{\linewidth}{!}{
\begin{tabular}{|l|c|c|ccc|ccc|ccc|ccc|ccc|ccc|ccc|ccc|ccc|ccc|ccc|}
\hline
\multirow{4}{*}{Instance} & \multirow{4}{*}{LB} & \multirow{4}{*}{UB} & \multicolumn{3}{c|}{\multirow{2}{*}{\textit{Product}}} & \multicolumn{3}{c|}{\multirow{2}{*}{\textit{Pairwise}}} & \multicolumn{3}{c|}{\multirow{2}{*}{Seq}} & \multicolumn{3}{c|}{\multirow{2}{*}{\textit{Duplex}}} & \multicolumn{3}{c|}{\multirow{2}{*}{$SCL_{seq}$}} & \multicolumn{3}{c|}{\multirow{2}{*}{$SCL_{par}$}} & \multicolumn{3}{c|}{\multirow{2}{*}{$F_e(k)$}} & \multicolumn{3}{c|}{\multirow{2}{*}{\textit{CPLEX}}} & \multicolumn{3}{c|}{\multirow{2}{*}{\textit{CP-MZ-Chuffed}}} & \multicolumn{3}{c|}{\multirow{2}{*}{\textit{CP-SAT}}} & \multicolumn{3}{c|}{\multirow{2}{*}{\textit{Gurobi}}} \\
&  &  & \multicolumn{3}{c|}{} & \multicolumn{3}{c|}{} & \multicolumn{3}{c|}{} & \multicolumn{3}{c|}{} & \multicolumn{3}{c|}{} & \multicolumn{3}{c|}{} & \multicolumn{3}{c|}{} & \multicolumn{3}{c|}{} & \multicolumn{3}{c|}{} & \multicolumn{3}{c|}{} & \multicolumn{3}{c|}{} \\ \cline{4-36} 
&  &  & \multicolumn{1}{c|}{\multirow{2}{*}{Obj-k}} & \multicolumn{1}{c|}{\multirow{2}{*}{Time(s)}} & \multirow{2}{*}{MB} & \multicolumn{1}{c|}{\multirow{2}{*}{Obj-k}} & \multicolumn{1}{c|}{\multirow{2}{*}{Time(s)}} & \multirow{2}{*}{MB} & \multicolumn{1}{c|}{\multirow{2}{*}{Obj-k}} & \multicolumn{1}{c|}{\multirow{2}{*}{Time(s)}} & \multirow{2}{*}{MB} & \multicolumn{1}{c|}{\multirow{2}{*}{Obj-k}} & \multicolumn{1}{c|}{\multirow{2}{*}{Time(s)}} & \multirow{2}{*}{MB} & \multicolumn{1}{c|}{\multirow{2}{*}{Obj-k}} & \multicolumn{1}{c|}{\multirow{2}{*}{Time(s)}} & \multirow{2}{*}{MB} & \multicolumn{1}{c|}{\multirow{2}{*}{Obj-k}} & \multicolumn{1}{c|}{\multirow{2}{*}{Time(s)}} & \multirow{2}{*}{MB} & \multicolumn{1}{c|}{\multirow{2}{*}{Obj-k}} & \multicolumn{1}{c|}{\multirow{2}{*}{Time(s)}} & \multirow{2}{*}{MB} & \multicolumn{1}{c|}{\multirow{2}{*}{Obj-k}} & \multicolumn{1}{c|}{\multirow{2}{*}{Time(s)}} & \multirow{2}{*}{MB} & \multicolumn{1}{c|}{\multirow{2}{*}{Obj-k}} & \multicolumn{1}{c|}{\multirow{2}{*}{Time(s)}} & \multirow{2}{*}{MB} & \multicolumn{1}{c|}{\multirow{2}{*}{Obj-k}} & \multicolumn{1}{c|}{\multirow{2}{*}{Time(s)}} & \multirow{2}{*}{MB} & \multicolumn{1}{c|}{\multirow{2}{*}{Obj-k}} & \multicolumn{1}{c|}{\multirow{2}{*}{Time(s)}} & \multirow{2}{*}{MB} \\
&  &  & \multicolumn{1}{c|}{} & \multicolumn{1}{c|}{} &  & \multicolumn{1}{c|}{} & \multicolumn{1}{c|}{} &  & \multicolumn{1}{c|}{} & \multicolumn{1}{c|}{} &  & \multicolumn{1}{c|}{} & \multicolumn{1}{c|}{} &  & \multicolumn{1}{c|}{} & \multicolumn{1}{c|}{} &  & \multicolumn{1}{c|}{} & \multicolumn{1}{c|}{} &  & \multicolumn{1}{c|}{} & \multicolumn{1}{c|}{} &  & \multicolumn{1}{c|}{} & \multicolumn{1}{c|}{} &  & \multicolumn{1}{c|}{} & \multicolumn{1}{c|}{} &  & \multicolumn{1}{c|}{} & \multicolumn{1}{c|}{} &  & \multicolumn{1}{c|}{} & \multicolumn{1}{c|}{} &  \\ \hline
\multirow{2}{*}{A-pores\_1} & \multirow{2}{*}{6} & \multirow{2}{*}{8} & \multicolumn{1}{c|}{\multirow{2}{*}{\textbf{6}}} & \multicolumn{1}{c|}{\multirow{2}{*}{122}} & \multirow{2}{*}{77.5} & \multicolumn{1}{c|}{\multirow{2}{*}{\textbf{6}}} & \multicolumn{1}{c|}{\multirow{2}{*}{65.7}} & \multirow{2}{*}{56.9} & \multicolumn{1}{c|}{\multirow{2}{*}{\textbf{6}}} & \multicolumn{1}{c|}{\multirow{2}{*}{47.7}} & \multirow{2}{*}{55.4} & \multicolumn{1}{c|}{\multirow{2}{*}{\textbf{6}}} & \multicolumn{1}{c|}{\multirow{2}{*}{\uline{3.4}}} & \multirow{2}{*}{14.1} & \multicolumn{1}{c|}{\multirow{2}{*}{\textbf{6}}} & \multicolumn{1}{c|}{\multirow{2}{*}{9.15}} & \multirow{2}{*}{19.8} & \multicolumn{1}{c|}{\multirow{2}{*}{\textbf{6}}} & \multicolumn{1}{c|}{\multirow{2}{*}{8.8}} & \multirow{2}{*}{23.6} & \multicolumn{1}{c|}{\multirow{2}{*}{\textbf{6}}} & \multicolumn{1}{c|}{\multirow{2}{*}{3.53}} & \multirow{2}{*}{74.9} & \multicolumn{1}{c|}{\multirow{2}{*}{7}} & \multicolumn{1}{c|}{\multirow{2}{*}{TO}} & \multirow{2}{*}{579.3} & \multicolumn{1}{c|}{\multirow{2}{*}{\textbf{6}}} & \multicolumn{1}{c|}{\multirow{2}{*}{4.75}} & \multirow{2}{*}{42.3} & \multicolumn{1}{c|}{\multirow{2}{*}{\textbf{6}}} & \multicolumn{1}{c|}{\multirow{2}{*}{5.17}} & \multirow{2}{*}{144.9} & \multicolumn{1}{c|}{\multirow{2}{*}{\textbf{6}}} & \multicolumn{1}{c|}{\multirow{2}{*}{9.84}} & \multirow{2}{*}{84.9} \\
&  &  & \multicolumn{1}{c|}{} & \multicolumn{1}{c|}{} &  & \multicolumn{1}{c|}{} & \multicolumn{1}{c|}{} &  & \multicolumn{1}{c|}{} & \multicolumn{1}{c|}{} &  & \multicolumn{1}{c|}{} & \multicolumn{1}{c|}{} &  & \multicolumn{1}{c|}{} & \multicolumn{1}{c|}{} &  & \multicolumn{1}{c|}{} & \multicolumn{1}{c|}{} &  & \multicolumn{1}{c|}{} & \multicolumn{1}{c|}{} &  & \multicolumn{1}{c|}{} & \multicolumn{1}{c|}{} &  & \multicolumn{1}{c|}{} & \multicolumn{1}{c|}{} &  & \multicolumn{1}{c|}{} & \multicolumn{1}{c|}{} &  & \multicolumn{1}{c|}{} & \multicolumn{1}{c|}{} &  \\ \hline
\multirow{2}{*}{B-ibm32} & \multirow{2}{*}{9} & \multirow{2}{*}{9} & \multicolumn{1}{c|}{\multirow{2}{*}{\textbf{9}}} & \multicolumn{1}{c|}{\multirow{2}{*}{6.02}} & \multirow{2}{*}{48.9} & \multicolumn{1}{c|}{\multirow{2}{*}{\textbf{9}}} & \multicolumn{1}{c|}{\multirow{2}{*}{18.7}} & \multirow{2}{*}{34.5} & \multicolumn{1}{c|}{\multirow{2}{*}{\textbf{9}}} & \multicolumn{1}{c|}{\multirow{2}{*}{2.13}} & \multirow{2}{*}{42.9} & \multicolumn{1}{c|}{\multirow{2}{*}{\textbf{9}}} & \multicolumn{1}{c|}{\multirow{2}{*}{\uline{0.19}}} & \multirow{2}{*}{8.8} & \multicolumn{1}{c|}{\multirow{2}{*}{\textbf{9}}} & \multicolumn{1}{c|}{\multirow{2}{*}{0.3}} & \multirow{2}{*}{7.8} & \multicolumn{1}{c|}{\multirow{2}{*}{\textbf{9}}} & \multicolumn{1}{c|}{\multirow{2}{*}{0.21}} & \multirow{2}{*}{9.2} & \multicolumn{1}{c|}{\multirow{2}{*}{\textbf{9}}} & \multicolumn{1}{c|}{\multirow{2}{*}{4.57}} & \multirow{2}{*}{106.8} & \multicolumn{1}{c|}{\multirow{2}{*}{\textbf{9}}} & \multicolumn{1}{c|}{\multirow{2}{*}{0.39}} & \multirow{2}{*}{41.3} & \multicolumn{1}{c|}{\multirow{2}{*}{\textbf{9}}} & \multicolumn{1}{c|}{\multirow{2}{*}{1.06}} & \multirow{2}{*}{29.3} & \multicolumn{1}{c|}{\multirow{2}{*}{\textbf{9}}} & \multicolumn{1}{c|}{\multirow{2}{*}{1.38}} & \multirow{2}{*}{120.5} & \multicolumn{1}{c|}{\multirow{2}{*}{\textbf{9}}} & \multicolumn{1}{c|}{\multirow{2}{*}{6.99}} & \multirow{2}{*}{108.2} \\
&  &  & \multicolumn{1}{c|}{} & \multicolumn{1}{c|}{} &  & \multicolumn{1}{c|}{} & \multicolumn{1}{c|}{} &  & \multicolumn{1}{c|}{} & \multicolumn{1}{c|}{} &  & \multicolumn{1}{c|}{} & \multicolumn{1}{c|}{} &  & \multicolumn{1}{c|}{} & \multicolumn{1}{c|}{} &  & \multicolumn{1}{c|}{} & \multicolumn{1}{c|}{} &  & \multicolumn{1}{c|}{} & \multicolumn{1}{c|}{} &  & \multicolumn{1}{c|}{} & \multicolumn{1}{c|}{} &  & \multicolumn{1}{c|}{} & \multicolumn{1}{c|}{} &  & \multicolumn{1}{c|}{} & \multicolumn{1}{c|}{} &  & \multicolumn{1}{c|}{} & \multicolumn{1}{c|}{} &  \\ \hline
\multirow{2}{*}{C-bcspwr01} & \multirow{2}{*}{16} & \multirow{2}{*}{17} & \multicolumn{1}{c|}{\multirow{2}{*}{\textbf{17}}} & \multicolumn{1}{c|}{\multirow{2}{*}{5.25}} & \multirow{2}{*}{41.4} & \multicolumn{1}{c|}{\multirow{2}{*}{\textbf{17}}} & \multicolumn{1}{c|}{\multirow{2}{*}{13}} & \multirow{2}{*}{40.1} & \multicolumn{1}{c|}{\multirow{2}{*}{\textbf{17}}} & \multicolumn{1}{c|}{\multirow{2}{*}{20.7}} & \multirow{2}{*}{58.2} & \multicolumn{1}{c|}{\multirow{2}{*}{\textbf{17}}} & \multicolumn{1}{c|}{\multirow{2}{*}{1.11}} & \multirow{2}{*}{11.2} & \multicolumn{1}{c|}{\multirow{2}{*}{\textbf{17}}} & \multicolumn{1}{c|}{\multirow{2}{*}{4.19}} & \multirow{2}{*}{18.6} & \multicolumn{1}{c|}{\multirow{2}{*}{\textbf{17}}} & \multicolumn{1}{c|}{\multirow{2}{*}{3.44}} & \multirow{2}{*}{16.8} & \multicolumn{1}{c|}{\multirow{2}{*}{\textbf{17}}} & \multicolumn{1}{c|}{\multirow{2}{*}{3.59}} & \multirow{2}{*}{97.5} & \multicolumn{1}{c|}{\multirow{2}{*}{\textbf{17}}} & \multicolumn{1}{c|}{\multirow{2}{*}{0.67}} & \multirow{2}{*}{41.0} & \multicolumn{1}{c|}{\multirow{2}{*}{\textbf{17}}} & \multicolumn{1}{c|}{\multirow{2}{*}{\uline{0.43}}} & \multirow{2}{*}{28.6} & \multicolumn{1}{c|}{\multirow{2}{*}{\textbf{17}}} & \multicolumn{1}{c|}{\multirow{2}{*}{1.46}} & \multirow{2}{*}{126.1} & \multicolumn{1}{c|}{\multirow{2}{*}{\textbf{17}}} & \multicolumn{1}{c|}{\multirow{2}{*}{5.1}} & \multirow{2}{*}{104.6} \\
&  &  & \multicolumn{1}{c|}{} & \multicolumn{1}{c|}{} &  & \multicolumn{1}{c|}{} & \multicolumn{1}{c|}{} &  & \multicolumn{1}{c|}{} & \multicolumn{1}{c|}{} &  & \multicolumn{1}{c|}{} & \multicolumn{1}{c|}{} &  & \multicolumn{1}{c|}{} & \multicolumn{1}{c|}{} &  & \multicolumn{1}{c|}{} & \multicolumn{1}{c|}{} &  & \multicolumn{1}{c|}{} & \multicolumn{1}{c|}{} &  & \multicolumn{1}{c|}{} & \multicolumn{1}{c|}{} &  & \multicolumn{1}{c|}{} & \multicolumn{1}{c|}{} &  & \multicolumn{1}{c|}{} & \multicolumn{1}{c|}{} &  & \multicolumn{1}{c|}{} & \multicolumn{1}{c|}{} &  \\ \hline
\multirow{2}{*}{D-bcsstk01} & \multirow{2}{*}{8} & \multirow{2}{*}{9} & \multicolumn{1}{c|}{\multirow{2}{*}{\textbf{9}}} & \multicolumn{1}{c|}{\multirow{2}{*}{9.43}} & \multirow{2}{*}{140.1} & \multicolumn{1}{c|}{\multirow{2}{*}{\textbf{9}}} & \multicolumn{1}{c|}{\multirow{2}{*}{5.36}} & \multirow{2}{*}{48.2} & \multicolumn{1}{c|}{\multirow{2}{*}{\textbf{9}}} & \multicolumn{1}{c|}{\multirow{2}{*}{11.0}} & \multirow{2}{*}{118.9} & \multicolumn{1}{c|}{\multirow{2}{*}{\textbf{9}}} & \multicolumn{1}{c|}{\multirow{2}{*}{0.54}} & \multirow{2}{*}{14.9} & \multicolumn{1}{c|}{\multirow{2}{*}{\textbf{9}}} & \multicolumn{1}{c|}{\multirow{2}{*}{0.19}} & \multirow{2}{*}{12.6} & \multicolumn{1}{c|}{\multirow{2}{*}{\textbf{9}}} & \multicolumn{1}{c|}{\multirow{2}{*}{\uline{0.13}}} & \multirow{2}{*}{13.2} & \multicolumn{1}{c|}{\multirow{2}{*}{\textbf{9}}} & \multicolumn{1}{c|}{\multirow{2}{*}{14.9}} & \multirow{2}{*}{147.0} & \multicolumn{1}{c|}{\multirow{2}{*}{\textbf{9}}} & \multicolumn{1}{c|}{\multirow{2}{*}{2.28}} & \multirow{2}{*}{46.8} & \multicolumn{1}{c|}{\multirow{2}{*}{\textbf{9}}} & \multicolumn{1}{c|}{\multirow{2}{*}{1.12}} & \multirow{2}{*}{41.0} & \multicolumn{1}{c|}{\multirow{2}{*}{\textbf{9}}} & \multicolumn{1}{c|}{\multirow{2}{*}{1.72}} & \multirow{2}{*}{158.4} & \multicolumn{1}{c|}{\multirow{2}{*}{\textbf{9}}} & \multicolumn{1}{c|}{\multirow{2}{*}{15.9}} & \multirow{2}{*}{252.9} \\
&  &  & \multicolumn{1}{c|}{} & \multicolumn{1}{c|}{} &  & \multicolumn{1}{c|}{} & \multicolumn{1}{c|}{} &  & \multicolumn{1}{c|}{} & \multicolumn{1}{c|}{} &  & \multicolumn{1}{c|}{} & \multicolumn{1}{c|}{} &  & \multicolumn{1}{c|}{} & \multicolumn{1}{c|}{} &  & \multicolumn{1}{c|}{} & \multicolumn{1}{c|}{} &  & \multicolumn{1}{c|}{} & \multicolumn{1}{c|}{} &  & \multicolumn{1}{c|}{} & \multicolumn{1}{c|}{} &  & \multicolumn{1}{c|}{} & \multicolumn{1}{c|}{} &  & \multicolumn{1}{c|}{} & \multicolumn{1}{c|}{} &  & \multicolumn{1}{c|}{} & \multicolumn{1}{c|}{} &  \\ \hline
\multirow{2}{*}{E-bcspwr02} & \multirow{2}{*}{21} & \multirow{2}{*}{22} & \multicolumn{1}{c|}{\multirow{2}{*}{\textbf{21}}} & \multicolumn{1}{c|}{\multirow{2}{*}{8.79}} & \multirow{2}{*}{83.5} & \multicolumn{1}{c|}{\multirow{2}{*}{\textbf{21}}} & \multicolumn{1}{c|}{\multirow{2}{*}{9.22}} & \multirow{2}{*}{45.8} & \multicolumn{1}{c|}{\multirow{2}{*}{\textbf{21}}} & \multicolumn{1}{c|}{\multirow{2}{*}{9.91}} & \multirow{2}{*}{86.1} & \multicolumn{1}{c|}{\multirow{2}{*}{\textbf{21}}} & \multicolumn{1}{c|}{\multirow{2}{*}{2.15}} & \multirow{2}{*}{14.9} & \multicolumn{1}{c|}{\multirow{2}{*}{\textbf{21}}} & \multicolumn{1}{c|}{\multirow{2}{*}{1.52}} & \multirow{2}{*}{11} & \multicolumn{1}{c|}{\multirow{2}{*}{\textbf{21}}} & \multicolumn{1}{c|}{\multirow{2}{*}{\uline{0.74}}} & \multirow{2}{*}{19.3} & \multicolumn{1}{c|}{\multirow{2}{*}{\textbf{21}}} & \multicolumn{1}{c|}{\multirow{2}{*}{17.8}} & \multirow{2}{*}{170.2} & \multicolumn{1}{c|}{\multirow{2}{*}{\textbf{21}}} & \multicolumn{1}{c|}{\multirow{2}{*}{15.83}} & \multirow{2}{*}{51.5} & \multicolumn{1}{c|}{\multirow{2}{*}{\textbf{21}}} & \multicolumn{1}{c|}{\multirow{2}{*}{0.8}} & \multirow{2}{*}{31.5} & \multicolumn{1}{c|}{\multirow{2}{*}{\textbf{21}}} & \multicolumn{1}{c|}{\multirow{2}{*}{3.83}} & \multirow{2}{*}{172.6} & \multicolumn{1}{c|}{\multirow{2}{*}{\textbf{21}}} & \multicolumn{1}{c|}{\multirow{2}{*}{14.3}} & \multirow{2}{*}{198.5} \\
&  &  & \multicolumn{1}{c|}{} & \multicolumn{1}{c|}{} &  & \multicolumn{1}{c|}{} & \multicolumn{1}{c|}{} &  & \multicolumn{1}{c|}{} & \multicolumn{1}{c|}{} &  & \multicolumn{1}{c|}{} & \multicolumn{1}{c|}{} &  & \multicolumn{1}{c|}{} & \multicolumn{1}{c|}{} &  & \multicolumn{1}{c|}{} & \multicolumn{1}{c|}{} &  & \multicolumn{1}{c|}{} & \multicolumn{1}{c|}{} &  & \multicolumn{1}{c|}{} & \multicolumn{1}{c|}{} &  & \multicolumn{1}{c|}{} & \multicolumn{1}{c|}{} &  & \multicolumn{1}{c|}{} & \multicolumn{1}{c|}{} &  & \multicolumn{1}{c|}{} & \multicolumn{1}{c|}{} &  \\ \hline
\multirow{2}{*}{F-curtis54} & \multirow{2}{*}{12} & \multirow{2}{*}{13} & \multicolumn{1}{c|}{\multirow{2}{*}{\textbf{13}}} & \multicolumn{1}{c|}{\multirow{2}{*}{13.6}} & \multirow{2}{*}{141.6} & \multicolumn{1}{c|}{\multirow{2}{*}{\textbf{13}}} & \multicolumn{1}{c|}{\multirow{2}{*}{1.52}} & \multirow{2}{*}{41.3} & \multicolumn{1}{c|}{\multirow{2}{*}{\textbf{13}}} & \multicolumn{1}{c|}{\multirow{2}{*}{11.1}} & \multirow{2}{*}{127} & \multicolumn{1}{c|}{\multirow{2}{*}{\textbf{13}}} & \multicolumn{1}{c|}{\multirow{2}{*}{0.27}} & \multirow{2}{*}{14.1} & \multicolumn{1}{c|}{\multirow{2}{*}{\textbf{13}}} & \multicolumn{1}{c|}{\multirow{2}{*}{0.28}} & \multirow{2}{*}{12.3} & \multicolumn{1}{c|}{\multirow{2}{*}{\textbf{13}}} & \multicolumn{1}{c|}{\multirow{2}{*}{\uline{0.13}}} & \multirow{2}{*}{23.9} & \multicolumn{1}{c|}{\multirow{2}{*}{\textbf{13}}} & \multicolumn{1}{c|}{\multirow{2}{*}{13.3}} & \multirow{2}{*}{145.8} & \multicolumn{1}{c|}{\multirow{2}{*}{\textbf{13}}} & \multicolumn{1}{c|}{\multirow{2}{*}{2.5}} & \multirow{2}{*}{47.2} & \multicolumn{1}{c|}{\multirow{2}{*}{\textbf{13}}} & \multicolumn{1}{c|}{\multirow{2}{*}{1.35}} & \multirow{2}{*}{38.8} & \multicolumn{1}{c|}{\multirow{2}{*}{\textbf{13}}} & \multicolumn{1}{c|}{\multirow{2}{*}{2.6}} & \multirow{2}{*}{179.8} & \multicolumn{1}{c|}{\multirow{2}{*}{\textbf{13}}} & \multicolumn{1}{c|}{\multirow{2}{*}{14.3}} & \multirow{2}{*}{255} \\
&  &  & \multicolumn{1}{c|}{} & \multicolumn{1}{c|}{} &  & \multicolumn{1}{c|}{} & \multicolumn{1}{c|}{} &  & \multicolumn{1}{c|}{} & \multicolumn{1}{c|}{} &  & \multicolumn{1}{c|}{} & \multicolumn{1}{c|}{} &  & \multicolumn{1}{c|}{} & \multicolumn{1}{c|}{} &  & \multicolumn{1}{c|}{} & \multicolumn{1}{c|}{} &  & \multicolumn{1}{c|}{} & \multicolumn{1}{c|}{} &  & \multicolumn{1}{c|}{} & \multicolumn{1}{c|}{} &  & \multicolumn{1}{c|}{} & \multicolumn{1}{c|}{} &  & \multicolumn{1}{c|}{} & \multicolumn{1}{c|}{} &  & \multicolumn{1}{c|}{} & \multicolumn{1}{c|}{} &  \\ \hline
\multirow{2}{*}{G-will57} & \multirow{2}{*}{12} & \multirow{2}{*}{14} & \multicolumn{1}{c|}{\multirow{2}{*}{\textbf{13}}} & \multicolumn{1}{c|}{\multirow{2}{*}{16.6}} & \multirow{2}{*}{147.9} & \multicolumn{1}{c|}{\multirow{2}{*}{\textbf{13}}} & \multicolumn{1}{c|}{\multirow{2}{*}{11.9}} & \multirow{2}{*}{73.2} & \multicolumn{1}{c|}{\multirow{2}{*}{\textbf{13}}} & \multicolumn{1}{c|}{\multirow{2}{*}{23.2}} & \multirow{2}{*}{133.8} & \multicolumn{1}{c|}{\multirow{2}{*}{\textbf{13}}} & \multicolumn{1}{c|}{\multirow{2}{*}{\uline{0.29}}} & \multirow{2}{*}{14.4} & \multicolumn{1}{c|}{\multirow{2}{*}{\textbf{13}}} & \multicolumn{1}{c|}{\multirow{2}{*}{0.47}} & \multirow{2}{*}{13.9} & \multicolumn{1}{c|}{\multirow{2}{*}{\textbf{13}}} & \multicolumn{1}{c|}{\multirow{2}{*}{0.44}} & \multirow{2}{*}{26.5} & \multicolumn{1}{c|}{\multirow{2}{*}{\textbf{13}}} & \multicolumn{1}{c|}{\multirow{2}{*}{18.5}} & \multirow{2}{*}{175.6} & \multicolumn{1}{c|}{\multirow{2}{*}{\textbf{13}}} & \multicolumn{1}{c|}{\multirow{2}{*}{14.18}} & \multirow{2}{*}{51.5} & \multicolumn{1}{c|}{\multirow{2}{*}{\textbf{13}}} & \multicolumn{1}{c|}{\multirow{2}{*}{1.09}} & \multirow{2}{*}{39.9} & \multicolumn{1}{c|}{\multirow{2}{*}{\textbf{13}}} & \multicolumn{1}{c|}{\multirow{2}{*}{2.16}} & \multirow{2}{*}{176.2} & \multicolumn{1}{c|}{\multirow{2}{*}{\textbf{13}}} & \multicolumn{1}{c|}{\multirow{2}{*}{21.5}} & \multirow{2}{*}{271.3} \\
&  &  & \multicolumn{1}{c|}{} & \multicolumn{1}{c|}{} &  & \multicolumn{1}{c|}{} & \multicolumn{1}{c|}{} &  & \multicolumn{1}{c|}{} & \multicolumn{1}{c|}{} &  & \multicolumn{1}{c|}{} & \multicolumn{1}{c|}{} &  & \multicolumn{1}{c|}{} & \multicolumn{1}{c|}{} &  & \multicolumn{1}{c|}{} & \multicolumn{1}{c|}{} &  & \multicolumn{1}{c|}{} & \multicolumn{1}{c|}{} &  & \multicolumn{1}{c|}{} & \multicolumn{1}{c|}{} &  & \multicolumn{1}{c|}{} & \multicolumn{1}{c|}{} &  & \multicolumn{1}{c|}{} & \multicolumn{1}{c|}{} &  & \multicolumn{1}{c|}{} & \multicolumn{1}{c|}{} &  \\ \hline
\multirow{2}{*}{H-impcol\_b} & \multirow{2}{*}{8} & \multirow{2}{*}{8} & \multicolumn{1}{c|}{\multirow{2}{*}{\textbf{8}}} & \multicolumn{1}{c|}{\multirow{2}{*}{4.38}} & \multirow{2}{*}{177} & \multicolumn{1}{c|}{\multirow{2}{*}{\textbf{8}}} & \multicolumn{1}{c|}{\multirow{2}{*}{0.38}} & \multirow{2}{*}{52.8} & \multicolumn{1}{c|}{\multirow{2}{*}{\textbf{8}}} & \multicolumn{1}{c|}{\multirow{2}{*}{5.22}} & \multirow{2}{*}{147.7} & \multicolumn{1}{c|}{\multirow{2}{*}{\textbf{8}}} & \multicolumn{1}{c|}{\multirow{2}{*}{0.26}} & \multirow{2}{*}{21.0} & \multicolumn{1}{c|}{\multirow{2}{*}{\textbf{8}}} & \multicolumn{1}{c|}{\multirow{2}{*}{0.23}} & \multirow{2}{*}{16.8} & \multicolumn{1}{c|}{\multirow{2}{*}{\textbf{8}}} & \multicolumn{1}{c|}{\multirow{2}{*}{\uline{0.08}}} & \multirow{2}{*}{4.8} & \multicolumn{1}{c|}{\multirow{2}{*}{\textbf{8}}} & \multicolumn{1}{c|}{\multirow{2}{*}{3.12}} & \multirow{2}{*}{167.4} & \multicolumn{1}{c|}{\multirow{2}{*}{\textbf{8}}} & \multicolumn{1}{c|}{\multirow{2}{*}{0.32}} & \multirow{2}{*}{44.7} & \multicolumn{1}{c|}{\multirow{2}{*}{\textbf{8}}} & \multicolumn{1}{c|}{\multirow{2}{*}{0.86}} & \multirow{2}{*}{55.5} & \multicolumn{1}{c|}{\multirow{2}{*}{\textbf{8}}} & \multicolumn{1}{c|}{\multirow{2}{*}{2.06}} & \multirow{2}{*}{190.2} & \multicolumn{1}{c|}{\multirow{2}{*}{\textbf{8}}} & \multicolumn{1}{c|}{\multirow{2}{*}{8.17}} & \multirow{2}{*}{262.5} \\
&  &  & \multicolumn{1}{c|}{} & \multicolumn{1}{c|}{} &  & \multicolumn{1}{c|}{} & \multicolumn{1}{c|}{} &  & \multicolumn{1}{c|}{} & \multicolumn{1}{c|}{} &  & \multicolumn{1}{c|}{} & \multicolumn{1}{c|}{} &  & \multicolumn{1}{c|}{} & \multicolumn{1}{c|}{} &  & \multicolumn{1}{c|}{} & \multicolumn{1}{c|}{} &  & \multicolumn{1}{c|}{} & \multicolumn{1}{c|}{} &  & \multicolumn{1}{c|}{} & \multicolumn{1}{c|}{} &  & \multicolumn{1}{c|}{} & \multicolumn{1}{c|}{} &  & \multicolumn{1}{c|}{} & \multicolumn{1}{c|}{} &  & \multicolumn{1}{c|}{} & \multicolumn{1}{c|}{} &  \\ \hline
\multirow{2}{*}{I-ash85} & \multirow{2}{*}{19} & \multirow{2}{*}{27} & \multicolumn{1}{c|}{\multirow{2}{*}{22}} & \multicolumn{1}{c|}{\multirow{2}{*}{TO}} & \multirow{2}{*}{870.8} & \multicolumn{1}{c|}{\multirow{2}{*}{22}} & \multicolumn{1}{c|}{\multirow{2}{*}{TO}} & \multirow{2}{*}{843.6} & \multicolumn{1}{c|}{\multirow{2}{*}{23}} & \multicolumn{1}{c|}{\multirow{2}{*}{TO}} & \multirow{2}{*}{619.8} & \multicolumn{1}{c|}{\multirow{2}{*}{\textbf{24}}} & \multicolumn{1}{c|}{\multirow{2}{*}{TO}} & \multirow{2}{*}{272} & \multicolumn{1}{c|}{\multirow{2}{*}{\textbf{24}}} & \multicolumn{1}{c|}{\multirow{2}{*}{TO}} & \multirow{2}{*}{381} & \multicolumn{1}{c|}{\multirow{2}{*}{\textbf{24}}} & \multicolumn{1}{c|}{\multirow{2}{*}{TO}} & \multirow{2}{*}{496.6} & \multicolumn{1}{c|}{\multirow{2}{*}{22}} & \multicolumn{1}{c|}{\multirow{2}{*}{TO}} & \multirow{2}{*}{460.3} & \multicolumn{1}{c|}{\multirow{2}{*}{23}} & \multicolumn{1}{c|}{\multirow{2}{*}{TO}} & \multirow{2}{*}{288.2} & \multicolumn{1}{c|}{\multirow{2}{*}{23}} & \multicolumn{1}{c|}{\multirow{2}{*}{TO}} & \multirow{2}{*}{374.8} & \multicolumn{1}{c|}{\multirow{2}{*}{\textbf{24}}} & \multicolumn{1}{c|}{\multirow{2}{*}{TO}} & \multirow{2}{*}{1723} & \multicolumn{1}{c|}{\multirow{2}{*}{22}} & \multicolumn{1}{c|}{\multirow{2}{*}{TO}} & \multirow{2}{*}{1219} \\
&  &  & \multicolumn{1}{c|}{} & \multicolumn{1}{c|}{} &  & \multicolumn{1}{c|}{} & \multicolumn{1}{c|}{} &  & \multicolumn{1}{c|}{} & \multicolumn{1}{c|}{} &  & \multicolumn{1}{c|}{} & \multicolumn{1}{c|}{} &  & \multicolumn{1}{c|}{} & \multicolumn{1}{c|}{} &  & \multicolumn{1}{c|}{} & \multicolumn{1}{c|}{} &  & \multicolumn{1}{c|}{} & \multicolumn{1}{c|}{} &  & \multicolumn{1}{c|}{} & \multicolumn{1}{c|}{} &  & \multicolumn{1}{c|}{} & \multicolumn{1}{c|}{} &  & \multicolumn{1}{c|}{} & \multicolumn{1}{c|}{} &  & \multicolumn{1}{c|}{} & \multicolumn{1}{c|}{} &  \\ \hline
\multirow{2}{*}{J-nos4} & \multirow{2}{*}{32} & \multirow{2}{*}{40} & \multicolumn{1}{c|}{\multirow{2}{*}{33}} & \multicolumn{1}{c|}{\multirow{2}{*}{TO}} & \multirow{2}{*}{1119.5} & \multicolumn{1}{c|}{\multirow{2}{*}{33}} & \multicolumn{1}{c|}{\multirow{2}{*}{TO}} & \multirow{2}{*}{1070.1} & \multicolumn{1}{c|}{\multirow{2}{*}{-}} & \multicolumn{1}{c|}{\multirow{2}{*}{TO}} & \multirow{2}{*}{1572.8} & \multicolumn{1}{c|}{\multirow{2}{*}{\textbf{35}}} & \multicolumn{1}{c|}{\multirow{2}{*}{472}} & \multirow{2}{*}{177} & \multicolumn{1}{c|}{\multirow{2}{*}{\textbf{35}}} & \multicolumn{1}{c|}{\multirow{2}{*}{980}} & \multirow{2}{*}{503} & \multicolumn{1}{c|}{\multirow{2}{*}{\textbf{35}}} & \multicolumn{1}{c|}{\multirow{2}{*}{720.3}} & \multirow{2}{*}{505.5} & \multicolumn{1}{c|}{\multirow{2}{*}{-}} & \multicolumn{1}{c|}{\multirow{2}{*}{TO}} & \multirow{2}{*}{927.0} & \multicolumn{1}{c|}{\multirow{2}{*}{35}} & \multicolumn{1}{c|}{\multirow{2}{*}{TO}} & \multirow{2}{*}{272.7} & \multicolumn{1}{c|}{\multirow{2}{*}{\textbf{35}}} & \multicolumn{1}{c|}{\multirow{2}{*}{\uline{384.4}}} & \multirow{2}{*}{343.9} & \multicolumn{1}{c|}{\multirow{2}{*}{36}} & \multicolumn{1}{c|}{\multirow{2}{*}{TO}} & \multirow{2}{*}{1818} & \multicolumn{1}{c|}{\multirow{2}{*}{33}} & \multicolumn{1}{c|}{\multirow{2}{*}{TO}} & \multirow{2}{*}{1690} \\
&  &  & \multicolumn{1}{c|}{} & \multicolumn{1}{c|}{} &  & \multicolumn{1}{c|}{} & \multicolumn{1}{c|}{} &  & \multicolumn{1}{c|}{} & \multicolumn{1}{c|}{} &  & \multicolumn{1}{c|}{} & \multicolumn{1}{c|}{} &  & \multicolumn{1}{c|}{} & \multicolumn{1}{c|}{} &  & \multicolumn{1}{c|}{} & \multicolumn{1}{c|}{} &  & \multicolumn{1}{c|}{} & \multicolumn{1}{c|}{} &  & \multicolumn{1}{c|}{} & \multicolumn{1}{c|}{} &  & \multicolumn{1}{c|}{} & \multicolumn{1}{c|}{} &  & \multicolumn{1}{c|}{} & \multicolumn{1}{c|}{} &  & \multicolumn{1}{c|}{} & \multicolumn{1}{c|}{} &  \\ \hline
\multirow{2}{*}{K-dwt\_\_234} & \multirow{2}{*}{46} & \multirow{2}{*}{58} & \multicolumn{1}{c|}{\multirow{2}{*}{49}} & \multicolumn{1}{c|}{\multirow{2}{*}{TO}} & \multirow{2}{*}{1373.3} & \multicolumn{1}{c|}{\multirow{2}{*}{51}} & \multicolumn{1}{c|}{\multirow{2}{*}{TO}} & \multirow{2}{*}{1192.6} & \multicolumn{1}{c|}{\multirow{2}{*}{48}} & \multicolumn{1}{c|}{\multirow{2}{*}{TO}} & \multirow{2}{*}{1110} & \multicolumn{1}{c|}{\multirow{2}{*}{51}} & \multicolumn{1}{c|}{\multirow{2}{*}{TO}} & \multirow{2}{*}{685} & \multicolumn{1}{c|}{\multirow{2}{*}{51}} & \multicolumn{1}{c|}{\multirow{2}{*}{TO}} & \multirow{2}{*}{530} & \multicolumn{1}{c|}{\multirow{2}{*}{\textbf{52}}} & \multicolumn{1}{c|}{\multirow{2}{*}{TO}} & \multirow{2}{*}{1377} & \multicolumn{1}{c|}{\multirow{2}{*}{50}} & \multicolumn{1}{c|}{\multirow{2}{*}{TO}} & \multirow{2}{*}{895.3} & \multicolumn{1}{c|}{\multirow{2}{*}{\textbf{52}}} & \multicolumn{1}{c|}{\multirow{2}{*}{TO}} & \multirow{2}{*}{243.8} & \multicolumn{1}{c|}{\multirow{2}{*}{47}} & \multicolumn{1}{c|}{\multirow{2}{*}{TO}} & \multirow{2}{*}{244.6} & \multicolumn{1}{c|}{\multirow{2}{*}{51}} & \multicolumn{1}{c|}{\multirow{2}{*}{TO}} & \multirow{2}{*}{1088} & \multicolumn{1}{c|}{\multirow{2}{*}{50}} & \multicolumn{1}{c|}{\multirow{2}{*}{TO}} & \multirow{2}{*}{784.4} \\
&  &  & \multicolumn{1}{c|}{} & \multicolumn{1}{c|}{} &  & \multicolumn{1}{c|}{} & \multicolumn{1}{c|}{} &  & \multicolumn{1}{c|}{} & \multicolumn{1}{c|}{} &  & \multicolumn{1}{c|}{} & \multicolumn{1}{c|}{} &  & \multicolumn{1}{c|}{} & \multicolumn{1}{c|}{} &  & \multicolumn{1}{c|}{} & \multicolumn{1}{c|}{} &  & \multicolumn{1}{c|}{} & \multicolumn{1}{c|}{} &  & \multicolumn{1}{c|}{} & \multicolumn{1}{c|}{} &  & \multicolumn{1}{c|}{} & \multicolumn{1}{c|}{} &  & \multicolumn{1}{c|}{} & \multicolumn{1}{c|}{} &  & \multicolumn{1}{c|}{} & \multicolumn{1}{c|}{} &  \\ \hline
\multirow{2}{*}{L-bcspwr03} & \multirow{2}{*}{39} & \multirow{2}{*}{39} & \multicolumn{1}{c|}{\multirow{2}{*}{\textbf{39}}} & \multicolumn{1}{c|}{\multirow{2}{*}{15.1}} & \multirow{2}{*}{682.1} & \multicolumn{1}{c|}{\multirow{2}{*}{\textbf{39}}} & \multicolumn{1}{c|}{\multirow{2}{*}{3.76}} & \multirow{2}{*}{436.5} & \multicolumn{1}{c|}{\multirow{2}{*}{\textbf{39}}} & \multicolumn{1}{c|}{\multirow{2}{*}{47.7}} & \multirow{2}{*}{856.6} & \multicolumn{1}{c|}{\multirow{2}{*}{\textbf{39}}} & \multicolumn{1}{c|}{\multirow{2}{*}{0.58}} & \multirow{2}{*}{53} & \multicolumn{1}{c|}{\multirow{2}{*}{\textbf{39}}} & \multicolumn{1}{c|}{\multirow{2}{*}{0.69}} & \multirow{2}{*}{40.9} & \multicolumn{1}{c|}{\multirow{2}{*}{\textbf{39}}} & \multicolumn{1}{c|}{\multirow{2}{*}{0.65}} & \multirow{2}{*}{41} & \multicolumn{1}{c|}{\multirow{2}{*}{\textbf{39}}} & \multicolumn{1}{c|}{\multirow{2}{*}{11.15}} & \multirow{2}{*}{559.0} & \multicolumn{1}{c|}{\multirow{2}{*}{\textbf{39}}} & \multicolumn{1}{c|}{\multirow{2}{*}{\uline{0.25}}} & \multirow{2}{*}{48.2} & \multicolumn{1}{c|}{\multirow{2}{*}{\textbf{39}}} & \multicolumn{1}{c|}{\multirow{2}{*}{2.34}} & \multirow{2}{*}{66.6} & \multicolumn{1}{c|}{\multirow{2}{*}{\textbf{39}}} & \multicolumn{1}{c|}{\multirow{2}{*}{7.15}} & \multirow{2}{*}{482.4} & \multicolumn{1}{c|}{\multirow{2}{*}{\textbf{39}}} & \multicolumn{1}{c|}{\multirow{2}{*}{65.9}} & \multirow{2}{*}{321.3} \\
&  &  & \multicolumn{1}{c|}{} & \multicolumn{1}{c|}{} &  & \multicolumn{1}{c|}{} & \multicolumn{1}{c|}{} &  & \multicolumn{1}{c|}{} & \multicolumn{1}{c|}{} &  & \multicolumn{1}{c|}{} & \multicolumn{1}{c|}{} &  & \multicolumn{1}{c|}{} & \multicolumn{1}{c|}{} &  & \multicolumn{1}{c|}{} & \multicolumn{1}{c|}{} &  & \multicolumn{1}{c|}{} & \multicolumn{1}{c|}{} &  & \multicolumn{1}{c|}{} & \multicolumn{1}{c|}{} &  & \multicolumn{1}{c|}{} & \multicolumn{1}{c|}{} &  & \multicolumn{1}{c|}{} & \multicolumn{1}{c|}{} &  & \multicolumn{1}{c|}{} & \multicolumn{1}{c|}{} &  \\ \hline
\hline
\multirow{2}{*}{M-bcsstk06} & \multirow{2}{*}{28} & \multirow{2}{*}{72} & \multicolumn{1}{c|}{\multirow{2}{*}{-}} & \multicolumn{1}{c|}{\multirow{2}{*}{-}} & \multirow{2}{*}{MO} & \multicolumn{1}{c|}{\multirow{2}{*}{31}} & \multicolumn{1}{c|}{\multirow{2}{*}{TO}} & \multirow{2}{*}{22469} & \multicolumn{1}{c|}{\multirow{2}{*}{-}} & \multicolumn{1}{c|}{\multirow{2}{*}{-}} & \multirow{2}{*}{MO} & \multicolumn{1}{c|}{\multirow{2}{*}{\textbf{35}}} & \multicolumn{1}{c|}{\multirow{2}{*}{TO}} & \multirow{2}{*}{1726} & \multicolumn{1}{c|}{\multirow{2}{*}{\textbf{35}}} & \multicolumn{1}{c|}{\multirow{2}{*}{TO}} & \multirow{2}{*}{1398} & \multicolumn{1}{c|}{\multirow{2}{*}{\textbf{35}}} & \multicolumn{1}{c|}{\multirow{2}{*}{TO}} & \multirow{2}{*}{5076} & \multicolumn{1}{c|}{\multirow{2}{*}{-}} & \multicolumn{1}{c|}{\multirow{2}{*}{TO}} & \multirow{2}{*}{12207} & \multicolumn{1}{c|}{\multirow{2}{*}{30}} & \multicolumn{1}{c|}{\multirow{2}{*}{TO}} & \multirow{2}{*}{738.4} & \multicolumn{1}{c|}{\multirow{2}{*}{-}} & \multicolumn{1}{c|}{\multirow{2}{*}{TO}} & \multirow{2}{*}{3016} & \multicolumn{1}{c|}{\multirow{2}{*}{33}} & \multicolumn{1}{c|}{\multirow{2}{*}{TO}} & \multirow{2}{*}{5243} & \multicolumn{1}{c|}{\multirow{2}{*}{-}} & \multicolumn{1}{c|}{\multirow{2}{*}{320.3}} & \multirow{2}{*}{MO} \\
&  &  & \multicolumn{1}{c|}{} & \multicolumn{1}{c|}{} &  & \multicolumn{1}{c|}{} & \multicolumn{1}{c|}{} &  & \multicolumn{1}{c|}{} & \multicolumn{1}{c|}{} &  & \multicolumn{1}{c|}{} & \multicolumn{1}{c|}{} &  & \multicolumn{1}{c|}{} & \multicolumn{1}{c|}{} &  & \multicolumn{1}{c|}{} & \multicolumn{1}{c|}{} &  & \multicolumn{1}{c|}{} & \multicolumn{1}{c|}{} &  & \multicolumn{1}{c|}{} & \multicolumn{1}{c|}{} &  & \multicolumn{1}{c|}{} & \multicolumn{1}{c|}{} &  & \multicolumn{1}{c|}{} & \multicolumn{1}{c|}{} &  & \multicolumn{1}{c|}{} & \multicolumn{1}{c|}{} &  \\ \hline
\multirow{2}{*}{N-bcsstk07} & \multirow{2}{*}{28} & \multirow{2}{*}{72} & \multicolumn{1}{c|}{\multirow{2}{*}{-}} & \multicolumn{1}{c|}{\multirow{2}{*}{-}} & \multirow{2}{*}{MO} & \multicolumn{1}{c|}{\multirow{2}{*}{31}} & \multicolumn{1}{c|}{\multirow{2}{*}{TO}} & \multirow{2}{*}{22470} & \multicolumn{1}{c|}{\multirow{2}{*}{-}} & \multicolumn{1}{c|}{\multirow{2}{*}{-}} & \multirow{2}{*}{MO} & \multicolumn{1}{c|}{\multirow{2}{*}{\textbf{35}}} & \multicolumn{1}{c|}{\multirow{2}{*}{TO}} & \multirow{2}{*}{1726} & \multicolumn{1}{c|}{\multirow{2}{*}{\textbf{35}}} & \multicolumn{1}{c|}{\multirow{2}{*}{TO}} & \multirow{2}{*}{1397} & \multicolumn{1}{c|}{\multirow{2}{*}{\textbf{35}}} & \multicolumn{1}{c|}{\multirow{2}{*}{TO}} & \multirow{2}{*}{5291} & \multicolumn{1}{c|}{\multirow{2}{*}{-}} & \multicolumn{1}{c|}{\multirow{2}{*}{TO}} & \multirow{2}{*}{12244} & \multicolumn{1}{c|}{\multirow{2}{*}{30}} & \multicolumn{1}{c|}{\multirow{2}{*}{TO}} & \multirow{2}{*}{737.9} & \multicolumn{1}{c|}{\multirow{2}{*}{-}} & \multicolumn{1}{c|}{\multirow{2}{*}{TO}} & \multirow{2}{*}{3016} & \multicolumn{1}{c|}{\multirow{2}{*}{33}} & \multicolumn{1}{c|}{\multirow{2}{*}{TO}} & \multirow{2}{*}{5250} & \multicolumn{1}{c|}{\multirow{2}{*}{-}} & \multicolumn{1}{c|}{\multirow{2}{*}{304.1}} & \multirow{2}{*}{MO} \\
&  &  & \multicolumn{1}{c|}{} & \multicolumn{1}{c|}{} &  & \multicolumn{1}{c|}{} & \multicolumn{1}{c|}{} &  & \multicolumn{1}{c|}{} & \multicolumn{1}{c|}{} &  & \multicolumn{1}{c|}{} & \multicolumn{1}{c|}{} &  & \multicolumn{1}{c|}{} & \multicolumn{1}{c|}{} &  & \multicolumn{1}{c|}{} & \multicolumn{1}{c|}{} &  & \multicolumn{1}{c|}{} & \multicolumn{1}{c|}{} &  & \multicolumn{1}{c|}{} & \multicolumn{1}{c|}{} &  & \multicolumn{1}{c|}{} & \multicolumn{1}{c|}{} &  & \multicolumn{1}{c|}{} & \multicolumn{1}{c|}{} &  & \multicolumn{1}{c|}{} & \multicolumn{1}{c|}{} &  \\ \hline
\multirow{2}{*}{O-impcol\_d} & \multirow{2}{*}{91} & \multirow{2}{*}{173} & \multicolumn{1}{c|}{\multirow{2}{*}{-}} & \multicolumn{1}{c|}{\multirow{2}{*}{-}} & \multirow{2}{*}{MO} & \multicolumn{1}{c|}{\multirow{2}{*}{96}} & \multicolumn{1}{c|}{\multirow{2}{*}{TO}} & \multirow{2}{*}{22681} & \multicolumn{1}{c|}{\multirow{2}{*}{-}} & \multicolumn{1}{c|}{\multirow{2}{*}{-}} & \multirow{2}{*}{MO} & \multicolumn{1}{c|}{\multirow{2}{*}{101}} & \multicolumn{1}{c|}{\multirow{2}{*}{TO}} & \multirow{2}{*}{1745} & \multicolumn{1}{c|}{\multirow{2}{*}{102}} & \multicolumn{1}{c|}{\multirow{2}{*}{TO}} & \multirow{2}{*}{1802} & \multicolumn{1}{c|}{\multirow{2}{*}{106}} & \multicolumn{1}{c|}{\multirow{2}{*}{TO}} & \multirow{2}{*}{5366} & \multicolumn{1}{c|}{\multirow{2}{*}{-}} & \multicolumn{1}{c|}{\multirow{2}{*}{-}} & \multirow{2}{*}{MO} & \multicolumn{1}{c|}{\multirow{2}{*}{\textbf{121}}} & \multicolumn{1}{c|}{\multirow{2}{*}{TO}} & \multirow{2}{*}{491.3} & \multicolumn{1}{c|}{\multirow{2}{*}{-}} & \multicolumn{1}{c|}{\multirow{2}{*}{TO}} & \multirow{2}{*}{1410} & \multicolumn{1}{c|}{\multirow{2}{*}{93}} & \multicolumn{1}{c|}{\multirow{2}{*}{TO}} & \multirow{2}{*}{5388} & \multicolumn{1}{c|}{\multirow{2}{*}{-}} & \multicolumn{1}{c|}{\multirow{2}{*}{341}} & \multirow{2}{*}{MO} \\
&  &  & \multicolumn{1}{c|}{} & \multicolumn{1}{c|}{} &  & \multicolumn{1}{c|}{} & \multicolumn{1}{c|}{} &  & \multicolumn{1}{c|}{} & \multicolumn{1}{c|}{} &  & \multicolumn{1}{c|}{} & \multicolumn{1}{c|}{} &  & \multicolumn{1}{c|}{} & \multicolumn{1}{c|}{} &  & \multicolumn{1}{c|}{} & \multicolumn{1}{c|}{} &  & \multicolumn{1}{c|}{} & \multicolumn{1}{c|}{} &  & \multicolumn{1}{c|}{} & \multicolumn{1}{c|}{} &  & \multicolumn{1}{c|}{} & \multicolumn{1}{c|}{} &  & \multicolumn{1}{c|}{} & \multicolumn{1}{c|}{} &  & \multicolumn{1}{c|}{} & \multicolumn{1}{c|}{} &  \\ \hline
\multirow{2}{*}{P-can\_\_445} & \multirow{2}{*}{78} & \multirow{2}{*}{120} & \multicolumn{1}{c|}{\multirow{2}{*}{-}} & \multicolumn{1}{c|}{\multirow{2}{*}{-}} & \multirow{2}{*}{MO} & \multicolumn{1}{c|}{\multirow{2}{*}{-}} & \multicolumn{1}{c|}{\multirow{2}{*}{TO}} & \multirow{2}{*}{27109} & \multicolumn{1}{c|}{\multirow{2}{*}{-}} & \multicolumn{1}{c|}{\multirow{2}{*}{-}} & \multirow{2}{*}{MO} & \multicolumn{1}{c|}{\multirow{2}{*}{-}} & \multicolumn{1}{c|}{\multirow{2}{*}{TO}} & \multirow{2}{*}{2464} & \multicolumn{1}{c|}{\multirow{2}{*}{79}} & \multicolumn{1}{c|}{\multirow{2}{*}{TO}} & \multirow{2}{*}{2156} & \multicolumn{1}{c|}{\multirow{2}{*}{\textbf{81}}} & \multicolumn{1}{c|}{\multirow{2}{*}{TO}} & \multirow{2}{*}{9758} & \multicolumn{1}{c|}{\multirow{2}{*}{-}} & \multicolumn{1}{c|}{\multirow{2}{*}{-}} & \multirow{2}{*}{MO} & \multicolumn{1}{c|}{\multirow{2}{*}{79}} & \multicolumn{1}{c|}{\multirow{2}{*}{TO}} & \multirow{2}{*}{869.2} & \multicolumn{1}{c|}{\multirow{2}{*}{-}} & \multicolumn{1}{c|}{\multirow{2}{*}{TO}} & \multirow{2}{*}{1843} & \multicolumn{1}{c|}{\multirow{2}{*}{-}} & \multicolumn{1}{c|}{\multirow{2}{*}{TO}} & \multirow{2}{*}{5732} & \multicolumn{1}{c|}{\multirow{2}{*}{-}} & \multicolumn{1}{c|}{\multirow{2}{*}{411.6}} & \multirow{2}{*}{MO} \\
&  &  & \multicolumn{1}{c|}{} & \multicolumn{1}{c|}{} &  & \multicolumn{1}{c|}{} & \multicolumn{1}{c|}{} &  & \multicolumn{1}{c|}{} & \multicolumn{1}{c|}{} &  & \multicolumn{1}{c|}{} & \multicolumn{1}{c|}{} &  & \multicolumn{1}{c|}{} & \multicolumn{1}{c|}{} &  & \multicolumn{1}{c|}{} & \multicolumn{1}{c|}{} &  & \multicolumn{1}{c|}{} & \multicolumn{1}{c|}{} &  & \multicolumn{1}{c|}{} & \multicolumn{1}{c|}{} &  & \multicolumn{1}{c|}{} & \multicolumn{1}{c|}{} &  & \multicolumn{1}{c|}{} & \multicolumn{1}{c|}{} &  & \multicolumn{1}{c|}{} & \multicolumn{1}{c|}{} &  \\ \hline
\multirow{2}{*}{Q-494\_bus} & \multirow{2}{*}{219} & \multirow{2}{*}{246} & \multicolumn{1}{c|}{\multirow{2}{*}{-}} & \multicolumn{1}{c|}{\multirow{2}{*}{TO}} & \multirow{2}{*}{25945} & \multicolumn{1}{c|}{\multirow{2}{*}{-}} & \multicolumn{1}{c|}{\multirow{2}{*}{TO}} & \multirow{2}{*}{29391} & \multicolumn{1}{c|}{\multirow{2}{*}{-}} & \multicolumn{1}{c|}{\multirow{2}{*}{-}} & \multirow{2}{*}{MO} & \multicolumn{1}{c|}{\multirow{2}{*}{-}} & \multicolumn{1}{c|}{\multirow{2}{*}{TO}} & \multirow{2}{*}{1443} & \multicolumn{1}{c|}{\multirow{2}{*}{\textbf{220}}} & \multicolumn{1}{c|}{\multirow{2}{*}{TO}} & \multirow{2}{*}{1301} & \multicolumn{1}{c|}{\multirow{2}{*}{\textbf{220}}} & \multicolumn{1}{c|}{\multirow{2}{*}{TO}} & \multirow{2}{*}{4713} & \multicolumn{1}{c|}{\multirow{2}{*}{-}} & \multicolumn{1}{c|}{\multirow{2}{*}{TO}} & \multirow{2}{*}{23775} & \multicolumn{1}{c|}{\multirow{2}{*}{\textbf{220}}} & \multicolumn{1}{c|}{\multirow{2}{*}{TO}} & \multirow{2}{*}{701.6} & \multicolumn{1}{c|}{\multirow{2}{*}{-}} & \multicolumn{1}{c|}{\multirow{2}{*}{TO}} & \multirow{2}{*}{763.8} & \multicolumn{1}{c|}{\multirow{2}{*}{-}} & \multicolumn{1}{c|}{\multirow{2}{*}{316.8}} & \multirow{2}{*}{MO} & \multicolumn{1}{c|}{\multirow{2}{*}{-}} & \multicolumn{1}{c|}{\multirow{2}{*}{TO}} & \multirow{2}{*}{18775} \\
&  &  & \multicolumn{1}{c|}{} & \multicolumn{1}{c|}{} &  & \multicolumn{1}{c|}{} & \multicolumn{1}{c|}{} &  & \multicolumn{1}{c|}{} & \multicolumn{1}{c|}{} &  & \multicolumn{1}{c|}{} & \multicolumn{1}{c|}{} &  & \multicolumn{1}{c|}{} & \multicolumn{1}{c|}{} &  & \multicolumn{1}{c|}{} & \multicolumn{1}{c|}{} &  & \multicolumn{1}{c|}{} & \multicolumn{1}{c|}{} &  & \multicolumn{1}{c|}{} & \multicolumn{1}{c|}{} &  & \multicolumn{1}{c|}{} & \multicolumn{1}{c|}{} &  & \multicolumn{1}{c|}{} & \multicolumn{1}{c|}{} &  & \multicolumn{1}{c|}{} & \multicolumn{1}{c|}{} &  \\ \hline
\multirow{2}{*}{R-dwt\_\_503} & \multirow{2}{*}{46} & \multirow{2}{*}{71} & \multicolumn{1}{c|}{\multirow{2}{*}{-}} & \multicolumn{1}{c|}{\multirow{2}{*}{-}} & \multirow{2}{*}{MO} & \multicolumn{1}{c|}{\multirow{2}{*}{-}} & \multicolumn{1}{c|}{\multirow{2}{*}{-}} & \multirow{2}{*}{MO} & \multicolumn{1}{c|}{\multirow{2}{*}{-}} & \multicolumn{1}{c|}{\multirow{2}{*}{-}} & \multirow{2}{*}{MO} & \multicolumn{1}{c|}{\multirow{2}{*}{\textbf{64}}} & \multicolumn{1}{c|}{\multirow{2}{*}{TO}} & \multirow{2}{*}{1634} & \multicolumn{1}{c|}{\multirow{2}{*}{\textbf{64}}} & \multicolumn{1}{c|}{\multirow{2}{*}{TO}} & \multirow{2}{*}{1474} & \multicolumn{1}{c|}{\multirow{2}{*}{\textbf{64}}} & \multicolumn{1}{c|}{\multirow{2}{*}{TO}} & \multirow{2}{*}{4991} & \multicolumn{1}{c|}{\multirow{2}{*}{-}} & \multicolumn{1}{c|}{\multirow{2}{*}{TO}} & \multirow{2}{*}{10633} & \multicolumn{1}{c|}{\multirow{2}{*}{56}} & \multicolumn{1}{c|}{\multirow{2}{*}{TO}} & \multirow{2}{*}{717.6} & \multicolumn{1}{c|}{\multirow{2}{*}{-}} & \multicolumn{1}{c|}{\multirow{2}{*}{TO}} & \multirow{2}{*}{1950} & \multicolumn{1}{c|}{\multirow{2}{*}{49}} & \multicolumn{1}{c|}{\multirow{2}{*}{TO}} & \multirow{2}{*}{7347} & \multicolumn{1}{c|}{\multirow{2}{*}{-}} & \multicolumn{1}{c|}{\multirow{2}{*}{424.5}} & \multirow{2}{*}{MO} \\
&  &  & \multicolumn{1}{c|}{} & \multicolumn{1}{c|}{} &  & \multicolumn{1}{c|}{} & \multicolumn{1}{c|}{} &  & \multicolumn{1}{c|}{} & \multicolumn{1}{c|}{} &  & \multicolumn{1}{c|}{} & \multicolumn{1}{c|}{} &  & \multicolumn{1}{c|}{} & \multicolumn{1}{c|}{} &  & \multicolumn{1}{c|}{} & \multicolumn{1}{c|}{} &  & \multicolumn{1}{c|}{} & \multicolumn{1}{c|}{} &  & \multicolumn{1}{c|}{} & \multicolumn{1}{c|}{} &  & \multicolumn{1}{c|}{} & \multicolumn{1}{c|}{} &  & \multicolumn{1}{c|}{} & \multicolumn{1}{c|}{} &  & \multicolumn{1}{c|}{} & \multicolumn{1}{c|}{} &  \\ \hline
\multirow{2}{*}{S-sherman4} & \multirow{2}{*}{256} & \multirow{2}{*}{272} & \multicolumn{1}{c|}{\multirow{2}{*}{-}} & \multicolumn{1}{c|}{\multirow{2}{*}{-}} & \multirow{2}{*}{MO} & \multicolumn{1}{c|}{\multirow{2}{*}{-}} & \multicolumn{1}{c|}{\multirow{2}{*}{-}} & \multirow{2}{*}{MO} & \multicolumn{1}{c|}{\multirow{2}{*}{-}} & \multicolumn{1}{c|}{\multirow{2}{*}{-}} & \multirow{2}{*}{MO} & \multicolumn{1}{c|}{\multirow{2}{*}{\textbf{-}}} & \multicolumn{1}{c|}{\multirow{2}{*}{TO}} & \multirow{2}{*}{1326} & \multicolumn{1}{c|}{\multirow{2}{*}{\textbf{-}}} & \multicolumn{1}{c|}{\multirow{2}{*}{TO}} & \multirow{2}{*}{1073} & \multicolumn{1}{c|}{\multirow{2}{*}{\textbf{-}}} & \multicolumn{1}{c|}{\multirow{2}{*}{TO}} & \multirow{2}{*}{4003} & \multicolumn{1}{c|}{\multirow{2}{*}{-}} & \multicolumn{1}{c|}{\multirow{2}{*}{-}} & \multirow{2}{*}{MO} & \multicolumn{1}{c|}{\multirow{2}{*}{-}} & \multicolumn{1}{c|}{\multirow{2}{*}{TO}} & \multirow{2}{*}{689.9} & \multicolumn{1}{c|}{\multirow{2}{*}{-}} & \multicolumn{1}{c|}{\multirow{2}{*}{TO}} & \multirow{2}{*}{888.8} & \multicolumn{1}{c|}{\multirow{2}{*}{-}} & \multicolumn{1}{c|}{\multirow{2}{*}{TO}} & \multirow{2}{*}{29192} & \multicolumn{1}{c|}{\multirow{2}{*}{-}} & \multicolumn{1}{c|}{\multirow{2}{*}{1013}} & \multirow{2}{*}{MO} \\
&  &  & \multicolumn{1}{c|}{} & \multicolumn{1}{c|}{} &  & \multicolumn{1}{c|}{} & \multicolumn{1}{c|}{} &  & \multicolumn{1}{c|}{} & \multicolumn{1}{c|}{} &  & \multicolumn{1}{c|}{} & \multicolumn{1}{c|}{} &  & \multicolumn{1}{c|}{} & \multicolumn{1}{c|}{} &  & \multicolumn{1}{c|}{} & \multicolumn{1}{c|}{} &  & \multicolumn{1}{c|}{} & \multicolumn{1}{c|}{} &  & \multicolumn{1}{c|}{} & \multicolumn{1}{c|}{} &  & \multicolumn{1}{c|}{} & \multicolumn{1}{c|}{} &  & \multicolumn{1}{c|}{} & \multicolumn{1}{c|}{} &  & \multicolumn{1}{c|}{} & \multicolumn{1}{c|}{} &  \\ \hline
\multirow{2}{*}{T-dwt\_\_592} & \multirow{2}{*}{103} & \multirow{2}{*}{150} & \multicolumn{1}{c|}{\multirow{2}{*}{-}} & \multicolumn{1}{c|}{\multirow{2}{*}{-}} & \multirow{2}{*}{MO} & \multicolumn{1}{c|}{\multirow{2}{*}{-}} & \multicolumn{1}{c|}{\multirow{2}{*}{-}} & \multirow{2}{*}{MO} & \multicolumn{1}{c|}{\multirow{2}{*}{-}} & \multicolumn{1}{c|}{\multirow{2}{*}{-}} & \multirow{2}{*}{MO} & \multicolumn{1}{c|}{\multirow{2}{*}{-}} & \multicolumn{1}{c|}{\multirow{2}{*}{TO}} & \multirow{2}{*}{5357} & \multicolumn{1}{c|}{\multirow{2}{*}{\textbf{104}}} & \multicolumn{1}{c|}{\multirow{2}{*}{TO}} & \multirow{2}{*}{2619} & \multicolumn{1}{c|}{\multirow{2}{*}{\textbf{104}}} & \multicolumn{1}{c|}{\multirow{2}{*}{TO}} & \multirow{2}{*}{11445} & \multicolumn{1}{c|}{\multirow{2}{*}{-}} & \multicolumn{1}{c|}{\multirow{2}{*}{-}} & \multirow{2}{*}{MO} & \multicolumn{1}{c|}{\multirow{2}{*}{\textbf{104}}} & \multicolumn{1}{c|}{\multirow{2}{*}{TO}} & \multirow{2}{*}{1366} & \multicolumn{1}{c|}{\multirow{2}{*}{-}} & \multicolumn{1}{c|}{\multirow{2}{*}{TO}} & \multirow{2}{*}{1905} & \multicolumn{1}{c|}{\multirow{2}{*}{-}} & \multicolumn{1}{c|}{\multirow{2}{*}{TO}} & \multirow{2}{*}{9331} & \multicolumn{1}{c|}{\multirow{2}{*}{-}} & \multicolumn{1}{c|}{\multirow{2}{*}{871.3}} & \multirow{2}{*}{MO} \\
&  &  & \multicolumn{1}{c|}{} & \multicolumn{1}{c|}{} &  & \multicolumn{1}{c|}{} & \multicolumn{1}{c|}{} &  & \multicolumn{1}{c|}{} & \multicolumn{1}{c|}{} &  & \multicolumn{1}{c|}{} & \multicolumn{1}{c|}{} &  & \multicolumn{1}{c|}{} & \multicolumn{1}{c|}{} &  & \multicolumn{1}{c|}{} & \multicolumn{1}{c|}{} &  & \multicolumn{1}{c|}{} & \multicolumn{1}{c|}{} &  & \multicolumn{1}{c|}{} & \multicolumn{1}{c|}{} &  & \multicolumn{1}{c|}{} & \multicolumn{1}{c|}{} &  & \multicolumn{1}{c|}{} & \multicolumn{1}{c|}{} &  & \multicolumn{1}{c|}{} & \multicolumn{1}{c|}{} &  \\ \hline
\multirow{2}{*}{U-662\_bus} & \multirow{2}{*}{219} & \multirow{2}{*}{220} & \multicolumn{1}{c|}{\multirow{2}{*}{-}} & \multicolumn{1}{c|}{\multirow{2}{*}{-}} & \multirow{2}{*}{MO} & \multicolumn{1}{c|}{\multirow{2}{*}{-}} & \multicolumn{1}{c|}{\multirow{2}{*}{-}} & \multirow{2}{*}{MO} & \multicolumn{1}{c|}{\multirow{2}{*}{-}} & \multicolumn{1}{c|}{\multirow{2}{*}{-}} & \multirow{2}{*}{MO} & \multicolumn{1}{c|}{\multirow{2}{*}{\textbf{220}}} & \multicolumn{1}{c|}{\multirow{2}{*}{471}} & \multirow{2}{*}{2498} & \multicolumn{1}{c|}{\multirow{2}{*}{\textbf{220}}} & \multicolumn{1}{c|}{\multirow{2}{*}{144}} & \multirow{2}{*}{1377} & \multicolumn{1}{c|}{\multirow{2}{*}{\textbf{220}}} & \multicolumn{1}{c|}{\multirow{2}{*}{122.9}} & \multirow{2}{*}{2600} & \multicolumn{1}{c|}{\multirow{2}{*}{-}} & \multicolumn{1}{c|}{\multirow{2}{*}{-}} & \multirow{2}{*}{MO} & \multicolumn{1}{c|}{\multirow{2}{*}{\textbf{220}}} & \multicolumn{1}{c|}{\multirow{2}{*}{\uline{3.15}}} & \multirow{2}{*}{250.4} & \multicolumn{1}{c|}{\multirow{2}{*}{-}} & \multicolumn{1}{c|}{\multirow{2}{*}{TO}} & \multirow{2}{*}{1061} & \multicolumn{1}{c|}{\multirow{2}{*}{-}} & \multicolumn{1}{c|}{\multirow{2}{*}{328}} & \multirow{2}{*}{MO} & \multicolumn{1}{c|}{\multirow{2}{*}{-}} & \multicolumn{1}{c|}{\multirow{2}{*}{851.3}} & \multirow{2}{*}{MO} \\
&  &  & \multicolumn{1}{c|}{} & \multicolumn{1}{c|}{} &  & \multicolumn{1}{c|}{} & \multicolumn{1}{c|}{} &  & \multicolumn{1}{c|}{} & \multicolumn{1}{c|}{} &  & \multicolumn{1}{c|}{} & \multicolumn{1}{c|}{} &  & \multicolumn{1}{c|}{} & \multicolumn{1}{c|}{} &  & \multicolumn{1}{c|}{} & \multicolumn{1}{c|}{} &  & \multicolumn{1}{c|}{} & \multicolumn{1}{c|}{} &  & \multicolumn{1}{c|}{} & \multicolumn{1}{c|}{} &  & \multicolumn{1}{c|}{} & \multicolumn{1}{c|}{} &  & \multicolumn{1}{c|}{} & \multicolumn{1}{c|}{} &  & \multicolumn{1}{c|}{} & \multicolumn{1}{c|}{} &  \\ \hline
\multirow{2}{*}{V-nos6} & \multirow{2}{*}{326} & \multirow{2}{*}{337} & \multicolumn{1}{c|}{\multirow{2}{*}{-}} & \multicolumn{1}{c|}{\multirow{2}{*}{-}} & \multirow{2}{*}{MO} & \multicolumn{1}{c|}{\multirow{2}{*}{-}} & \multicolumn{1}{c|}{\multirow{2}{*}{-}} & \multirow{2}{*}{MO} & \multicolumn{1}{c|}{\multirow{2}{*}{-}} & \multicolumn{1}{c|}{\multirow{2}{*}{-}} & \multirow{2}{*}{MO} & \multicolumn{1}{c|}{\multirow{2}{*}{-}} & \multicolumn{1}{c|}{\multirow{2}{*}{TO}} & \multirow{2}{*}{1836} & \multicolumn{1}{c|}{\multirow{2}{*}{-}} & \multicolumn{1}{c|}{\multirow{2}{*}{TO}} & \multirow{2}{*}{1532} & \multicolumn{1}{c|}{\multirow{2}{*}{-}} & \multicolumn{1}{c|}{\multirow{2}{*}{TO}} & \multirow{2}{*}{6317} & \multicolumn{1}{c|}{\multirow{2}{*}{-}} & \multicolumn{1}{c|}{\multirow{2}{*}{-}} & \multirow{2}{*}{MO} & \multicolumn{1}{c|}{\multirow{2}{*}{-}} & \multicolumn{1}{c|}{\multirow{2}{*}{TO}} & \multirow{2}{*}{967.9} & \multicolumn{1}{c|}{\multirow{2}{*}{-}} & \multicolumn{1}{c|}{\multirow{2}{*}{TO}} & \multirow{2}{*}{1168} & \multicolumn{1}{c|}{\multirow{2}{*}{-}} & \multicolumn{1}{c|}{\multirow{2}{*}{1260}} & \multirow{2}{*}{MO} & \multicolumn{1}{c|}{\multirow{2}{*}{-}} & \multicolumn{1}{c|}{\multirow{2}{*}{1071}} & \multirow{2}{*}{MO} \\
&  &  & \multicolumn{1}{c|}{} & \multicolumn{1}{c|}{} &  & \multicolumn{1}{c|}{} & \multicolumn{1}{c|}{} &  & \multicolumn{1}{c|}{} & \multicolumn{1}{c|}{} &  & \multicolumn{1}{c|}{} & \multicolumn{1}{c|}{} &  & \multicolumn{1}{c|}{} & \multicolumn{1}{c|}{} &  & \multicolumn{1}{c|}{} & \multicolumn{1}{c|}{} &  & \multicolumn{1}{c|}{} & \multicolumn{1}{c|}{} &  & \multicolumn{1}{c|}{} & \multicolumn{1}{c|}{} &  & \multicolumn{1}{c|}{} & \multicolumn{1}{c|}{} &  & \multicolumn{1}{c|}{} & \multicolumn{1}{c|}{} &  & \multicolumn{1}{c|}{} & \multicolumn{1}{c|}{} &  \\ \hline
\multirow{2}{*}{W-685\_bus} & \multirow{2}{*}{136} & \multirow{2}{*}{136} & \multicolumn{1}{c|}{\multirow{2}{*}{-}} & \multicolumn{1}{c|}{\multirow{2}{*}{-}} & \multirow{2}{*}{MO} & \multicolumn{1}{c|}{\multirow{2}{*}{-}} & \multicolumn{1}{c|}{\multirow{2}{*}{-}} & \multirow{2}{*}{MO} & \multicolumn{1}{c|}{\multirow{2}{*}{-}} & \multicolumn{1}{c|}{\multirow{2}{*}{-}} & \multirow{2}{*}{MO} & \multicolumn{1}{c|}{\multirow{2}{*}{\textbf{136}}} & \multicolumn{1}{c|}{\multirow{2}{*}{10.7}} & \multirow{2}{*}{1431} & \multicolumn{1}{c|}{\multirow{2}{*}{\textbf{136}}} & \multicolumn{1}{c|}{\multirow{2}{*}{7.57}} & \multirow{2}{*}{1164} & \multicolumn{1}{c|}{\multirow{2}{*}{\textbf{136}}} & \multicolumn{1}{c|}{\multirow{2}{*}{7.6}} & \multirow{2}{*}{1162} & \multicolumn{1}{c|}{\multirow{2}{*}{-}} & \multicolumn{1}{c|}{\multirow{2}{*}{-}} & \multirow{2}{*}{MO} & \multicolumn{1}{c|}{\multirow{2}{*}{\textbf{136}}} & \multicolumn{1}{c|}{\multirow{2}{*}{\uline{1.95}}} & \multirow{2}{*}{332.6} & \multicolumn{1}{c|}{\multirow{2}{*}{-}} & \multicolumn{1}{c|}{\multirow{2}{*}{TO}} & \multirow{2}{*}{1325} & \multicolumn{1}{c|}{\multirow{2}{*}{\textbf{136}}} & \multicolumn{1}{c|}{\multirow{2}{*}{73.3}} & \multirow{2}{*}{8223} & \multicolumn{1}{c|}{\multirow{2}{*}{-}} & \multicolumn{1}{c|}{\multirow{2}{*}{855}} & \multirow{2}{*}{MO} \\
&  &  & \multicolumn{1}{c|}{} & \multicolumn{1}{c|}{} &  & \multicolumn{1}{c|}{} & \multicolumn{1}{c|}{} &  & \multicolumn{1}{c|}{} & \multicolumn{1}{c|}{} &  & \multicolumn{1}{c|}{} & \multicolumn{1}{c|}{} &  & \multicolumn{1}{c|}{} & \multicolumn{1}{c|}{} &  & \multicolumn{1}{c|}{} & \multicolumn{1}{c|}{} &  & \multicolumn{1}{c|}{} & \multicolumn{1}{c|}{} &  & \multicolumn{1}{c|}{} & \multicolumn{1}{c|}{} &  & \multicolumn{1}{c|}{} & \multicolumn{1}{c|}{} &  & \multicolumn{1}{c|}{} & \multicolumn{1}{c|}{} &  & \multicolumn{1}{c|}{} & \multicolumn{1}{c|}{} &  \\ \hline
\multirow{2}{*}{X-can\_\_715} & \multirow{2}{*}{112} & \multirow{2}{*}{142} & \multicolumn{1}{c|}{\multirow{2}{*}{-}} & \multicolumn{1}{c|}{\multirow{2}{*}{-}} & \multirow{2}{*}{MO} & \multicolumn{1}{c|}{\multirow{2}{*}{-}} & \multicolumn{1}{c|}{\multirow{2}{*}{-}} & \multirow{2}{*}{MO} & \multicolumn{1}{c|}{\multirow{2}{*}{-}} & \multicolumn{1}{c|}{\multirow{2}{*}{-}} & \multirow{2}{*}{MO} & \multicolumn{1}{c|}{\multirow{2}{*}{-}} & \multicolumn{1}{c|}{\multirow{2}{*}{TO}} & \multirow{2}{*}{3060} & \multicolumn{1}{c|}{\multirow{2}{*}{113}} & \multicolumn{1}{c|}{\multirow{2}{*}{TO}} & \multirow{2}{*}{4599} & \multicolumn{1}{c|}{\multirow{2}{*}{\textbf{116}}} & \multicolumn{1}{c|}{\multirow{2}{*}{TO}} & \multirow{2}{*}{10388} & \multicolumn{1}{c|}{\multirow{2}{*}{-}} & \multicolumn{1}{c|}{\multirow{2}{*}{-}} & \multirow{2}{*}{MO} & \multicolumn{1}{c|}{\multirow{2}{*}{113}} & \multicolumn{1}{c|}{\multirow{2}{*}{TO}} & \multirow{2}{*}{1380} & \multicolumn{1}{c|}{\multirow{2}{*}{-}} & \multicolumn{1}{c|}{\multirow{2}{*}{TO}} & \multirow{2}{*}{2442} & \multicolumn{1}{c|}{\multirow{2}{*}{-}} & \multicolumn{1}{c|}{\multirow{2}{*}{TO}} & \multirow{2}{*}{11514} & \multicolumn{1}{c|}{\multirow{2}{*}{-}} & \multicolumn{1}{c|}{\multirow{2}{*}{1013}} & \multirow{2}{*}{MO} \\
&  &  & \multicolumn{1}{c|}{} & \multicolumn{1}{c|}{} &  & \multicolumn{1}{c|}{} & \multicolumn{1}{c|}{} &  & \multicolumn{1}{c|}{} & \multicolumn{1}{c|}{} &  & \multicolumn{1}{c|}{} & \multicolumn{1}{c|}{} &  & \multicolumn{1}{c|}{} & \multicolumn{1}{c|}{} &  & \multicolumn{1}{c|}{} & \multicolumn{1}{c|}{} &  & \multicolumn{1}{c|}{} & \multicolumn{1}{c|}{} &  & \multicolumn{1}{c|}{} & \multicolumn{1}{c|}{} &  & \multicolumn{1}{c|}{} & \multicolumn{1}{c|}{} &  & \multicolumn{1}{c|}{} & \multicolumn{1}{c|}{} &  & \multicolumn{1}{c|}{} & \multicolumn{1}{c|}{} &  \\ \hline
\end{tabular}
}
\label{tab:resultABP}
\end{table*}
\end{landscape}
\newpage
\restoregeometry
}

Table \ref{tab:resultABP} summarizes our experimental results on 24 selected matrices, with a time limit of 1800 seconds and a memory limit of 30 GB applied to the \textit{Product}, \textit{Pairwise}, \textit{Sequential Counter} (Seq), \textit{Duplex}, \( SCL_{seq} \), \( SCL_{par} \), \( F_e(k) \), \textit{CPLEX}, \textit{CP-MZ-Chuffed}, \textit{CP-SAT}, and \textit{Gurobi} approaches. For each problem, the table displays the best solution found by each method, along with the solving time (in seconds) and memory consumption (in MB). The best anti-bandwidth value is highlighted in bold, and the best solving time is underlined.
If the computation time exceeds 1800 seconds or the memory usage surpasses 30 GB, a termination signal is triggered, resulting in the process stopping due to a timeout (TO) or memory overload (MO), respectively. In such cases, the highest solved anti-bandwidth is reported as the best result achieved by the technique. If the method fails to solve the problem using the initial lower bound (LB) value, the result is indicated with a ``-'' character.

\begin{figure*}[ht!]
    \centering
    \begin{minipage}{0.49\textwidth}
        \centering
        \includegraphics[width=\textwidth]{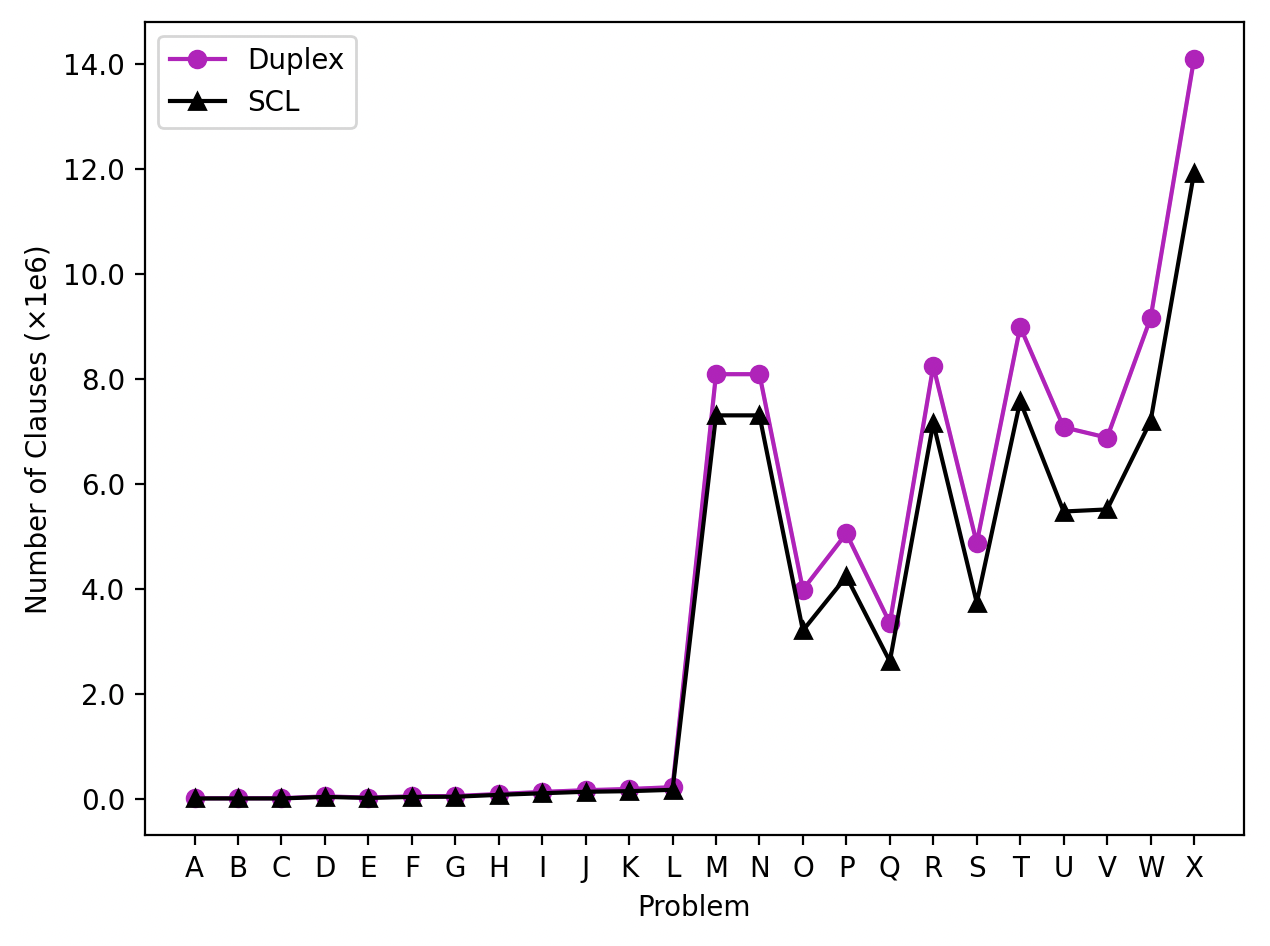}
        \caption{Comparison of generated clauses between \textit{SCL} and \textit{Duplex}.}
        \label{fig:num_clauses}
    \end{minipage}%
    \hfill
    \begin{minipage}{0.49\textwidth}
        \centering
        \includegraphics[width=\textwidth]{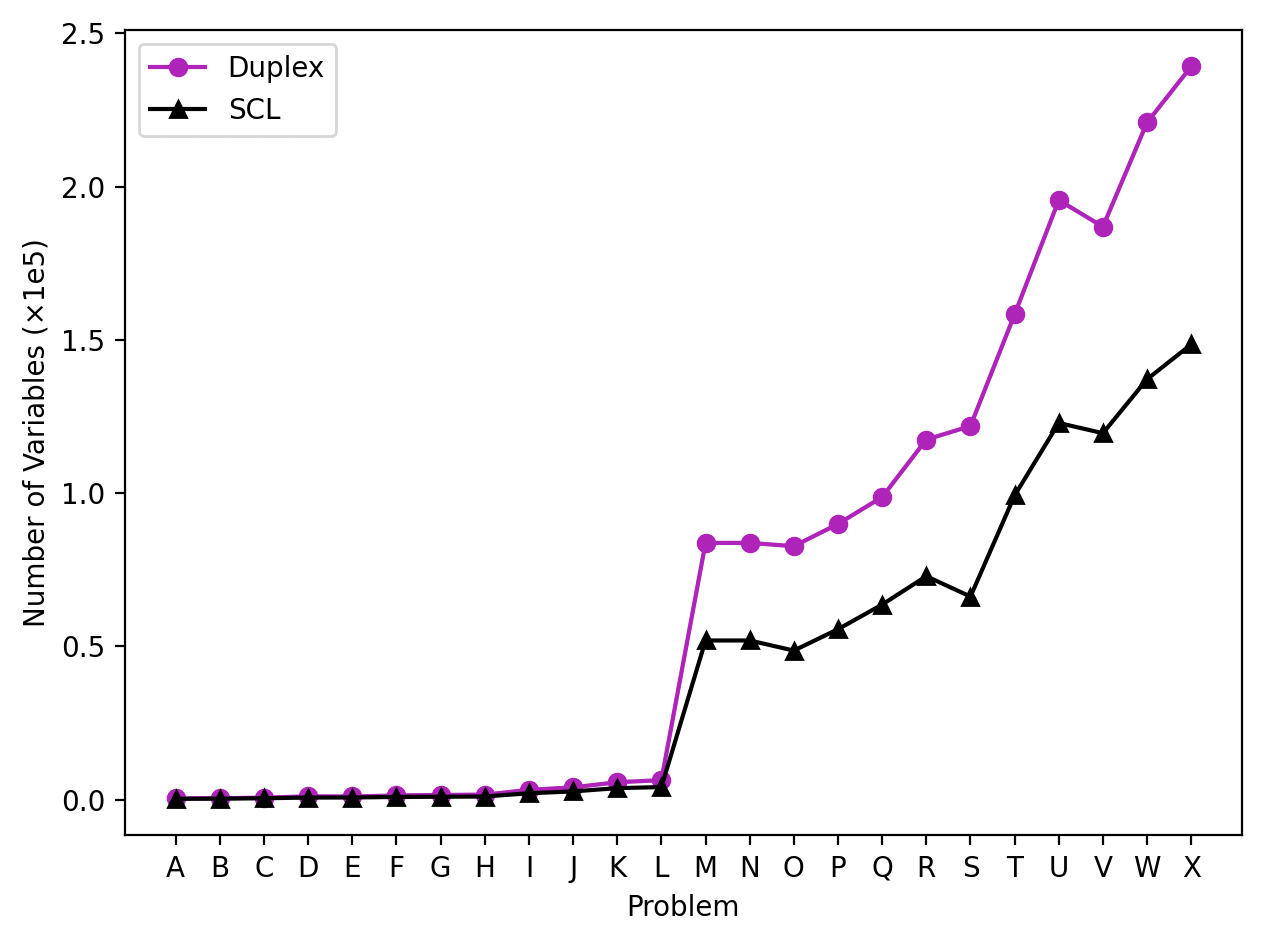}
        \caption{Comparison of auxiliary variables between \textit{SCL} and \textit{Duplex}.}
        \label{fig:num_vars}
    \end{minipage}
\end{figure*}

Figures \ref{fig:num_clauses} and \ref{fig:num_vars} compare the number of generated clauses and auxiliary variables used in the ABP encoding across 24 instances. This evaluation considers both the \textit{SCL} and \textit{Duplex} encodings, with \textit{Duplex} recognized as the best encoding according to \cite{fazekas2020duplex}. Each instance is represented by the first letter of its name (e.g., \textit{A} for \textit{A-pores\_1} and \textit{X} for \textit{X-can\_\_715)}. We analyzed the encodings at the maximum anti-bandwidth achievable by both encodings within specified time and memory limits. The results show that as the size of the instance increases, indicated by the increasing number of vertices $|V|$ and edges $|E|$, the $SCL$ encoding requires fewer clauses and auxiliary variables than the \textit{Duplex} encoding. This finding highlights why $SCL$ can outperform \textit{Duplex} in solving certain ABP instances, particularly in 12 large instances, and it also demonstrates more efficient memory usage.

When solving 12 small graph instances, the $SCL$ encoding yields the best results compared to other methods. The performance of $SCL_{seq}$ is competitive with \textit{Duplex} and outperforms the other SAT-based encodings.
Compared to \textit{CPLEX}, which is one of the top-performing methods among CP and MIP approaches, $SCL_{par}$ finds the optimal anti-bandwidth values for the two instances \textit{A-pores\_1} and \textit{J-nos4}, which are 6 and 35, respectively. In contrast, \textit{CPLEX} only proves feasible anti-bandwidths of 6 and 34 for those instances. Moreover, $SCL_{par}$ also performs better on the \textit{I-ash85} instance.
Compared to the other efficient CP method, \textit{CP-SAT}, $SCL_{par}$ delivers better results on the instances \textit{J-nos4} and \textit{K-dwt\_\_234}. MIP-based methods such as \textit{Gurobi} and \( F_e(k) \) exhibit a weaker performance than $SCL_{par}$.

\begin{table*}[ht!]
\centering
\caption{ABP solving results of the five best approaches in 12 large instances with TO = 1800s and MO = 30GB.}
\resizebox{\textwidth}{!}{
\begin{tabular}{|l|c|c|ccc|ccc|ccc|ccc|ccc|}
\hline
\multirow{4}{*}{Instance} & \multirow{4}{*}{LB} & \multirow{4}{*}{UB} & \multicolumn{3}{c|}{\multirow{2}{*}{\textit{Duplex}}} & \multicolumn{3}{c|}{\multirow{2}{*}{$SCL_{seq}$}} & \multicolumn{3}{c|}{\multirow{2}{*}{$SCL_{par}$}} & \multicolumn{3}{c|}{\multirow{2}{*}{\textit{CPLEX}}} & \multicolumn{3}{c|}{\multirow{2}{*}{\textit{CP-SAT}}} \\
 &  &  & \multicolumn{3}{c|}{} & \multicolumn{3}{c|}{} & \multicolumn{3}{c|}{} & \multicolumn{3}{c|}{} & \multicolumn{3}{c|}{} \\ \cline{4-18} 
 &  &  & \multicolumn{1}{c|}{\multirow{2}{*}{Obj-k}} & \multicolumn{1}{c|}{\multirow{2}{*}{Time(s)}} & \multirow{2}{*}{MB} & \multicolumn{1}{c|}{\multirow{2}{*}{Obj-k}} & \multicolumn{1}{c|}{\multirow{2}{*}{Time(s)}} & \multirow{2}{*}{MB} & \multicolumn{1}{c|}{\multirow{2}{*}{Obj-k}} & \multicolumn{1}{c|}{\multirow{2}{*}{Time(s)}} & \multirow{2}{*}{MB} & \multicolumn{1}{c|}{\multirow{2}{*}{Obj-k}} & \multicolumn{1}{c|}{\multirow{2}{*}{Time(s)}} & \multirow{2}{*}{MB} & \multicolumn{1}{c|}{\multirow{2}{*}{Obj-k}} & \multicolumn{1}{c|}{\multirow{2}{*}{Time(s)}} & \multirow{2}{*}{MB} \\
 &  &  & \multicolumn{1}{c|}{} & \multicolumn{1}{c|}{} &  & \multicolumn{1}{c|}{} & \multicolumn{1}{c|}{} &  & \multicolumn{1}{c|}{} & \multicolumn{1}{c|}{} &  & \multicolumn{1}{c|}{} & \multicolumn{1}{c|}{} &  & \multicolumn{1}{c|}{} & \multicolumn{1}{c|}{} &  \\ \hline
\multirow{2}{*}{M-bcsstk06} & \multirow{2}{*}{28} & \multirow{2}{*}{72} & \multicolumn{1}{c|}{\multirow{2}{*}{\textbf{35}}} & \multicolumn{1}{c|}{\multirow{2}{*}{TO}} & \multirow{2}{*}{1726} & \multicolumn{1}{c|}{\multirow{2}{*}{\textbf{35}}} & \multicolumn{1}{c|}{\multirow{2}{*}{TO}} & \multirow{2}{*}{1398} & \multicolumn{1}{c|}{\multirow{2}{*}{\textbf{35}}} & \multicolumn{1}{c|}{\multirow{2}{*}{TO}} & \multirow{2}{*}{5076} & \multicolumn{1}{c|}{\multirow{2}{*}{30}} & \multicolumn{1}{c|}{\multirow{2}{*}{TO}} & \multirow{2}{*}{738.4} & \multicolumn{1}{c|}{\multirow{2}{*}{33}} & \multicolumn{1}{c|}{\multirow{2}{*}{TO}} & \multirow{2}{*}{5243} \\
 &  &  & \multicolumn{1}{c|}{} & \multicolumn{1}{c|}{} &  & \multicolumn{1}{c|}{} & \multicolumn{1}{c|}{} &  & \multicolumn{1}{c|}{} & \multicolumn{1}{c|}{} &  & \multicolumn{1}{c|}{} & \multicolumn{1}{c|}{} &  & \multicolumn{1}{c|}{} & \multicolumn{1}{c|}{} &  \\ \hline
\multirow{2}{*}{N-bcsstk07} & \multirow{2}{*}{28} & \multirow{2}{*}{72} & \multicolumn{1}{c|}{\multirow{2}{*}{\textbf{35}}} & \multicolumn{1}{c|}{\multirow{2}{*}{TO}} & \multirow{2}{*}{1726} & \multicolumn{1}{c|}{\multirow{2}{*}{\textbf{35}}} & \multicolumn{1}{c|}{\multirow{2}{*}{TO}} & \multirow{2}{*}{1397} & \multicolumn{1}{c|}{\multirow{2}{*}{\textbf{35}}} & \multicolumn{1}{c|}{\multirow{2}{*}{TO}} & \multirow{2}{*}{5291} & \multicolumn{1}{c|}{\multirow{2}{*}{30}} & \multicolumn{1}{c|}{\multirow{2}{*}{TO}} & \multirow{2}{*}{737.9} & \multicolumn{1}{c|}{\multirow{2}{*}{33}} & \multicolumn{1}{c|}{\multirow{2}{*}{TO}} & \multirow{2}{*}{5250} \\
 &  &  & \multicolumn{1}{c|}{} & \multicolumn{1}{c|}{} &  & \multicolumn{1}{c|}{} & \multicolumn{1}{c|}{} &  & \multicolumn{1}{c|}{} & \multicolumn{1}{c|}{} &  & \multicolumn{1}{c|}{} & \multicolumn{1}{c|}{} &  & \multicolumn{1}{c|}{} & \multicolumn{1}{c|}{} &  \\ \hline
\multirow{2}{*}{O-impcol\_d} & \multirow{2}{*}{91} & \multirow{2}{*}{173} & \multicolumn{1}{c|}{\multirow{2}{*}{101}} & \multicolumn{1}{c|}{\multirow{2}{*}{TO}} & \multirow{2}{*}{1745} & \multicolumn{1}{c|}{\multirow{2}{*}{102}} & \multicolumn{1}{c|}{\multirow{2}{*}{TO}} & \multirow{2}{*}{1802} & \multicolumn{1}{c|}{\multirow{2}{*}{106}} & \multicolumn{1}{c|}{\multirow{2}{*}{TO}} & \multirow{2}{*}{5366} & \multicolumn{1}{c|}{\multirow{2}{*}{\textbf{121}}} & \multicolumn{1}{c|}{\multirow{2}{*}{TO}} & \multirow{2}{*}{491.3} & \multicolumn{1}{c|}{\multirow{2}{*}{93}} & \multicolumn{1}{c|}{\multirow{2}{*}{TO}} & \multirow{2}{*}{5388} \\
 &  &  & \multicolumn{1}{c|}{} & \multicolumn{1}{c|}{} &  & \multicolumn{1}{c|}{} & \multicolumn{1}{c|}{} &  & \multicolumn{1}{c|}{} & \multicolumn{1}{c|}{} &  & \multicolumn{1}{c|}{} & \multicolumn{1}{c|}{} &  & \multicolumn{1}{c|}{} & \multicolumn{1}{c|}{} &  \\ \hline
\multirow{2}{*}{P-can\_\_445} & \multirow{2}{*}{78} & \multirow{2}{*}{120} & \multicolumn{1}{c|}{\multirow{2}{*}{-}} & \multicolumn{1}{c|}{\multirow{2}{*}{TO}} & \multirow{2}{*}{2464} & \multicolumn{1}{c|}{\multirow{2}{*}{79}} & \multicolumn{1}{c|}{\multirow{2}{*}{TO}} & \multirow{2}{*}{2156} & \multicolumn{1}{c|}{\multirow{2}{*}{\textbf{81}}} & \multicolumn{1}{c|}{\multirow{2}{*}{TO}} & \multirow{2}{*}{9758} & \multicolumn{1}{c|}{\multirow{2}{*}{79}} & \multicolumn{1}{c|}{\multirow{2}{*}{TO}} & \multirow{2}{*}{869.2} & \multicolumn{1}{c|}{\multirow{2}{*}{-}} & \multicolumn{1}{c|}{\multirow{2}{*}{TO}} & \multirow{2}{*}{5732} \\
 &  &  & \multicolumn{1}{c|}{} & \multicolumn{1}{c|}{} &  & \multicolumn{1}{c|}{} & \multicolumn{1}{c|}{} &  & \multicolumn{1}{c|}{} & \multicolumn{1}{c|}{} &  & \multicolumn{1}{c|}{} & \multicolumn{1}{c|}{} &  & \multicolumn{1}{c|}{} & \multicolumn{1}{c|}{} &  \\ \hline
\multirow{2}{*}{Q-494\_bus} & \multirow{2}{*}{219} & \multirow{2}{*}{246} & \multicolumn{1}{c|}{\multirow{2}{*}{-}} & \multicolumn{1}{c|}{\multirow{2}{*}{TO}} & \multirow{2}{*}{1443} & \multicolumn{1}{c|}{\multirow{2}{*}{\textbf{220}}} & \multicolumn{1}{c|}{\multirow{2}{*}{TO}} & \multirow{2}{*}{1301} & \multicolumn{1}{c|}{\multirow{2}{*}{\textbf{220}}} & \multicolumn{1}{c|}{\multirow{2}{*}{TO}} & \multirow{2}{*}{4713} & \multicolumn{1}{c|}{\multirow{2}{*}{\textbf{220}}} & \multicolumn{1}{c|}{\multirow{2}{*}{TO}} & \multirow{2}{*}{701.6} & \multicolumn{1}{c|}{\multirow{2}{*}{-}} & \multicolumn{1}{c|}{\multirow{2}{*}{316.8}} & \multirow{2}{*}{MO} \\
 &  &  & \multicolumn{1}{c|}{} & \multicolumn{1}{c|}{} &  & \multicolumn{1}{c|}{} & \multicolumn{1}{c|}{} &  & \multicolumn{1}{c|}{} & \multicolumn{1}{c|}{} &  & \multicolumn{1}{c|}{} & \multicolumn{1}{c|}{} &  & \multicolumn{1}{c|}{} & \multicolumn{1}{c|}{} &  \\ \hline
\multirow{2}{*}{R-dwt\_\_503} & \multirow{2}{*}{46} & \multirow{2}{*}{71} & \multicolumn{1}{c|}{\multirow{2}{*}{\textbf{64}}} & \multicolumn{1}{c|}{\multirow{2}{*}{TO}} & \multirow{2}{*}{1634} & \multicolumn{1}{c|}{\multirow{2}{*}{\textbf{64}}} & \multicolumn{1}{c|}{\multirow{2}{*}{TO}} & \multirow{2}{*}{1474} & \multicolumn{1}{c|}{\multirow{2}{*}{\textbf{64}}} & \multicolumn{1}{c|}{\multirow{2}{*}{TO}} & \multirow{2}{*}{4991} & \multicolumn{1}{c|}{\multirow{2}{*}{56}} & \multicolumn{1}{c|}{\multirow{2}{*}{TO}} & \multirow{2}{*}{717.6} & \multicolumn{1}{c|}{\multirow{2}{*}{49}} & \multicolumn{1}{c|}{\multirow{2}{*}{TO}} & \multirow{2}{*}{7347} \\
 &  &  & \multicolumn{1}{c|}{} & \multicolumn{1}{c|}{} &  & \multicolumn{1}{c|}{} & \multicolumn{1}{c|}{} &  & \multicolumn{1}{c|}{} & \multicolumn{1}{c|}{} &  & \multicolumn{1}{c|}{} & \multicolumn{1}{c|}{} &  & \multicolumn{1}{c|}{} & \multicolumn{1}{c|}{} &  \\ \hline
\multirow{2}{*}{S-sherman4} & \multirow{2}{*}{256} & \multirow{2}{*}{272} & \multicolumn{1}{c|}{\multirow{2}{*}{\textbf{-}}} & \multicolumn{1}{c|}{\multirow{2}{*}{TO}} & \multirow{2}{*}{1326} & \multicolumn{1}{c|}{\multirow{2}{*}{\textbf{-}}} & \multicolumn{1}{c|}{\multirow{2}{*}{TO}} & \multirow{2}{*}{1073} & \multicolumn{1}{c|}{\multirow{2}{*}{\textbf{-}}} & \multicolumn{1}{c|}{\multirow{2}{*}{TO}} & \multirow{2}{*}{4003} & \multicolumn{1}{c|}{\multirow{2}{*}{-}} & \multicolumn{1}{c|}{\multirow{2}{*}{TO}} & \multirow{2}{*}{689.9} & \multicolumn{1}{c|}{\multirow{2}{*}{-}} & \multicolumn{1}{c|}{\multirow{2}{*}{TO}} & \multirow{2}{*}{29192} \\
 &  &  & \multicolumn{1}{c|}{} & \multicolumn{1}{c|}{} &  & \multicolumn{1}{c|}{} & \multicolumn{1}{c|}{} &  & \multicolumn{1}{c|}{} & \multicolumn{1}{c|}{} &  & \multicolumn{1}{c|}{} & \multicolumn{1}{c|}{} &  & \multicolumn{1}{c|}{} & \multicolumn{1}{c|}{} &  \\ \hline
\multirow{2}{*}{T-dwt\_\_592} & \multirow{2}{*}{103} & \multirow{2}{*}{150} & \multicolumn{1}{c|}{\multirow{2}{*}{-}} & \multicolumn{1}{c|}{\multirow{2}{*}{TO}} & \multirow{2}{*}{5357} & \multicolumn{1}{c|}{\multirow{2}{*}{\textbf{104}}} & \multicolumn{1}{c|}{\multirow{2}{*}{TO}} & \multirow{2}{*}{2619} & \multicolumn{1}{c|}{\multirow{2}{*}{\textbf{104}}} & \multicolumn{1}{c|}{\multirow{2}{*}{TO}} & \multirow{2}{*}{11445} & \multicolumn{1}{c|}{\multirow{2}{*}{\textbf{104}}} & \multicolumn{1}{c|}{\multirow{2}{*}{TO}} & \multirow{2}{*}{1366} & \multicolumn{1}{c|}{\multirow{2}{*}{-}} & \multicolumn{1}{c|}{\multirow{2}{*}{TO}} & \multirow{2}{*}{9331} \\
 &  &  & \multicolumn{1}{c|}{} & \multicolumn{1}{c|}{} &  & \multicolumn{1}{c|}{} & \multicolumn{1}{c|}{} &  & \multicolumn{1}{c|}{} & \multicolumn{1}{c|}{} &  & \multicolumn{1}{c|}{} & \multicolumn{1}{c|}{} &  & \multicolumn{1}{c|}{} & \multicolumn{1}{c|}{} &  \\ \hline
\multirow{2}{*}{U-662\_bus} & \multirow{2}{*}{219} & \multirow{2}{*}{220} & \multicolumn{1}{c|}{\multirow{2}{*}{\textbf{220}}} & \multicolumn{1}{c|}{\multirow{2}{*}{471}} & \multirow{2}{*}{2498} & \multicolumn{1}{c|}{\multirow{2}{*}{\textbf{220}}} & \multicolumn{1}{c|}{\multirow{2}{*}{144}} & \multirow{2}{*}{1377} & \multicolumn{1}{c|}{\multirow{2}{*}{\textbf{220}}} & \multicolumn{1}{c|}{\multirow{2}{*}{122.9}} & \multirow{2}{*}{2600} & \multicolumn{1}{c|}{\multirow{2}{*}{\textbf{220}}} & \multicolumn{1}{c|}{\multirow{2}{*}{\uline{3.15}}} & \multirow{2}{*}{250.4} & \multicolumn{1}{c|}{\multirow{2}{*}{-}} & \multicolumn{1}{c|}{\multirow{2}{*}{328}} & \multirow{2}{*}{MO} \\
 &  &  & \multicolumn{1}{c|}{} & \multicolumn{1}{c|}{} &  & \multicolumn{1}{c|}{} & \multicolumn{1}{c|}{} &  & \multicolumn{1}{c|}{} & \multicolumn{1}{c|}{} &  & \multicolumn{1}{c|}{} & \multicolumn{1}{c|}{} &  & \multicolumn{1}{c|}{} & \multicolumn{1}{c|}{} &  \\ \hline
\multirow{2}{*}{V-nos6} & \multirow{2}{*}{326} & \multirow{2}{*}{337} & \multicolumn{1}{c|}{\multirow{2}{*}{-}} & \multicolumn{1}{c|}{\multirow{2}{*}{TO}} & \multirow{2}{*}{1836} & \multicolumn{1}{c|}{\multirow{2}{*}{-}} & \multicolumn{1}{c|}{\multirow{2}{*}{TO}} & \multirow{2}{*}{1532} & \multicolumn{1}{c|}{\multirow{2}{*}{-}} & \multicolumn{1}{c|}{\multirow{2}{*}{TO}} & \multirow{2}{*}{6317} & \multicolumn{1}{c|}{\multirow{2}{*}{-}} & \multicolumn{1}{c|}{\multirow{2}{*}{TO}} & \multirow{2}{*}{967.9} & \multicolumn{1}{c|}{\multirow{2}{*}{-}} & \multicolumn{1}{c|}{\multirow{2}{*}{1260}} & \multirow{2}{*}{MO} \\
 &  &  & \multicolumn{1}{c|}{} & \multicolumn{1}{c|}{} &  & \multicolumn{1}{c|}{} & \multicolumn{1}{c|}{} &  & \multicolumn{1}{c|}{} & \multicolumn{1}{c|}{} &  & \multicolumn{1}{c|}{} & \multicolumn{1}{c|}{} &  & \multicolumn{1}{c|}{} & \multicolumn{1}{c|}{} &  \\ \hline
\multirow{2}{*}{W-685\_bus} & \multirow{2}{*}{136} & \multirow{2}{*}{136} & \multicolumn{1}{c|}{\multirow{2}{*}{\textbf{136}}} & \multicolumn{1}{c|}{\multirow{2}{*}{10.7}} & \multirow{2}{*}{1431} & \multicolumn{1}{c|}{\multirow{2}{*}{\textbf{136}}} & \multicolumn{1}{c|}{\multirow{2}{*}{7.57}} & \multirow{2}{*}{1164} & \multicolumn{1}{c|}{\multirow{2}{*}{\textbf{136}}} & \multicolumn{1}{c|}{\multirow{2}{*}{7.6}} & \multirow{2}{*}{1162} & \multicolumn{1}{c|}{\multirow{2}{*}{\textbf{136}}} & \multicolumn{1}{c|}{\multirow{2}{*}{\uline{1.95}}} & \multirow{2}{*}{332.6} & \multicolumn{1}{c|}{\multirow{2}{*}{\textbf{136}}} & \multicolumn{1}{c|}{\multirow{2}{*}{73.3}} & \multirow{2}{*}{8223} \\
 &  &  & \multicolumn{1}{c|}{} & \multicolumn{1}{c|}{} &  & \multicolumn{1}{c|}{} & \multicolumn{1}{c|}{} &  & \multicolumn{1}{c|}{} & \multicolumn{1}{c|}{} &  & \multicolumn{1}{c|}{} & \multicolumn{1}{c|}{} &  & \multicolumn{1}{c|}{} & \multicolumn{1}{c|}{} &  \\ \hline
\multirow{2}{*}{X-can\_\_715} & \multirow{2}{*}{112} & \multirow{2}{*}{142} & \multicolumn{1}{c|}{\multirow{2}{*}{-}} & \multicolumn{1}{c|}{\multirow{2}{*}{TO}} & \multirow{2}{*}{3060} & \multicolumn{1}{c|}{\multirow{2}{*}{113}} & \multicolumn{1}{c|}{\multirow{2}{*}{TO}} & \multirow{2}{*}{4599} & \multicolumn{1}{c|}{\multirow{2}{*}{\textbf{116}}} & \multicolumn{1}{c|}{\multirow{2}{*}{TO}} & \multirow{2}{*}{10388} & \multicolumn{1}{c|}{\multirow{2}{*}{113}} & \multicolumn{1}{c|}{\multirow{2}{*}{TO}} & \multirow{2}{*}{1380} & \multicolumn{1}{c|}{\multirow{2}{*}{-}} & \multicolumn{1}{c|}{\multirow{2}{*}{TO}} & \multirow{2}{*}{11514} \\
 &  &  & \multicolumn{1}{c|}{} & \multicolumn{1}{c|}{} &  & \multicolumn{1}{c|}{} & \multicolumn{1}{c|}{} &  & \multicolumn{1}{c|}{} & \multicolumn{1}{c|}{} &  & \multicolumn{1}{c|}{} & \multicolumn{1}{c|}{} &  & \multicolumn{1}{c|}{} & \multicolumn{1}{c|}{} &  \\ \hline
\end{tabular}
}
\label{tab:result-ABP-5-best-encodings}
\end{table*}

Table~\ref{tab:result-ABP-5-best-encodings} summarizes the experimental results on 12 large problem instances for the five best-performing approaches: \textit{Duplex}, $SCL_{seq}$, $SCL_{par}$, \textit{CPLEX}, and \textit{CP-SAT}. 
Overall, $SCL_{par}$ achieves the best performance among all methods, except for the \textit{O-impcol\_d} instance, where \textit{CPLEX} performs significantly better.
Compared to \textit{CPLEX}, $SCL_{par}$ outperforms it on five instances: \textit{M-bcsstk06}, \textit{N-bcsstk07}, \textit{P-can\_\_445}, \textit{R-dwt\_\_503}, and \textit{X-can\_\_715}.
In the sequential setting, $SCL_{seq}$ consistently matches or improves upon the results obtained by \textit{Duplex} across all 12 large instances, demonstrating that the advantage over \textit{Duplex} is inherent in the proposed \textit{SCL} encoding. Building on this, $SCL_{par}$ represents an improved variant of $SCL_{seq}$ that further exploits parallel search to enhance performance, outperforming \textit{Duplex} on five instances, including \textit{O-impcol\_d}, \textit{P-can\_\_445}, \textit{Q-494\_bus}, \textit{T-dwt\_\_592}, and \textit{X-can\_\_715}. Notably, for \textit{P-can\_\_445} and \textit{X-can\_\_715}, no other approach achieves results comparable to those of $SCL_{par}$.

\subsection{Computation Results for the Nurse Rostering Problem}
\subsubsection{Datasets and Implementation Details}

The third framework evaluates the performance of the \textit{SCL} encoding on a Nurse Rostering Problem (NRP) variant used by Bergman et al. \cite{bergman2014mdd}. This NRP variant involves scheduling for a single nurse over a planning horizon ranging from 40 to 100 days, subject to 11 schedule requirements described in Section~\ref{sec:sat-nrp}.

In this framework, we implemented the $SCL_{\leq k}$ alongside several SAT encoding techniques, including \textit{BDD}, \textit{Card}, \textit{Adder}, \textit{Pairwise}, \textit{SCL}, and \textit{Seq}.
For the SAT-based approaches, constraints 1, 4, and 8 in the ILP model are encoded using \textit{Pairwise} encoding, since they correspond to At-Least-One, At-Most-One, or Exactly-One constraints. For the remaining constraints of ALK type, particularly constraints (2), (3), (6), and (9), if a given SAT encoding cannot handle them directly, we first reformulate these constraints into equivalent AMK constraints, as described in Section~\ref{sec:nrp-sat-implementation}.

We also compare $SCL_{\leq k}$ with state-of-the-art CP and MIP solvers, namely \textit{CPLEX}, \textit{CP-SAT}, and \textit{Gurobi}. As in the experiments on the Anti-bandwidth Problem, the \textit{CPLEX} approach leverages the CP APIs of IBM ILOG CPLEX Optimization Studio version 22.1.1, while \textit{CP-SAT} is provided by Google OR-Tools version 9.11.4210, and \textit{Gurobi} uses Gurobi Optimizer version 11.0.3.

\subsubsection{Performance Evaluation}

\begin{figure*}[ht!]
    \centering
    \begin{minipage}{0.49\textwidth}
        \centering
        \includegraphics[width=\textwidth]{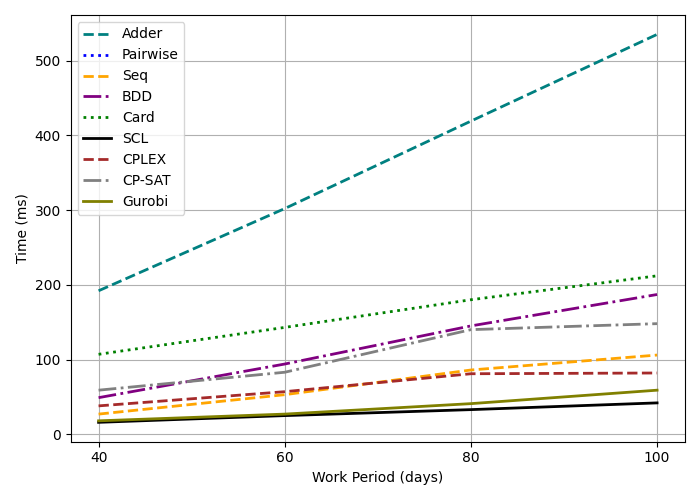}
        \caption{\small NRP solving time of different approaches.}
        \label{fig:nrp-solving-time}
    \end{minipage}%
    \hfill
    \begin{minipage}{0.49\textwidth}
        \centering
        \includegraphics[width=\textwidth]{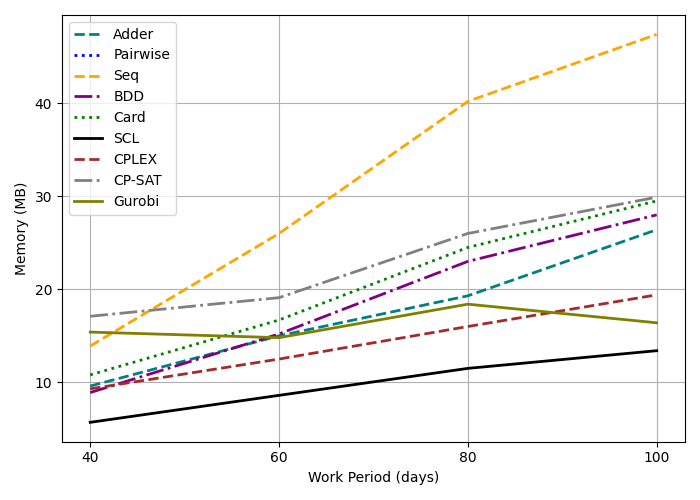}
        \caption{\small NRP memory usage of different approaches.}
        \label{fig:nrp-memory-consumption}
    \end{minipage}
\end{figure*}

Since almost all approaches can solve the NRP dataset quickly, we execute all the approaches on each instance 10 times and report the average solving time and memory consumption. For each run, the time limit and memory limit are set to 600 seconds and 5 GB, respectively.
The results, including both solving time and memory consumption, are summarized in Table~\ref{tab:nrp_solving_results} and also illustrated in Figures~\ref{fig:nrp-solving-time} and \ref{fig:nrp-memory-consumption}. Figure~\ref{fig:nrp-solving-time} presents the solving time, while Figure~\ref{fig:nrp-memory-consumption} reports the memory consumption.

\begin{table}[ht!]
    \centering
    \resizebox{\textwidth}{!}{
    \begin{tabular}{|l|cc|cc|cc|cc|}
    \hline
    \multicolumn{1}{|c|}{\multirow{2}{*}{Work period}} & \multicolumn{2}{c|}{40 days} & \multicolumn{2}{c|}{60 days} & \multicolumn{2}{c|}{80 days} & \multicolumn{2}{c|}{100 days \rule{0pt}{12pt}} \\ \cline{2-9} 
    \multicolumn{1}{|c|}{} & \multicolumn{1}{c|}{Time(ms)} & MB & \multicolumn{1}{c|}{Time(ms)} & MB & \multicolumn{1}{c|}{Time(ms)} & MB & \multicolumn{1}{c|}{Time(ms)} & MB \rule{0pt}{12pt} \\ \hline
    \textit{Adder} & \multicolumn{1}{c|}{192} & 9.6 & \multicolumn{1}{c|}{302} & 15 & \multicolumn{1}{c|}{419} & 19.3 & \multicolumn{1}{c|}{535} & 26.4 \rule{0pt}{12pt} \\ \hline
    \textit{BDD} & \multicolumn{1}{c|}{49} & 8.9 & \multicolumn{1}{c|}{94} & 15.2 & \multicolumn{1}{c|}{145} & 23 & \multicolumn{1}{c|}{187} & 28 \rule{0pt}{12pt} \\ \hline
    \textit{Card} & \multicolumn{1}{c|}{107} & 10.8 & \multicolumn{1}{c|}{143} & 16.7 & \multicolumn{1}{c|}{180} & 24.5 & \multicolumn{1}{c|}{212} & 29.5 \rule{0pt}{12pt} \\ \hline
    \textit{Pairwise} & \multicolumn{1}{c|}{-} & MO & \multicolumn{1}{c|}{-} & MO & \multicolumn{1}{c|}{-} & MO & \multicolumn{1}{c|}{-} & MO \rule{0pt}{12pt} \\ \hline
    \textit{SCL} & \multicolumn{1}{c|}{\textbf{16}} & \textbf{5.7} & \multicolumn{1}{c|}{\textbf{25}} & \textbf{8.6} & \multicolumn{1}{c|}{\textbf{33}} & \textbf{11.5} & \multicolumn{1}{c|}{\textbf{42}} & \textbf{13.4} \rule{0pt}{12pt} \\ \hline
    \textit{Seq} & \multicolumn{1}{c|}{27} & 13.9 & \multicolumn{1}{c|}{53} & 26 & \multicolumn{1}{c|}{86} & 40.2 & \multicolumn{1}{c|}{106} & 47.4 \rule{0pt}{12pt} \\ \hline
    \textit{CPLEX} & \multicolumn{1}{c|}{38} & 9.3 & \multicolumn{1}{c|}{57} & 12.5 & \multicolumn{1}{c|}{81} & 16 & \multicolumn{1}{c|}{82} & 19.4 \rule{0pt}{12pt} \\ \hline
    \textit{CP-SAT} & \multicolumn{1}{c|}{59} & 17.1 & \multicolumn{1}{c|}{83} & 19.1 & \multicolumn{1}{c|}{140} & 26 & \multicolumn{1}{c|}{148} & 29.9 \rule{0pt}{12pt} \\ \hline
    \textit{Gurobi} & \multicolumn{1}{c|}{18} & 15.4 & \multicolumn{1}{c|}{27} & 14.8 & \multicolumn{1}{c|}{41} & 18.4 & \multicolumn{1}{c|}{59} & 16.4 \rule{0pt}{12pt} \\ \hline
    \end{tabular}
    }
    \caption{NRP solving results of different approaches with TO = 600s and MO = 5GB.}
    \label{tab:nrp_solving_results}
\end{table}

In terms of solving time, \textit{SCL} encoding consistently outperforms other SAT-based encodings across all instances. Compared with the best-performing alternative SAT-based encoding, namely \textit{Seq}, \textit{SCL} solves all instances approximately twice as fast. For example, on the smallest instance (40 days), \textit{SCL} finds a feasible schedule in 16 ms, whereas \textit{Seq} requires 27 ms. On the largest instance (100 days), \textit{SCL} requires only 42 ms, compared to 106 ms for \textit{Seq}. These results demonstrate a substantial runtime improvement of \textit{SCL} encoding over existing SAT encodings that encode each cardinality constraint separately.

Importantly, this speed advantage is not achieved at the expense of memory usage. \textit{SCL} encoding also maintains the lowest memory consumption among all SAT-based encodings. Its memory usage increases moderately from 5.7\,MB at 40 days to 13.4\,MB at 100 days. In comparison, \textit{Adder}, \textit{BDD} and \textit{Card} require roughly twice as much memory, while \textit{Seq} consumes approximately three times more. Besides, \textit{Pairwise} encoding fails on all instances due to memory overflow, confirming its lack of scalability when encoding AMK and ALK constraints.

When compared with CP and MIP solvers, \textit{SCL} encoding remains competitive in solving time while maintaining lower memory consumption, especially on medium and large instances. On the 40-day instance, \textit{SCL} finds a feasible schedule in 16 ms, slightly faster than 18 ms of \textit{Gurobi}. The performance gap widens as the problem size increases. On the 100-day instance, \textit{SCL} requires only 42 ms, whereas \textit{Gurobi} needs 59 ms, meaning that \textit{SCL} solves the instance in roughly two-thirds of the time. 

It is also worth noting that all the CP and MIP solvers used in this experiment run in parallel using default settings with 8 concurrent threads, whereas the SAT-based approaches operate in single-threaded mode. The fact that \textit{SCL} achieves better performance despite this difference further highlights its efficiency for problems involving Ladder constraints.

\section{Conclusion}\label{sec13}
 
SAT solving has emerged as a competitive and effective alternative to CP and MIP solvers for various combinatorial, scheduling, and OR problems. Alongside significant advancements in the capabilities of SAT solvers, the development of efficient encodings tailored to specific types of constraints can greatly enhance overall solution performance. Numerous studies have shown that optimizing SAT encoding can lead to substantial improvements, especially when replacing direct encoding techniques with more compact representations. These optimized encodings can reduce the number of auxiliary variables, generated clauses, and clause sizes.

In this paper, we introduced a novel SAT encoding for sets of AMK cardinality constraints arranged in a ladder shape. Rather than encoding each AMK constraint individually, our approach generates encodings for blocks that represent adjacent AMK constraints with the aim of encoding a set of AMK constraints simultaneously. To our knowledge, previous research in this area has mainly focused on using \textit{BDD} to represent Ladder AMO constraints \cite{fazekas2020duplex}, and no proposed encodings exist for Ladder AMK constraints where \( k > 1 \). Compared to the \textit{BDD}-based representation and other SAT-based encodings that encode each AMK constraint separately, the proposed encoding, built upon the \textit{Sequential Counter} encoding, achieves significant improvements for Ladder AMO constraints and demonstrates effectiveness for Ladder AMK constraints.

We further applied the proposed encoding to address the Anti-bandwidth Problem and Nurse Rostering Problem. The experimental results show that our approach is not only competitive with other SAT-based encodings but also capable of outperforming state-of-the-art CP and MIP solvers such as \textit{CPLEX}, \textit{Gurobi}, and \textit{CP-SAT}. 
These findings highlight the practical potential of the proposed encoding for real-world applications, such as Car Sequencing problem \cite{artigues2014sat, siala2015search}, Employee Scheduling problem \cite{nieuwenhuis2021employee}, and various other scheduling challenges featuring sequence constraints with ladder structures.

Finally, while this work focuses on a block-based \textit{SC} encoding tailored to Ladder AMK constraints, exploring the suitability of alternative cardinality encodings for enabling intermediate variable reuse remains an interesting direction for future research. In particular, it would be valuable to systematically investigate how other encoding structures can be adapted or extended to better support sequence constraints, especially Ladder AMK constraints. Such an analysis could provide deeper insights into the interaction between encoding structure and constraint overlap, and may further enhance the performance of SAT-based approaches for Ladder AMK constraints, as well as sequence constraints.

\backmatter

\bmhead{Acknowledgements}

We sincerely thank the reviewers for their careful reading of the manuscript and constructive comments. Their precise and valuable suggestions have significantly improved the clarity, quality, and technical soundness of this work.

\bmhead{Author Contributions}

Hieu Truong Xuan: conceptualization, methodology, formal analysis, investigation, data curation, software, visualization, writing - original draft, review \& editing;\\
Tuyen Kieu Van: methodology, writing - review \& editing;\\
Khanh To Van: conceptualization, investigation, methodology, formal analysis, supervision, funding acquisition, writing - original draft, review \& editing.


\bmhead{Availability of data and materials} 
The experimental results are available at \url{https://github.com/TruongXuanHieu-H/SCL-Encoding}.

\bmhead{Code availability}
The source code for these experiments is also available at \url{https://github.com/TruongXuanHieu-H/SCL-Encoding}.

\section*{Declaration}
\bmhead{Ethics approval} Not applicable.
\bmhead{Consent to participate} Not applicable. 
\bmhead{Consent for publication} Not applicable.  
\bmhead{Conflict of interest} Not applicable. 





\begin{appendices}
\section{Correctness of the SC Encoding}
\label{app:sc-correctness}

In this appendix, we prove that equations \eqref{eq:ladder-amk-init-bit}--\eqref{eq:ladder-amk-restrict-bit} constitute a sound and complete encoding of the At-Most-K (AMK) constraint over decomposed blocks.

Let a block $i$ of $w$ variables, and define
\[
S_{i,j} = \sum_{t=1}^{j} x_{i,t}.
\]

The encoded constraint is:
\[
\sum_{j=1}^{w} x_{i,j} \le k.
\]

The register bits $R_{i,j,s}$ are intended to represent:
\[
R_{i,j,s} \;\;\equiv\;\; (S_{i,j} \ge s),
\quad
1 \le s \le \min(j,k).
\]

\subsection{Soundness}

\begin{theorem}
Any assignment satisfying equations~\eqref{eq:ladder-amk-init-bit}--\eqref{eq:ladder-amk-restrict-bit} satisfies
\[
\sum_{j=1}^{w} x_{i,j} \le k.
\]
\label{theorem:soundness_SC}
\end{theorem}

\begin{proof}
Assume, for contradiction, that there exists a satisfying assignment such that
\[
\sum_{j=1}^{w} x_{i,j} \ge k+1.
\]

Let $j$ be the smallest index such that
\[
S_{i,j} = k+1.
\]

Then necessarily
\[
S_{i,j-1} = k
\quad \text{and} \quad
x_{i,j} = 1.
\]


Since \(S_{i,j-1} = k\), $R_{i,j-1,k} = 1$ (or $true$).

However, equation~\eqref{eq:ladder-amk-restrict-bit} enforces, for all $j > k$,
\[
x_{i,j} \rightarrow \neg R_{i,j-1,k}.
\]

Since $x_{i,j}=1$ and $R_{i,j-1,k}=1$,
we obtain a contradiction.
Therefore,
\[
\sum_{j=1}^{w} x_{i,j} \le k.
\]
\end{proof}

\subsection{Completeness}

\begin{theorem}
If
\[
\sum_{j=1}^{w} x_{i,j} \le k,
\]
then there exists an assignment to the register bits $R_{i,j,s}$ satisfying equations~\eqref{eq:ladder-amk-init-bit}--\eqref{eq:ladder-amk-restrict-bit}.
\label{theorem:completeness_SC}
\end{theorem}

\begin{proof}
Assume
\[
\sum_{j=1}^{w} x_{i,j} \le k.
\]

Define the register bits as follows:
\[
R_{i,j,s} :=
\begin{cases}
1 & \text{if } S_{i,j} \ge s, \\
0 & \text{otherwise.}
\end{cases}
\]

By construction, the relation
\(
S_{i,j} = S_{i,j-1} + x_{i,j}
\)
ensures that equations~\eqref{eq:ladder-amk-init-bit}--\eqref{eq:ladder-amk-derive-true-bit} are satisfied.

Similarly, if $S_{i,j} < s$, then either $S_{i,j-1} < s - 1$ or ($S_{i,j-1} < s$ and $x_{i,j}=0$),
which guarantees that equations~\eqref{eq:ladder-amk-init-false-bit}--\eqref{eq:ladder-amk-derive-false-bit} are satisfied.

Finally, since $S_{i,j} \le k$ for all $j$, we never have both $x_{i,j}=1$ and $S_{i,j-1}=k$ simultaneously. Hence, equation~\eqref{eq:ladder-amk-restrict-bit} is satisfied.

Therefore, a satisfying assignment of the register bits exists.
\end{proof}

\medskip

\noindent
Together, Theorems \ref{theorem:soundness_SC} and \ref{theorem:completeness_SC} establish that the \textit{SC} encoding is sound and complete for encoding AMK constraints.
\end{appendices}


\bibliography{sn-bibliography}

\end{document}